\pdfoutput=1
\documentclass[11pt]{article}

\usepackage[preprint]{acl}

\usepackage{times}
\usepackage{latexsym}
\usepackage[T1]{fontenc}
\usepackage[utf8]{inputenc}
\usepackage{microtype}
\usepackage{inconsolata}
\usepackage{graphicx}
\usepackage{enumitem}
\usepackage{xcolor}
\usepackage{soul}
\usepackage{multirow}
\usepackage{array}
\usepackage{lscape}
\usepackage{tcolorbox}
\usepackage{amsmath}
\usepackage{subcaption}
\usepackage{lineno}
\usepackage{longtable}
\usepackage{caption}
\usepackage{xurl}
\usepackage{float}
\usepackage{dingbat}
\usepackage{pifont}
\usepackage{fontawesome5}    %

\newcommand{\hficon}{\raisebox{-0.15em}{\includegraphics[height=0.9em]{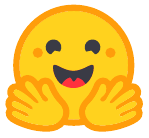}}~}
\newcommand{\ghicon}{\faGithub~}

\usepackage{booktabs}        %
\usepackage{tabularx}        %
\usepackage{adjustbox}       %

\usepackage{siunitx}

\newcommand{\PreserveBackslash}[1]{\let\temp=\\#1\let\\=\temp}
\newcolumntype{C}[1]{>{\PreserveBackslash\centering}p{#1}}
\newcolumntype{R}[1]{>{\PreserveBackslash\raggedleft}p{#1}}
\newcolumntype{L}[1]{>{\PreserveBackslash\raggedright}p{#1}}

\newcolumntype{P}[1]{>{\centering\arraybackslash}p{#1}}

\usepackage{colortbl}
\usepackage{dblfloatfix}
\usepackage{placeins}        %

\definecolor{hidecolor}{rgb}{1,1,1,0.0} 
\sethlcolor{hidecolor}

\sethlcolor{yellow}

\definecolor{lightblue}{RGB}{173,216,230}

\definecolor{lightgreen}{RGB}{144,238,144}

\definecolor{lightorange}{RGB}{255,200,124}

\definecolor{lightred}{RGB}{255,182,193}

\title{Prompting Is All You Need: Multi-view Prompting Large Language Models for Aspect-Based Sentiment Analysis} %

\author{
Nils Constantin Hellwig$^{1}$ \quad Niklas Donhauser$^{1}$ \quad Jakob Fehle$^{1}$ \\
\normalfont \textbf{Udo Kruschwitz}$^{2}$ \quad \textbf{Christian Wolff}$^{1}$ \\[0.5em]
$^{1}$Media Informatics Group, University of Regensburg, Germany \\
$^{2}$Information Science Group, University of Regensburg, Germany \\[0.5em]
\texttt{\{nils-constantin.hellwig,niklas.donhauser,jakob.fehle,} \\
\texttt{udo.kruschwitz,christian.wolff\}@ur.de}
}

\providecommand{\val}[1]{\ifcsname llmmvpval:#1\endcsname\csname llmmvpval:#1\endcsname\else\textbf{??#1??}\PackageWarning{numbers}{Undefined value `#1'}\fi}
\expandafter\def\csname llmmvpval:energy.asqp.ft\endcsname{15.92}
\expandafter\def\csname llmmvpval:energy.asqp.mvp.0\endcsname{219.23}
\expandafter\def\csname llmmvpval:energy.asqp.mvp.0.int\endcsname{219}
\expandafter\def\csname llmmvpval:energy.asqp.mvp.10\endcsname{264.86}
\expandafter\def\csname llmmvpval:energy.asqp.mvp.10.int\endcsname{265}
\expandafter\def\csname llmmvpval:energy.asqp.mvp.100\endcsname{352.60}
\expandafter\def\csname llmmvpval:energy.asqp.mvp.100.int\endcsname{353}
\expandafter\def\csname llmmvpval:energy.asqp.mvp.50\endcsname{307.62}
\expandafter\def\csname llmmvpval:energy.asqp.mvp.50.int\endcsname{308}
\expandafter\def\csname llmmvpval:energy.asqp.reasoning.0\endcsname{14.21}
\expandafter\def\csname llmmvpval:energy.asqp.reasoning.10\endcsname{16.40}
\expandafter\def\csname llmmvpval:energy.asqp.reasoning.100\endcsname{25.29}
\expandafter\def\csname llmmvpval:energy.asqp.reasoning.50\endcsname{20.87}
\expandafter\def\csname llmmvpval:energy.asqp.sc-nc.0\endcsname{248.99}
\expandafter\def\csname llmmvpval:energy.asqp.sc-nc.10\endcsname{440.01}
\expandafter\def\csname llmmvpval:energy.asqp.sc-nc.100\endcsname{2618.89}
\expandafter\def\csname llmmvpval:energy.asqp.sc-nc.50\endcsname{1363.75}
\expandafter\def\csname llmmvpval:energy.asqp.sc.0\endcsname{34.85}
\expandafter\def\csname llmmvpval:energy.asqp.sc.10\endcsname{33.41}
\expandafter\def\csname llmmvpval:energy.asqp.sc.100\endcsname{40.32}
\expandafter\def\csname llmmvpval:energy.asqp.sc.50\endcsname{37.92}
\expandafter\def\csname llmmvpval:energy.asqp.so-nc.0\endcsname{50.27}
\expandafter\def\csname llmmvpval:energy.asqp.so-nc.0.int\endcsname{50}
\expandafter\def\csname llmmvpval:energy.asqp.so-nc.10\endcsname{88.15}
\expandafter\def\csname llmmvpval:energy.asqp.so-nc.10.int\endcsname{88}
\expandafter\def\csname llmmvpval:energy.asqp.so-nc.100\endcsname{516.45}
\expandafter\def\csname llmmvpval:energy.asqp.so-nc.100.int\endcsname{516}
\expandafter\def\csname llmmvpval:energy.asqp.so-nc.50\endcsname{270.39}
\expandafter\def\csname llmmvpval:energy.asqp.so-nc.50.int\endcsname{270}
\expandafter\def\csname llmmvpval:energy.asqp.so.0\endcsname{9.28}
\expandafter\def\csname llmmvpval:energy.asqp.so.10\endcsname{10.62}
\expandafter\def\csname llmmvpval:energy.asqp.so.100\endcsname{14.09}
\expandafter\def\csname llmmvpval:energy.asqp.so.50\endcsname{12.33}
\expandafter\def\csname llmmvpval:energy.tasd.ft\endcsname{12.61}
\expandafter\def\csname llmmvpval:energy.tasd.mvp.0\endcsname{49.65}
\expandafter\def\csname llmmvpval:energy.tasd.mvp.0.int\endcsname{50}
\expandafter\def\csname llmmvpval:energy.tasd.mvp.10\endcsname{43.06}
\expandafter\def\csname llmmvpval:energy.tasd.mvp.10.int\endcsname{43}
\expandafter\def\csname llmmvpval:energy.tasd.mvp.100\endcsname{64.41}
\expandafter\def\csname llmmvpval:energy.tasd.mvp.100.int\endcsname{64}
\expandafter\def\csname llmmvpval:energy.tasd.mvp.50\endcsname{54.00}
\expandafter\def\csname llmmvpval:energy.tasd.mvp.50.int\endcsname{54}
\expandafter\def\csname llmmvpval:energy.tasd.mvp.max\endcsname{64.41}
\expandafter\def\csname llmmvpval:energy.tasd.mvp.min\endcsname{43.06}
\expandafter\def\csname llmmvpval:energy.tasd.reasoning.0\endcsname{10.25}
\expandafter\def\csname llmmvpval:energy.tasd.reasoning.10\endcsname{13.29}
\expandafter\def\csname llmmvpval:energy.tasd.reasoning.100\endcsname{19.17}
\expandafter\def\csname llmmvpval:energy.tasd.reasoning.50\endcsname{16.68}
\expandafter\def\csname llmmvpval:energy.tasd.sc-nc.0\endcsname{206.57}
\expandafter\def\csname llmmvpval:energy.tasd.sc-nc.10\endcsname{366.24}
\expandafter\def\csname llmmvpval:energy.tasd.sc-nc.100\endcsname{2107.03}
\expandafter\def\csname llmmvpval:energy.tasd.sc-nc.50\endcsname{1144.76}
\expandafter\def\csname llmmvpval:energy.tasd.sc.0\endcsname{26.91}
\expandafter\def\csname llmmvpval:energy.tasd.sc.10\endcsname{23.94}
\expandafter\def\csname llmmvpval:energy.tasd.sc.100\endcsname{28.96}
\expandafter\def\csname llmmvpval:energy.tasd.sc.50\endcsname{27.04}
\expandafter\def\csname llmmvpval:energy.tasd.so-nc.0\endcsname{41.49}
\expandafter\def\csname llmmvpval:energy.tasd.so-nc.0.int\endcsname{41}
\expandafter\def\csname llmmvpval:energy.tasd.so-nc.10\endcsname{72.99}
\expandafter\def\csname llmmvpval:energy.tasd.so-nc.10.int\endcsname{73}
\expandafter\def\csname llmmvpval:energy.tasd.so-nc.100\endcsname{417.06}
\expandafter\def\csname llmmvpval:energy.tasd.so-nc.100.int\endcsname{417}
\expandafter\def\csname llmmvpval:energy.tasd.so-nc.50\endcsname{226.65}
\expandafter\def\csname llmmvpval:energy.tasd.so-nc.50.int\endcsname{227}
\expandafter\def\csname llmmvpval:energy.tasd.so.0\endcsname{7.77}
\expandafter\def\csname llmmvpval:energy.tasd.so.10\endcsname{7.21}
\expandafter\def\csname llmmvpval:energy.tasd.so.100\endcsname{10.26}
\expandafter\def\csname llmmvpval:energy.tasd.so.50\endcsname{9.44}
\expandafter\def\csname llmmvpval:f1.coursera.asqp.ft-llm\endcsname{33.66}
\expandafter\def\csname llmmvpval:f1.coursera.asqp.ft-mvp\endcsname{32.50}
\expandafter\def\csname llmmvpval:f1.coursera.asqp.ft-para\endcsname{32.34}
\expandafter\def\csname llmmvpval:f1.coursera.asqp.mvp.0\endcsname{9.46}
\expandafter\def\csname llmmvpval:f1.coursera.asqp.mvp.10\endcsname{22.93}
\expandafter\def\csname llmmvpval:f1.coursera.asqp.mvp.100\endcsname{30.13}
\expandafter\def\csname llmmvpval:f1.coursera.asqp.mvp.50\endcsname{28.75}
\expandafter\def\csname llmmvpval:f1.coursera.asqp.mvp.avg\endcsname{22.82}
\expandafter\def\csname llmmvpval:f1.coursera.asqp.mvpeff.0\endcsname{7.17}
\expandafter\def\csname llmmvpval:f1.coursera.asqp.mvpeff.10\endcsname{22.12}
\expandafter\def\csname llmmvpval:f1.coursera.asqp.mvpeff.100\endcsname{28.49}
\expandafter\def\csname llmmvpval:f1.coursera.asqp.mvpeff.50\endcsname{27.50}
\expandafter\def\csname llmmvpval:f1.coursera.asqp.nogd.0\endcsname{5.25}
\expandafter\def\csname llmmvpval:f1.coursera.asqp.nogd.10\endcsname{23.55}
\expandafter\def\csname llmmvpval:f1.coursera.asqp.nogd.100\endcsname{32.28}
\expandafter\def\csname llmmvpval:f1.coursera.asqp.nogd.50\endcsname{30.12}
\expandafter\def\csname llmmvpval:f1.coursera.asqp.reasoning.0\endcsname{13.57}
\expandafter\def\csname llmmvpval:f1.coursera.asqp.reasoning.10\endcsname{19.71}
\expandafter\def\csname llmmvpval:f1.coursera.asqp.reasoning.100\endcsname{23.97}
\expandafter\def\csname llmmvpval:f1.coursera.asqp.reasoning.50\endcsname{21.97}
\expandafter\def\csname llmmvpval:f1.coursera.asqp.reasoning.avg\endcsname{19.81}
\expandafter\def\csname llmmvpval:f1.coursera.asqp.sc.0\endcsname{7.47}
\expandafter\def\csname llmmvpval:f1.coursera.asqp.sc.10\endcsname{24.28}
\expandafter\def\csname llmmvpval:f1.coursera.asqp.sc.100\endcsname{30.70}
\expandafter\def\csname llmmvpval:f1.coursera.asqp.sc.50\endcsname{29.50}
\expandafter\def\csname llmmvpval:f1.coursera.asqp.sc.avg\endcsname{22.99}
\expandafter\def\csname llmmvpval:f1.coursera.asqp.so.0\endcsname{6.26}
\expandafter\def\csname llmmvpval:f1.coursera.asqp.so.10\endcsname{22.01}
\expandafter\def\csname llmmvpval:f1.coursera.asqp.so.100\endcsname{28.14}
\expandafter\def\csname llmmvpval:f1.coursera.asqp.so.50\endcsname{27.49}
\expandafter\def\csname llmmvpval:f1.coursera.asqp.so.avg\endcsname{20.97}
\expandafter\def\csname llmmvpval:f1.coursera.tasd.ft-llm\endcsname{52.94}
\expandafter\def\csname llmmvpval:f1.coursera.tasd.ft-mvp\endcsname{50.97}
\expandafter\def\csname llmmvpval:f1.coursera.tasd.ft-para\endcsname{51.86}
\expandafter\def\csname llmmvpval:f1.coursera.tasd.mvp.0\endcsname{21.81}
\expandafter\def\csname llmmvpval:f1.coursera.tasd.mvp.10\endcsname{41.01}
\expandafter\def\csname llmmvpval:f1.coursera.tasd.mvp.100\endcsname{49.12}
\expandafter\def\csname llmmvpval:f1.coursera.tasd.mvp.50\endcsname{47.22}
\expandafter\def\csname llmmvpval:f1.coursera.tasd.mvp.avg\endcsname{39.79}
\expandafter\def\csname llmmvpval:f1.coursera.tasd.mvpeff.0\endcsname{21.64}
\expandafter\def\csname llmmvpval:f1.coursera.tasd.mvpeff.10\endcsname{39.19}
\expandafter\def\csname llmmvpval:f1.coursera.tasd.mvpeff.100\endcsname{46.84}
\expandafter\def\csname llmmvpval:f1.coursera.tasd.mvpeff.50\endcsname{44.91}
\expandafter\def\csname llmmvpval:f1.coursera.tasd.nogd.0\endcsname{20.22}
\expandafter\def\csname llmmvpval:f1.coursera.tasd.nogd.10\endcsname{39.19}
\expandafter\def\csname llmmvpval:f1.coursera.tasd.nogd.100\endcsname{48.83}
\expandafter\def\csname llmmvpval:f1.coursera.tasd.nogd.50\endcsname{47.17}
\expandafter\def\csname llmmvpval:f1.coursera.tasd.reasoning.0\endcsname{26.26}
\expandafter\def\csname llmmvpval:f1.coursera.tasd.reasoning.10\endcsname{33.50}
\expandafter\def\csname llmmvpval:f1.coursera.tasd.reasoning.100\endcsname{39.96}
\expandafter\def\csname llmmvpval:f1.coursera.tasd.reasoning.50\endcsname{36.92}
\expandafter\def\csname llmmvpval:f1.coursera.tasd.reasoning.avg\endcsname{34.16}
\expandafter\def\csname llmmvpval:f1.coursera.tasd.sc.0\endcsname{22.36}
\expandafter\def\csname llmmvpval:f1.coursera.tasd.sc.10\endcsname{39.43}
\expandafter\def\csname llmmvpval:f1.coursera.tasd.sc.100\endcsname{47.20}
\expandafter\def\csname llmmvpval:f1.coursera.tasd.sc.50\endcsname{44.42}
\expandafter\def\csname llmmvpval:f1.coursera.tasd.sc.avg\endcsname{38.35}
\expandafter\def\csname llmmvpval:f1.coursera.tasd.so.0\endcsname{21.13}
\expandafter\def\csname llmmvpval:f1.coursera.tasd.so.10\endcsname{38.52}
\expandafter\def\csname llmmvpval:f1.coursera.tasd.so.100\endcsname{46.09}
\expandafter\def\csname llmmvpval:f1.coursera.tasd.so.50\endcsname{44.06}
\expandafter\def\csname llmmvpval:f1.coursera.tasd.so.avg\endcsname{37.45}
\expandafter\def\csname llmmvpval:f1.flightabsa.asqp.ft-llm\endcsname{64.49}
\expandafter\def\csname llmmvpval:f1.flightabsa.asqp.ft-mvp\endcsname{57.90}
\expandafter\def\csname llmmvpval:f1.flightabsa.asqp.ft-para\endcsname{57.76}
\expandafter\def\csname llmmvpval:f1.flightabsa.asqp.mvp.0\endcsname{34.06}
\expandafter\def\csname llmmvpval:f1.flightabsa.asqp.mvp.10\endcsname{50.47}
\expandafter\def\csname llmmvpval:f1.flightabsa.asqp.mvp.100\endcsname{57.37}
\expandafter\def\csname llmmvpval:f1.flightabsa.asqp.mvp.50\endcsname{55.80}
\expandafter\def\csname llmmvpval:f1.flightabsa.asqp.mvp.avg\endcsname{49.42}
\expandafter\def\csname llmmvpval:f1.flightabsa.asqp.mvpeff.0\endcsname{25.62}
\expandafter\def\csname llmmvpval:f1.flightabsa.asqp.mvpeff.10\endcsname{48.20}
\expandafter\def\csname llmmvpval:f1.flightabsa.asqp.mvpeff.100\endcsname{55.62}
\expandafter\def\csname llmmvpval:f1.flightabsa.asqp.mvpeff.50\endcsname{53.91}
\expandafter\def\csname llmmvpval:f1.flightabsa.asqp.nogd.0\endcsname{27.26}
\expandafter\def\csname llmmvpval:f1.flightabsa.asqp.nogd.10\endcsname{49.07}
\expandafter\def\csname llmmvpval:f1.flightabsa.asqp.nogd.100\endcsname{56.82}
\expandafter\def\csname llmmvpval:f1.flightabsa.asqp.nogd.50\endcsname{54.64}
\expandafter\def\csname llmmvpval:f1.flightabsa.asqp.reasoning.0\endcsname{38.43}
\expandafter\def\csname llmmvpval:f1.flightabsa.asqp.reasoning.10\endcsname{42.62}
\expandafter\def\csname llmmvpval:f1.flightabsa.asqp.reasoning.100\endcsname{45.28}
\expandafter\def\csname llmmvpval:f1.flightabsa.asqp.reasoning.50\endcsname{46.88}
\expandafter\def\csname llmmvpval:f1.flightabsa.asqp.reasoning.avg\endcsname{43.30}
\expandafter\def\csname llmmvpval:f1.flightabsa.asqp.sc.0\endcsname{28.62}
\expandafter\def\csname llmmvpval:f1.flightabsa.asqp.sc.10\endcsname{48.75}
\expandafter\def\csname llmmvpval:f1.flightabsa.asqp.sc.100\endcsname{56.24}
\expandafter\def\csname llmmvpval:f1.flightabsa.asqp.sc.50\endcsname{55.12}
\expandafter\def\csname llmmvpval:f1.flightabsa.asqp.sc.avg\endcsname{47.18}
\expandafter\def\csname llmmvpval:f1.flightabsa.asqp.so.0\endcsname{24.32}
\expandafter\def\csname llmmvpval:f1.flightabsa.asqp.so.10\endcsname{47.29}
\expandafter\def\csname llmmvpval:f1.flightabsa.asqp.so.100\endcsname{54.90}
\expandafter\def\csname llmmvpval:f1.flightabsa.asqp.so.50\endcsname{53.33}
\expandafter\def\csname llmmvpval:f1.flightabsa.asqp.so.avg\endcsname{44.96}
\expandafter\def\csname llmmvpval:f1.flightabsa.tasd.ft-llm\endcsname{73.36}
\expandafter\def\csname llmmvpval:f1.flightabsa.tasd.ft-mvp\endcsname{68.67}
\expandafter\def\csname llmmvpval:f1.flightabsa.tasd.ft-para\endcsname{69.74}
\expandafter\def\csname llmmvpval:f1.flightabsa.tasd.mvp.0\endcsname{39.11}
\expandafter\def\csname llmmvpval:f1.flightabsa.tasd.mvp.10\endcsname{62.22}
\expandafter\def\csname llmmvpval:f1.flightabsa.tasd.mvp.100\endcsname{70.11}
\expandafter\def\csname llmmvpval:f1.flightabsa.tasd.mvp.50\endcsname{68.36}
\expandafter\def\csname llmmvpval:f1.flightabsa.tasd.mvp.avg\endcsname{59.95}
\expandafter\def\csname llmmvpval:f1.flightabsa.tasd.mvpeff.0\endcsname{37.62}
\expandafter\def\csname llmmvpval:f1.flightabsa.tasd.mvpeff.10\endcsname{60.74}
\expandafter\def\csname llmmvpval:f1.flightabsa.tasd.mvpeff.100\endcsname{68.44}
\expandafter\def\csname llmmvpval:f1.flightabsa.tasd.mvpeff.50\endcsname{67.15}
\expandafter\def\csname llmmvpval:f1.flightabsa.tasd.nogd.0\endcsname{37.07}
\expandafter\def\csname llmmvpval:f1.flightabsa.tasd.nogd.10\endcsname{60.55}
\expandafter\def\csname llmmvpval:f1.flightabsa.tasd.nogd.100\endcsname{69.56}
\expandafter\def\csname llmmvpval:f1.flightabsa.tasd.nogd.50\endcsname{67.66}
\expandafter\def\csname llmmvpval:f1.flightabsa.tasd.reasoning.0\endcsname{54.34}
\expandafter\def\csname llmmvpval:f1.flightabsa.tasd.reasoning.10\endcsname{58.62}
\expandafter\def\csname llmmvpval:f1.flightabsa.tasd.reasoning.100\endcsname{63.09}
\expandafter\def\csname llmmvpval:f1.flightabsa.tasd.reasoning.50\endcsname{61.43}
\expandafter\def\csname llmmvpval:f1.flightabsa.tasd.reasoning.avg\endcsname{59.37}
\expandafter\def\csname llmmvpval:f1.flightabsa.tasd.sc.0\endcsname{40.96}
\expandafter\def\csname llmmvpval:f1.flightabsa.tasd.sc.10\endcsname{59.07}
\expandafter\def\csname llmmvpval:f1.flightabsa.tasd.sc.100\endcsname{68.17}
\expandafter\def\csname llmmvpval:f1.flightabsa.tasd.sc.50\endcsname{67.43}
\expandafter\def\csname llmmvpval:f1.flightabsa.tasd.sc.avg\endcsname{58.91}
\expandafter\def\csname llmmvpval:f1.flightabsa.tasd.so.0\endcsname{37.93}
\expandafter\def\csname llmmvpval:f1.flightabsa.tasd.so.10\endcsname{59.27}
\expandafter\def\csname llmmvpval:f1.flightabsa.tasd.so.100\endcsname{67.59}
\expandafter\def\csname llmmvpval:f1.flightabsa.tasd.so.50\endcsname{66.38}
\expandafter\def\csname llmmvpval:f1.flightabsa.tasd.so.avg\endcsname{57.79}
\expandafter\def\csname llmmvpval:f1.hotels.asqp.ft-llm\endcsname{56.54}
\expandafter\def\csname llmmvpval:f1.hotels.asqp.ft-mvp\endcsname{55.03}
\expandafter\def\csname llmmvpval:f1.hotels.asqp.ft-para\endcsname{53.87}
\expandafter\def\csname llmmvpval:f1.hotels.asqp.mvp.0\endcsname{27.40}
\expandafter\def\csname llmmvpval:f1.hotels.asqp.mvp.10\endcsname{47.90}
\expandafter\def\csname llmmvpval:f1.hotels.asqp.mvp.100\endcsname{54.16}
\expandafter\def\csname llmmvpval:f1.hotels.asqp.mvp.50\endcsname{52.78}
\expandafter\def\csname llmmvpval:f1.hotels.asqp.mvp.avg\endcsname{45.56}
\expandafter\def\csname llmmvpval:f1.hotels.asqp.mvpeff.0\endcsname{19.66}
\expandafter\def\csname llmmvpval:f1.hotels.asqp.mvpeff.10\endcsname{45.38}
\expandafter\def\csname llmmvpval:f1.hotels.asqp.mvpeff.100\endcsname{50.48}
\expandafter\def\csname llmmvpval:f1.hotels.asqp.mvpeff.50\endcsname{48.27}
\expandafter\def\csname llmmvpval:f1.hotels.asqp.nogd.0\endcsname{18.11}
\expandafter\def\csname llmmvpval:f1.hotels.asqp.nogd.10\endcsname{46.15}
\expandafter\def\csname llmmvpval:f1.hotels.asqp.nogd.100\endcsname{52.65}
\expandafter\def\csname llmmvpval:f1.hotels.asqp.nogd.50\endcsname{50.53}
\expandafter\def\csname llmmvpval:f1.hotels.asqp.reasoning.0\endcsname{32.61}
\expandafter\def\csname llmmvpval:f1.hotels.asqp.reasoning.10\endcsname{39.91}
\expandafter\def\csname llmmvpval:f1.hotels.asqp.reasoning.100\endcsname{42.26}
\expandafter\def\csname llmmvpval:f1.hotels.asqp.reasoning.50\endcsname{42.30}
\expandafter\def\csname llmmvpval:f1.hotels.asqp.reasoning.avg\endcsname{39.27}
\expandafter\def\csname llmmvpval:f1.hotels.asqp.sc.0\endcsname{21.21}
\expandafter\def\csname llmmvpval:f1.hotels.asqp.sc.10\endcsname{47.23}
\expandafter\def\csname llmmvpval:f1.hotels.asqp.sc.100\endcsname{50.95}
\expandafter\def\csname llmmvpval:f1.hotels.asqp.sc.50\endcsname{49.45}
\expandafter\def\csname llmmvpval:f1.hotels.asqp.sc.avg\endcsname{42.21}
\expandafter\def\csname llmmvpval:f1.hotels.asqp.so.0\endcsname{18.91}
\expandafter\def\csname llmmvpval:f1.hotels.asqp.so.10\endcsname{44.27}
\expandafter\def\csname llmmvpval:f1.hotels.asqp.so.100\endcsname{49.66}
\expandafter\def\csname llmmvpval:f1.hotels.asqp.so.50\endcsname{47.74}
\expandafter\def\csname llmmvpval:f1.hotels.asqp.so.avg\endcsname{40.15}
\expandafter\def\csname llmmvpval:f1.hotels.tasd.ft-llm\endcsname{71.47}
\expandafter\def\csname llmmvpval:f1.hotels.tasd.ft-mvp\endcsname{69.37}
\expandafter\def\csname llmmvpval:f1.hotels.tasd.ft-para\endcsname{67.70}
\expandafter\def\csname llmmvpval:f1.hotels.tasd.mvp.0\endcsname{37.07}
\expandafter\def\csname llmmvpval:f1.hotels.tasd.mvp.10\endcsname{64.33}
\expandafter\def\csname llmmvpval:f1.hotels.tasd.mvp.100\endcsname{68.38}
\expandafter\def\csname llmmvpval:f1.hotels.tasd.mvp.50\endcsname{67.42}
\expandafter\def\csname llmmvpval:f1.hotels.tasd.mvp.avg\endcsname{59.30}
\expandafter\def\csname llmmvpval:f1.hotels.tasd.mvpeff.0\endcsname{34.81}
\expandafter\def\csname llmmvpval:f1.hotels.tasd.mvpeff.10\endcsname{62.69}
\expandafter\def\csname llmmvpval:f1.hotels.tasd.mvpeff.100\endcsname{67.24}
\expandafter\def\csname llmmvpval:f1.hotels.tasd.mvpeff.50\endcsname{65.49}
\expandafter\def\csname llmmvpval:f1.hotels.tasd.nogd.0\endcsname{36.52}
\expandafter\def\csname llmmvpval:f1.hotels.tasd.nogd.10\endcsname{62.11}
\expandafter\def\csname llmmvpval:f1.hotels.tasd.nogd.100\endcsname{67.06}
\expandafter\def\csname llmmvpval:f1.hotels.tasd.nogd.50\endcsname{67.07}
\expandafter\def\csname llmmvpval:f1.hotels.tasd.reasoning.0\endcsname{44.09}
\expandafter\def\csname llmmvpval:f1.hotels.tasd.reasoning.10\endcsname{53.88}
\expandafter\def\csname llmmvpval:f1.hotels.tasd.reasoning.100\endcsname{56.75}
\expandafter\def\csname llmmvpval:f1.hotels.tasd.reasoning.50\endcsname{57.26}
\expandafter\def\csname llmmvpval:f1.hotels.tasd.reasoning.avg\endcsname{53.00}
\expandafter\def\csname llmmvpval:f1.hotels.tasd.sc.0\endcsname{36.94}
\expandafter\def\csname llmmvpval:f1.hotels.tasd.sc.10\endcsname{62.20}
\expandafter\def\csname llmmvpval:f1.hotels.tasd.sc.100\endcsname{66.72}
\expandafter\def\csname llmmvpval:f1.hotels.tasd.sc.50\endcsname{65.14}
\expandafter\def\csname llmmvpval:f1.hotels.tasd.sc.avg\endcsname{57.75}
\expandafter\def\csname llmmvpval:f1.hotels.tasd.so.0\endcsname{33.93}
\expandafter\def\csname llmmvpval:f1.hotels.tasd.so.10\endcsname{61.09}
\expandafter\def\csname llmmvpval:f1.hotels.tasd.so.100\endcsname{66.06}
\expandafter\def\csname llmmvpval:f1.hotels.tasd.so.50\endcsname{64.44}
\expandafter\def\csname llmmvpval:f1.hotels.tasd.so.avg\endcsname{56.38}
\expandafter\def\csname llmmvpval:f1.rest15.asqp.ft-llm\endcsname{56.78}
\expandafter\def\csname llmmvpval:f1.rest15.asqp.ft-mvp\endcsname{51.04}
\expandafter\def\csname llmmvpval:f1.rest15.asqp.ft-para\endcsname{46.93}
\expandafter\def\csname llmmvpval:f1.rest15.asqp.mvp.0\endcsname{32.88}
\expandafter\def\csname llmmvpval:f1.rest15.asqp.mvp.10\endcsname{45.56}
\expandafter\def\csname llmmvpval:f1.rest15.asqp.mvp.100\endcsname{53.40}
\expandafter\def\csname llmmvpval:f1.rest15.asqp.mvp.50\endcsname{50.99}
\expandafter\def\csname llmmvpval:f1.rest15.asqp.mvp.avg\endcsname{45.71}
\expandafter\def\csname llmmvpval:f1.rest15.asqp.mvpeff.0\endcsname{27.51}
\expandafter\def\csname llmmvpval:f1.rest15.asqp.mvpeff.10\endcsname{42.08}
\expandafter\def\csname llmmvpval:f1.rest15.asqp.mvpeff.100\endcsname{51.99}
\expandafter\def\csname llmmvpval:f1.rest15.asqp.mvpeff.50\endcsname{49.19}
\expandafter\def\csname llmmvpval:f1.rest15.asqp.nogd.0\endcsname{20.33}
\expandafter\def\csname llmmvpval:f1.rest15.asqp.nogd.10\endcsname{43.00}
\expandafter\def\csname llmmvpval:f1.rest15.asqp.nogd.100\endcsname{52.65}
\expandafter\def\csname llmmvpval:f1.rest15.asqp.nogd.50\endcsname{49.91}
\expandafter\def\csname llmmvpval:f1.rest15.asqp.reasoning.0\endcsname{33.33}
\expandafter\def\csname llmmvpval:f1.rest15.asqp.reasoning.10\endcsname{36.39}
\expandafter\def\csname llmmvpval:f1.rest15.asqp.reasoning.100\endcsname{39.95}
\expandafter\def\csname llmmvpval:f1.rest15.asqp.reasoning.50\endcsname{37.51}
\expandafter\def\csname llmmvpval:f1.rest15.asqp.reasoning.avg\endcsname{36.79}
\expandafter\def\csname llmmvpval:f1.rest15.asqp.sc.0\endcsname{27.10}
\expandafter\def\csname llmmvpval:f1.rest15.asqp.sc.10\endcsname{41.99}
\expandafter\def\csname llmmvpval:f1.rest15.asqp.sc.100\endcsname{51.74}
\expandafter\def\csname llmmvpval:f1.rest15.asqp.sc.50\endcsname{49.27}
\expandafter\def\csname llmmvpval:f1.rest15.asqp.sc.avg\endcsname{42.53}
\expandafter\def\csname llmmvpval:f1.rest15.asqp.so.0\endcsname{25.59}
\expandafter\def\csname llmmvpval:f1.rest15.asqp.so.10\endcsname{40.43}
\expandafter\def\csname llmmvpval:f1.rest15.asqp.so.100\endcsname{50.70}
\expandafter\def\csname llmmvpval:f1.rest15.asqp.so.50\endcsname{47.99}
\expandafter\def\csname llmmvpval:f1.rest15.asqp.so.avg\endcsname{41.18}
\expandafter\def\csname llmmvpval:f1.rest15.tasd.ft-llm\endcsname{72.41}
\expandafter\def\csname llmmvpval:f1.rest15.tasd.ft-mvp\endcsname{64.53}
\expandafter\def\csname llmmvpval:f1.rest15.tasd.ft-para\endcsname{63.06}
\expandafter\def\csname llmmvpval:f1.rest15.tasd.mvp.0\endcsname{40.88}
\expandafter\def\csname llmmvpval:f1.rest15.tasd.mvp.10\endcsname{63.29}
\expandafter\def\csname llmmvpval:f1.rest15.tasd.mvp.100\endcsname{70.50}
\expandafter\def\csname llmmvpval:f1.rest15.tasd.mvp.50\endcsname{68.18}
\expandafter\def\csname llmmvpval:f1.rest15.tasd.mvp.avg\endcsname{60.71}
\expandafter\def\csname llmmvpval:f1.rest15.tasd.mvpeff.0\endcsname{41.03}
\expandafter\def\csname llmmvpval:f1.rest15.tasd.mvpeff.10\endcsname{60.90}
\expandafter\def\csname llmmvpval:f1.rest15.tasd.mvpeff.100\endcsname{67.37}
\expandafter\def\csname llmmvpval:f1.rest15.tasd.mvpeff.50\endcsname{65.38}
\expandafter\def\csname llmmvpval:f1.rest15.tasd.nogd.0\endcsname{41.02}
\expandafter\def\csname llmmvpval:f1.rest15.tasd.nogd.10\endcsname{62.26}
\expandafter\def\csname llmmvpval:f1.rest15.tasd.nogd.100\endcsname{67.97}
\expandafter\def\csname llmmvpval:f1.rest15.tasd.nogd.50\endcsname{66.14}
\expandafter\def\csname llmmvpval:f1.rest15.tasd.reasoning.0\endcsname{48.18}
\expandafter\def\csname llmmvpval:f1.rest15.tasd.reasoning.10\endcsname{50.55}
\expandafter\def\csname llmmvpval:f1.rest15.tasd.reasoning.100\endcsname{57.08}
\expandafter\def\csname llmmvpval:f1.rest15.tasd.reasoning.50\endcsname{55.06}
\expandafter\def\csname llmmvpval:f1.rest15.tasd.reasoning.avg\endcsname{52.72}
\expandafter\def\csname llmmvpval:f1.rest15.tasd.sc.0\endcsname{42.14}
\expandafter\def\csname llmmvpval:f1.rest15.tasd.sc.10\endcsname{59.53}
\expandafter\def\csname llmmvpval:f1.rest15.tasd.sc.100\endcsname{66.14}
\expandafter\def\csname llmmvpval:f1.rest15.tasd.sc.50\endcsname{63.42}
\expandafter\def\csname llmmvpval:f1.rest15.tasd.sc.avg\endcsname{57.81}
\expandafter\def\csname llmmvpval:f1.rest15.tasd.so.0\endcsname{40.48}
\expandafter\def\csname llmmvpval:f1.rest15.tasd.so.10\endcsname{58.82}
\expandafter\def\csname llmmvpval:f1.rest15.tasd.so.100\endcsname{65.27}
\expandafter\def\csname llmmvpval:f1.rest15.tasd.so.50\endcsname{63.05}
\expandafter\def\csname llmmvpval:f1.rest15.tasd.so.avg\endcsname{56.91}
\expandafter\def\csname llmmvpval:f1.rest16.asqp.ft-llm\endcsname{66.49}
\expandafter\def\csname llmmvpval:f1.rest16.asqp.ft-mvp\endcsname{60.39}
\expandafter\def\csname llmmvpval:f1.rest16.asqp.ft-para\endcsname{57.93}
\expandafter\def\csname llmmvpval:f1.rest16.asqp.mvp.0\endcsname{39.15}
\expandafter\def\csname llmmvpval:f1.rest16.asqp.mvp.10\endcsname{52.46}
\expandafter\def\csname llmmvpval:f1.rest16.asqp.mvp.100\endcsname{60.61}
\expandafter\def\csname llmmvpval:f1.rest16.asqp.mvp.50\endcsname{58.08}
\expandafter\def\csname llmmvpval:f1.rest16.asqp.mvp.avg\endcsname{52.57}
\expandafter\def\csname llmmvpval:f1.rest16.asqp.mvpeff.0\endcsname{32.64}
\expandafter\def\csname llmmvpval:f1.rest16.asqp.mvpeff.10\endcsname{49.87}
\expandafter\def\csname llmmvpval:f1.rest16.asqp.mvpeff.100\endcsname{58.16}
\expandafter\def\csname llmmvpval:f1.rest16.asqp.mvpeff.50\endcsname{56.46}
\expandafter\def\csname llmmvpval:f1.rest16.asqp.nogd.0\endcsname{29.91}
\expandafter\def\csname llmmvpval:f1.rest16.asqp.nogd.10\endcsname{50.85}
\expandafter\def\csname llmmvpval:f1.rest16.asqp.nogd.100\endcsname{59.82}
\expandafter\def\csname llmmvpval:f1.rest16.asqp.nogd.50\endcsname{57.04}
\expandafter\def\csname llmmvpval:f1.rest16.asqp.reasoning.0\endcsname{38.27}
\expandafter\def\csname llmmvpval:f1.rest16.asqp.reasoning.10\endcsname{42.05}
\expandafter\def\csname llmmvpval:f1.rest16.asqp.reasoning.100\endcsname{46.62}
\expandafter\def\csname llmmvpval:f1.rest16.asqp.reasoning.50\endcsname{44.67}
\expandafter\def\csname llmmvpval:f1.rest16.asqp.reasoning.avg\endcsname{42.90}
\expandafter\def\csname llmmvpval:f1.rest16.asqp.sc.0\endcsname{34.52}
\expandafter\def\csname llmmvpval:f1.rest16.asqp.sc.10\endcsname{50.52}
\expandafter\def\csname llmmvpval:f1.rest16.asqp.sc.100\endcsname{58.31}
\expandafter\def\csname llmmvpval:f1.rest16.asqp.sc.50\endcsname{56.38}
\expandafter\def\csname llmmvpval:f1.rest16.asqp.sc.avg\endcsname{49.93}
\expandafter\def\csname llmmvpval:f1.rest16.asqp.so.0\endcsname{30.13}
\expandafter\def\csname llmmvpval:f1.rest16.asqp.so.10\endcsname{48.98}
\expandafter\def\csname llmmvpval:f1.rest16.asqp.so.100\endcsname{56.36}
\expandafter\def\csname llmmvpval:f1.rest16.asqp.so.50\endcsname{54.70}
\expandafter\def\csname llmmvpval:f1.rest16.asqp.so.avg\endcsname{47.54}
\expandafter\def\csname llmmvpval:f1.rest16.tasd.ft-llm\endcsname{79.78}
\expandafter\def\csname llmmvpval:f1.rest16.tasd.ft-mvp\endcsname{72.76}
\expandafter\def\csname llmmvpval:f1.rest16.tasd.ft-para\endcsname{71.97}
\expandafter\def\csname llmmvpval:f1.rest16.tasd.mvp.0\endcsname{51.03}
\expandafter\def\csname llmmvpval:f1.rest16.tasd.mvp.10\endcsname{67.79}
\expandafter\def\csname llmmvpval:f1.rest16.tasd.mvp.100\endcsname{74.95}
\expandafter\def\csname llmmvpval:f1.rest16.tasd.mvp.50\endcsname{73.19}
\expandafter\def\csname llmmvpval:f1.rest16.tasd.mvp.avg\endcsname{66.74}
\expandafter\def\csname llmmvpval:f1.rest16.tasd.mvpeff.0\endcsname{50.52}
\expandafter\def\csname llmmvpval:f1.rest16.tasd.mvpeff.10\endcsname{65.46}
\expandafter\def\csname llmmvpval:f1.rest16.tasd.mvpeff.100\endcsname{72.96}
\expandafter\def\csname llmmvpval:f1.rest16.tasd.mvpeff.50\endcsname{71.45}
\expandafter\def\csname llmmvpval:f1.rest16.tasd.nogd.0\endcsname{50.30}
\expandafter\def\csname llmmvpval:f1.rest16.tasd.nogd.10\endcsname{67.19}
\expandafter\def\csname llmmvpval:f1.rest16.tasd.nogd.100\endcsname{73.70}
\expandafter\def\csname llmmvpval:f1.rest16.tasd.nogd.50\endcsname{71.35}
\expandafter\def\csname llmmvpval:f1.rest16.tasd.reasoning.0\endcsname{54.59}
\expandafter\def\csname llmmvpval:f1.rest16.tasd.reasoning.10\endcsname{55.97}
\expandafter\def\csname llmmvpval:f1.rest16.tasd.reasoning.100\endcsname{60.00}
\expandafter\def\csname llmmvpval:f1.rest16.tasd.reasoning.50\endcsname{59.80}
\expandafter\def\csname llmmvpval:f1.rest16.tasd.reasoning.avg\endcsname{57.59}
\expandafter\def\csname llmmvpval:f1.rest16.tasd.sc.0\endcsname{50.98}
\expandafter\def\csname llmmvpval:f1.rest16.tasd.sc.10\endcsname{64.68}
\expandafter\def\csname llmmvpval:f1.rest16.tasd.sc.100\endcsname{71.89}
\expandafter\def\csname llmmvpval:f1.rest16.tasd.sc.50\endcsname{69.81}
\expandafter\def\csname llmmvpval:f1.rest16.tasd.sc.avg\endcsname{64.34}
\expandafter\def\csname llmmvpval:f1.rest16.tasd.so.0\endcsname{50.10}
\expandafter\def\csname llmmvpval:f1.rest16.tasd.so.10\endcsname{64.30}
\expandafter\def\csname llmmvpval:f1.rest16.tasd.so.100\endcsname{70.74}
\expandafter\def\csname llmmvpval:f1.rest16.tasd.so.50\endcsname{69.30}
\expandafter\def\csname llmmvpval:f1.rest16.tasd.so.avg\endcsname{63.61}
\expandafter\def\csname llmmvpval:gain.mvp.10to100.max\endcsname{8.15}
\expandafter\def\csname llmmvpval:gain.mvp.10to100.max.int\endcsname{8} %
\expandafter\def\csname llmmvpval:gap.ft.max\endcsname{7.1} %
\expandafter\def\csname llmmvpval:gap.ft.max.int\endcsname{7}
\expandafter\def\csname llmmvpval:gap.ft.min\endcsname{1.9} %
\expandafter\def\csname llmmvpval:gap.ft.min.int\endcsname{2}
\expandafter\def\csname llmmvpval:intro.ft.mvp\endcsname{51.04}
\expandafter\def\csname llmmvpval:intro.gap\endcsname{9}
\expandafter\def\csname llmmvpval:intro.hellwig.best\endcsname{41.74} %
\expandafter\def\csname llmmvpval:intro.hellwig.best.shots\endcsname{50}
\expandafter\def\csname llmmvpval:mvar.asqp.best.avg\endcsname{44.36}
\expandafter\def\csname llmmvpval:mvar.asqp.best.m\endcsname{17}
\expandafter\def\csname llmmvpval:mvar.asqp.m1.avg\endcsname{41.54}
\expandafter\def\csname llmmvpval:mvar.asqp.m11.avg\endcsname{43.89}
\expandafter\def\csname llmmvpval:mvar.asqp.m13.avg\endcsname{44.12}
\expandafter\def\csname llmmvpval:mvar.asqp.m15.avg\endcsname{44.27}
\expandafter\def\csname llmmvpval:mvar.asqp.m17.avg\endcsname{44.36}
\expandafter\def\csname llmmvpval:mvar.asqp.m19.avg\endcsname{44.33}
\expandafter\def\csname llmmvpval:mvar.asqp.m21.avg\endcsname{44.27}
\expandafter\def\csname llmmvpval:mvar.asqp.m23.avg\endcsname{44.33}
\expandafter\def\csname llmmvpval:mvar.asqp.m24.avg\endcsname{44.29}
\expandafter\def\csname llmmvpval:mvar.asqp.m3.avg\endcsname{42.80}
\expandafter\def\csname llmmvpval:mvar.asqp.m5.avg\endcsname{43.22}
\expandafter\def\csname llmmvpval:mvar.asqp.m7.avg\endcsname{43.51}
\expandafter\def\csname llmmvpval:mvar.asqp.m9.avg\endcsname{43.72}
\expandafter\def\csname llmmvpval:mvar.tasd.best.avg\endcsname{57.30}
\expandafter\def\csname llmmvpval:mvar.tasd.best.m\endcsname{5}
\expandafter\def\csname llmmvpval:mvar.tasd.m1.avg\endcsname{56.32}
\expandafter\def\csname llmmvpval:mvar.tasd.m3.avg\endcsname{57.11}
\expandafter\def\csname llmmvpval:mvar.tasd.m5.avg\endcsname{57.30}
\expandafter\def\csname llmmvpval:mvar.tasd.m6.avg\endcsname{57.07}
\expandafter\def\csname llmmvpval:p.energy.asqp.ge10.mode\endcsname{0.524} %
\expandafter\def\csname llmmvpval:p.energy.tasd.0.sonc-mvp\endcsname{0.445}
\expandafter\def\csname llmmvpval:p.energy.tasd.100.reasoning-ft\endcsname{0.036}
\expandafter\def\csname llmmvpval:p.energy.tasd.100.so-ft\endcsname{0.426}
\expandafter\def\csname llmmvpval:perm.asqp.beats.oracle\endcsname{14} %
\expandafter\def\csname llmmvpval:perm.asqp.mvp.avg\endcsname{43.22} %
\expandafter\def\csname llmmvpval:perm.asqp.oracle.avg\endcsname{42.71} %
\expandafter\def\csname llmmvpval:perm.few.gap.max\endcsname{0.74} %
\expandafter\def\csname llmmvpval:perm.tasd.beats.oracle\endcsname{12} %
\expandafter\def\csname llmmvpval:perm.tasd.mvp.avg\endcsname{57.30} %
\expandafter\def\csname llmmvpval:perm.tasd.oracle.avg\endcsname{57.20} %
\expandafter\def\csname llmmvpval:prompts.total\endcsname{7,448,672} %
\expandafter\def\csname llmmvpval:sig.n.comparisons\endcsname{84}
\expandafter\def\csname llmmvpval:sig.n.pairs\endcsname{21}
\expandafter\def\csname llmmvpval:time.asqp.ft\endcsname{0.103}
\expandafter\def\csname llmmvpval:time.asqp.mvp.0\endcsname{1.319}
\expandafter\def\csname llmmvpval:time.asqp.mvp.10\endcsname{1.593}
\expandafter\def\csname llmmvpval:time.asqp.mvp.100\endcsname{2.121}
\expandafter\def\csname llmmvpval:time.asqp.mvp.50\endcsname{1.850}
\expandafter\def\csname llmmvpval:time.asqp.mvp.max\endcsname{2.1}
\expandafter\def\csname llmmvpval:time.asqp.mvp.min\endcsname{1.3}
\expandafter\def\csname llmmvpval:time.asqp.reasoning.0\endcsname{0.086}
\expandafter\def\csname llmmvpval:time.asqp.reasoning.10\endcsname{0.099}
\expandafter\def\csname llmmvpval:time.asqp.reasoning.100\endcsname{0.202}
\expandafter\def\csname llmmvpval:time.asqp.reasoning.50\endcsname{0.134}
\expandafter\def\csname llmmvpval:time.asqp.sc-nc.0\endcsname{1.495}
\expandafter\def\csname llmmvpval:time.asqp.sc-nc.10\endcsname{2.641}
\expandafter\def\csname llmmvpval:time.asqp.sc-nc.100\endcsname{15.715}
\expandafter\def\csname llmmvpval:time.asqp.sc-nc.50\endcsname{8.183}
\expandafter\def\csname llmmvpval:time.asqp.sc.0\endcsname{0.209}
\expandafter\def\csname llmmvpval:time.asqp.sc.10\endcsname{0.201}
\expandafter\def\csname llmmvpval:time.asqp.sc.100\endcsname{0.243}
\expandafter\def\csname llmmvpval:time.asqp.sc.50\endcsname{0.228}
\expandafter\def\csname llmmvpval:time.asqp.so-nc.0\endcsname{0.302}
\expandafter\def\csname llmmvpval:time.asqp.so-nc.10\endcsname{0.529}
\expandafter\def\csname llmmvpval:time.asqp.so-nc.100\endcsname{3.099}
\expandafter\def\csname llmmvpval:time.asqp.so-nc.50\endcsname{1.623}
\expandafter\def\csname llmmvpval:time.asqp.so.0\endcsname{0.056}
\expandafter\def\csname llmmvpval:time.asqp.so.10\endcsname{0.064}
\expandafter\def\csname llmmvpval:time.asqp.so.100\endcsname{0.085}
\expandafter\def\csname llmmvpval:time.asqp.so.50\endcsname{0.074}
\expandafter\def\csname llmmvpval:time.tasd.ft\endcsname{0.083}
\expandafter\def\csname llmmvpval:time.tasd.mvp.0\endcsname{0.299}
\expandafter\def\csname llmmvpval:time.tasd.mvp.10\endcsname{0.259}
\expandafter\def\csname llmmvpval:time.tasd.mvp.100\endcsname{0.388}
\expandafter\def\csname llmmvpval:time.tasd.mvp.50\endcsname{0.325}
\expandafter\def\csname llmmvpval:time.tasd.mvp.max.ceil\endcsname{0.4} %
\expandafter\def\csname llmmvpval:time.tasd.reasoning.0\endcsname{0.062}
\expandafter\def\csname llmmvpval:time.tasd.reasoning.10\endcsname{0.080}
\expandafter\def\csname llmmvpval:time.tasd.reasoning.100\endcsname{0.141}
\expandafter\def\csname llmmvpval:time.tasd.reasoning.50\endcsname{0.103}
\expandafter\def\csname llmmvpval:time.tasd.sc-nc.0\endcsname{1.240}
\expandafter\def\csname llmmvpval:time.tasd.sc-nc.10\endcsname{2.198}
\expandafter\def\csname llmmvpval:time.tasd.sc-nc.100\endcsname{12.644}
\expandafter\def\csname llmmvpval:time.tasd.sc-nc.50\endcsname{6.869}
\expandafter\def\csname llmmvpval:time.tasd.sc.0\endcsname{0.162}
\expandafter\def\csname llmmvpval:time.tasd.sc.10\endcsname{0.144}
\expandafter\def\csname llmmvpval:time.tasd.sc.100\endcsname{0.174}
\expandafter\def\csname llmmvpval:time.tasd.sc.50\endcsname{0.163}
\expandafter\def\csname llmmvpval:time.tasd.so-nc.0\endcsname{0.249}
\expandafter\def\csname llmmvpval:time.tasd.so-nc.10\endcsname{0.438}
\expandafter\def\csname llmmvpval:time.tasd.so-nc.100\endcsname{2.503}
\expandafter\def\csname llmmvpval:time.tasd.so-nc.50\endcsname{1.360}
\expandafter\def\csname llmmvpval:time.tasd.so.0\endcsname{0.047}
\expandafter\def\csname llmmvpval:time.tasd.so.10\endcsname{0.043}
\expandafter\def\csname llmmvpval:time.tasd.so.100\endcsname{0.062}
\expandafter\def\csname llmmvpval:time.tasd.so.50\endcsname{0.057}

\begin{document}
\maketitle
\begin{abstract}
Recent work explored the capabilities of Large Language Models (LLMs) in Aspect-Based Sentiment Analysis (ABSA) through few-shot prompting, requiring substantially fewer annotated examples while achieving notable improvements over zero-shot baselines. However, a performance gap remained compared to models fine-tuned on about a thousand examples, and the computational costs of LLM inference present practical barriers to deployment. We introduce LLM-based Multi-view Prompting (LLM-MvP), which adapts the multi-view principle of considering multiple element orderings to LLM prompting. By combining schema-constrained decoding with prefix caching, LLM-MvP substantially reduces the computational overhead of processing multiple views. Experiments on five benchmark datasets show that LLM-MvP narrows the gap between few-shot prompting and fine-tuned models, surpassing fine-tuned T5-based models in several settings.
\end{abstract}
\section{Introduction}

Aspect-Based Sentiment Analysis (ABSA) extends conventional sentiment analysis by identifying fine-grained opinions towards specific aspects within a text, enabling a more nuanced understanding of expressed sentiment than document- or sentence-level approaches \citep{zhang2022survey}. ABSA tasks extract up to four sentiment elements per aspect: aspect term (\textit{a}), the specific entity discussed; aspect category (\textit{c}), its broader semantic class; opinion term (\textit{o}), the sentiment-bearing phrase; and polarity (\textit{p}), indicating positive, negative, or neutral orientation \citep{zhang2022survey} (see Appendix~\ref{appendix:tuples-examples} for annotated examples). Among various subtasks, Aspect Sentiment Quad Prediction (ASQP) is the most challenging, requiring extraction of all four sentiment elements \citep{zhang2021aspect}.

While recent few-shot prompting approaches have narrowed the gap to fine-tuned models \citep{hellwig2025still, ventirozos-etal-2025-aspect}, a performance gap remains, particularly for the more challenging ASQP task. On the SemEval-2015 restaurant dataset, the best previously reported few-shot prompting score, 50.7 (Gemini 2.0 Flash, 100-shot) \citep{ventirozos-etal-2025-aspect}, still trails the fine-tuned MvP baseline of \val{intro.ft.mvp} \citep{gou2023mvp}. Prompting Large Language Models (LLMs) for ABSA also introduces several practical challenges. First, the computational cost and energy consumption of LLM inference can be prohibitive. Second, LLMs frequently generate invalid outputs, such as predicting aspect or opinion terms that do not appear in the input text, or producing categories and polarities outside the predefined set. These invalid generations have previously been tackled by validation and regeneration, further increasing computational overhead \citep{hellwig2025still}.

To address the above challenges, we propose \textit{\textbf{LLM}-based \textbf{M}ulti-\textbf{v}iew \textbf{P}rompting} (LLM-MvP). LLM-MvP considers multiple permutations of sentiment elements and selects outputs based on model confidence measured by token-level entropy. Furthermore, we address the challenge of invalid outputs through schema-constrained decoding using context-free grammars, and mitigate computational overhead via prefix caching.

We evaluate LLM-MvP on both a quadruple extraction task, ASQP, and a triplet extraction task, Target Aspect Sentiment Detection (TASD), across five diverse datasets spanning the restaurant, airline, e-learning, and hotel domains. In particular, we compare LLM-MvP against several commonly used prompting strategies, including standard single-order prompting \citep{hellwig2025still}, self-consistency prompting \citep{hellwig2025still, gou2023mvp}, and reasoning-augmented prompting \citep{10.5555/3600270.3602070}, as well as against state-of-the-art (SOTA) fine-tuned models. Our investigation is guided by the following research questions:

\begin{itemize}
    \item \textbf{RQ1:} Does LLM-MvP yield performance improvements compared to (1) single-order prompting with a single LLM execution per example, (2) self-consistency prompting, and (3) reasoning-augmented prompting?
    \item \textbf{RQ2:} Does LLM-MvP surpass the performance of fine-tuned models?
    \item \textbf{RQ3:} To what extent does prefix caching mitigate the computational overhead associated with processing multiple permutations per input instance?
\end{itemize}

Our main contributions are as follows:

\begin{itemize}
    \item We demonstrate that LLM-MvP achieves performance competitive with or superior to fine-tuned models while relying on only a few demonstrations.
    \item We release all code and results on GitHub\footnote{\ghicon\url{https://github.com/NilsHellwig/llmmvp}}, enabling adaptation to other domains and languages.
\end{itemize}

\section{Related Work}

The highest performance scores across ABSA tasks have been achieved using generative models fine-tuned on human-annotated examples \citep{zhang2022survey, vsmid2024llama, fehle2026leveraging}. Prior work has predominantly employed Small Language Models (SLMs), specifically Google's T5 base (220M) encoder-decoder model \citep{raffel2020exploring}, with individual approaches primarily distinguished by their representation of output tuples. A fundamental limitation of these approaches, however, is their reliance on a fixed left-to-right generation order, which fails to account for the interdependence of sentiment elements, is sensitive to the choice of ordering \citep{hu-etal-2022-improving-aspect}, and introduces error accumulation across sequentially predicted elements.

\citet{gou2023mvp} addressed these limitations by considering multiple orderings of sentiment elements within tuples, aggregating predictions across views via majority voting. Specifically, the top-$m$ orderings are selected by scoring the gold-annotated target sequences of the training set under each ordering with a pre-trained T5 model and ranking the orderings by their average conditional generation probability, thereby identifying the orderings of highest generation confidence. The model is then fine-tuned on all $m$ selected orderings per training example, and at inference, predictions from all $m$ orderings are aggregated via majority voting, retaining only tuples that appear in at least half of the generated outputs. Building on this principle, \citet{jun-lee-2025-dynamic} observed that applying the same number of views to every instance is unnecessary, and proposed a dynamic order template method that uses instance-level entropy to predict the required number of views per example from labeled data, reducing inference overhead while preserving the benefits of multi-view ensembling. Beyond T5-based approaches, the strongest results were reported for LLMs fine-tuned via Low-Rank Adaptation (LoRA)~\citep{hu2022lora}, a parameter-efficient fine-tuning technique that constrains weight updates to low-rank matrices~\citep{vsmid2024llama, fehle2026leveraging}.

Since training task-specific models requires substantial amounts of human-annotated data, there has been growing interest in tackling NLP tasks in settings where no or only a few annotated examples are available \citep{lepagnol2024small, munker2025zero}, both in sentiment analysis \citep{zhang2024sentiment, hasan2024zero, nevsic2024advancing} and in ABSA \citep{hellwig2025still, bai2024compound, gou2023mvp}. In the field of ABSA, research has primarily focused on zero-shot and few-shot prompting. \citet{hellwig2025still} reported performance scores for 50-shot prompting that approached those achieved by fine-tuned methods such as MvP \citep{gou2023mvp}. For instance, they achieved an F1 score of 62.12 compared to 64.53 for MvP on the restaurant-domain dataset from SemEval-2015 \citep{pontiki2015semeval} for the TASD task, which considers all sentiment elements except the opinion term \citep{zhang2021aspect}. Furthermore, \citet{hellwig2025still} demonstrated that self-consistency prompting, which executes the prompt multiple times with different random seeds and aggregates the results via majority voting, can improve performance across tasks and shot counts.
\begin{figure*}[t]
    \centering
    \includegraphics[width=\textwidth]{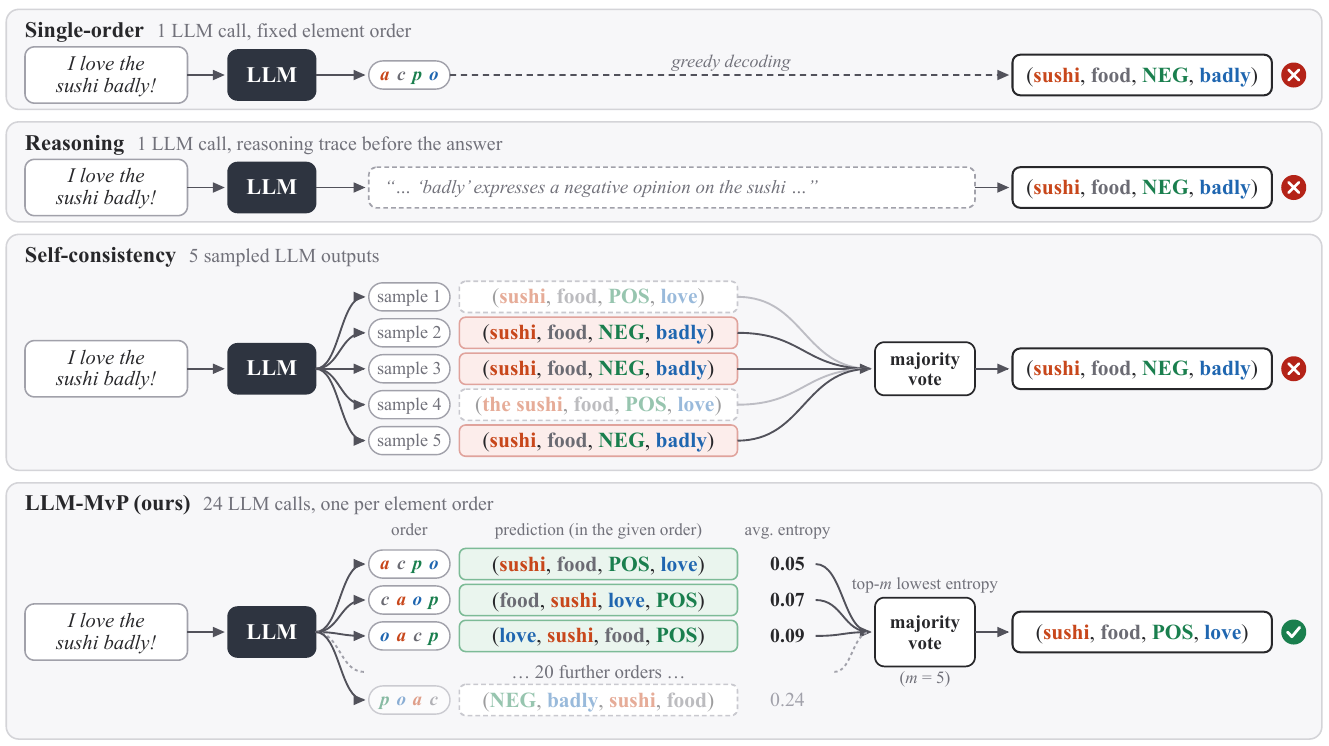}
    \caption{\textbf{Comparison of LLM-MvP against other prompting approaches.} For each instance, LLM-MvP queries the LLM across all permutations of the sentiment elements in a given tuple, e.g., resulting in 24 distinct orderings and thus 24 LLM executions for the illustrated ASQP task. The five outputs with the lowest average token-level entropy are selected and combined via majority voting. In comparison, the self-consistency mechanism proposed by \citet{hellwig2025still} samples five generations with varying seeds, followed by majority voting over the five outputs.}
    \label{fig:main-approach}
\end{figure*}

\section{System Description}

The LLM-MvP pipeline is illustrated in Figure~\ref{fig:main-approach}. In contrast to the single-order prediction employed by previous approaches \citep{hellwig2025still, bai2024compound} and to self-consistency, we consider all possible permutations of sentiment elements (i.e., 6 for triplets and 24 for quadruplets). In the following, we describe each component in detail.

\subsection{Prompt Formulation}

\begin{figure*}[t]
    \centering
    \includegraphics[width=\textwidth]{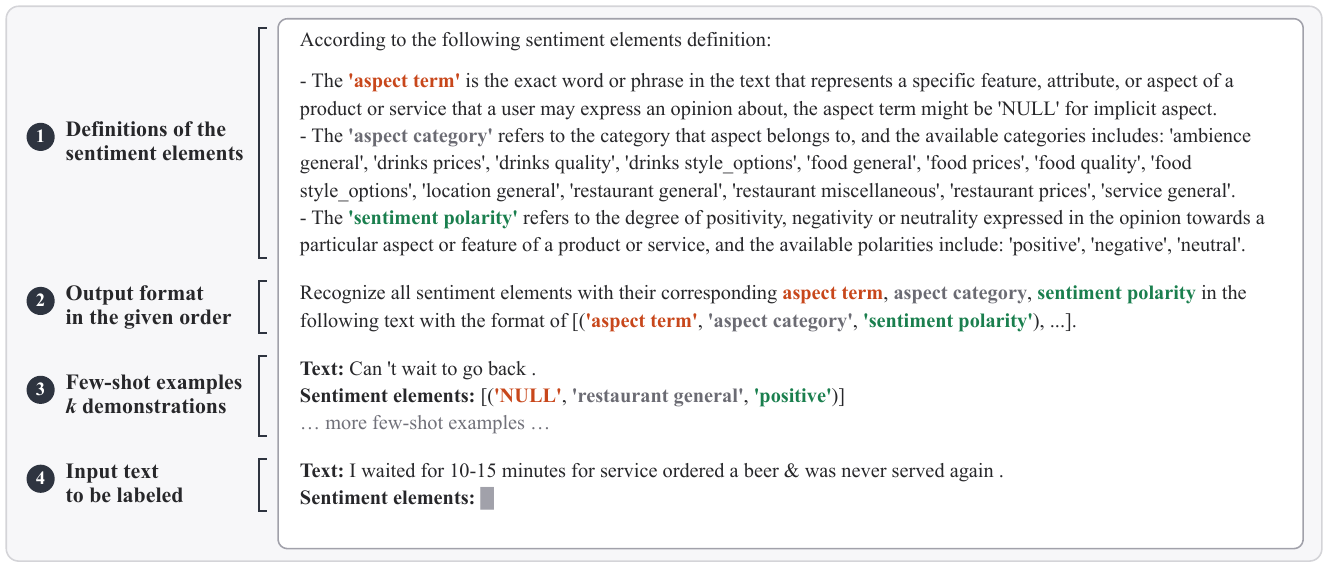}
    \caption{\textbf{Example prompt used for ABSA prompting.} For ASQP, the same prompt is extended with a fourth bullet point that defines opinion terms, while the few-shot examples include opinion term annotations.}
    \label{fig:prompt}
\end{figure*}

Following \citet{hellwig2025still}, \citet{bai2024compound}, and \citet{gou2023mvp}, the prompt comprises four components: (1) descriptions of the sentiment elements under consideration (e.g., the definition of an aspect term), (2) output format specifications, (3) annotated examples in the case of few-shot prompting, and (4) the input text for which ABSA is to be performed. All few-shot examples include both the text and labels, where the sentiment elements within the label tuples are ordered according to the desired output format. The same fixed set of $k$ random few-shot examples is included in the prompts used to predict the aspect tuples of all examples in a given test set. An example is provided in Figure~\ref{fig:prompt}.

\subsection{Multi-view Inference}

While \citet{gou2023mvp} originally adapted the multi-view principle for fine-tuning on about a thousand labeled examples, using a gold-annotated training set to identify the $m$ permutations with the lowest average entropy on a pre-trained T5 model, our setting differs fundamentally. Since we assume neither a training nor a development set to be available, the selection of top-$m$ permutations cannot rely on supervised signals. Consequently, the multi-view principle must be reformulated for zero- and few-shot prompting.

LLM-MvP generates predictions for each input text across all possible permutations of sentiment elements. For each LLM output, we compute the entropy at each generated token position and subsequently average it over all token positions, yielding a mean entropy score. Lower entropy corresponds to higher model confidence.

Let $y = (y_1, y_2, \ldots, y_T)$ denote the generated output sequence of length $T$ for a given permutation. The entropy at each token position $t$ is defined as:

\begin{equation}
    H(y_t) = -\sum_{v \in \mathcal{V}} P(y_t = v) \log P(y_t = v)
\end{equation}

where $\mathcal{V}$ represents the vocabulary and $P(y_t = v)$ is the probability assigned to token $v$ at position $t$. The average entropy for output $y$ is computed as:
\begin{equation}
    \bar{H}(y) = \frac{1}{T} \sum_{t=1}^{T} H(y_t)
\end{equation}

We average over all generated tokens, including structural ones such as quotation marks, commas, and brackets: under schema-constrained decoding (Section~\ref{sec:constrained-decoding}), these positions rarely admit only a single token but encode whether an aspect or opinion term ends and whether a further tuple follows.

For a given example, the $m$ permutations with the lowest average entropy are selected for aggregation via majority voting. 

\subsection{Multi-view Results Aggregation}

For an input sentence $x$, let $T'_{p_i}$ denote the set of predicted tuples for permutation $p_i$. The final aggregated result $T'_{\text{MvP}}$ is obtained by:

\begin{equation}
    T'_{\text{MvP}} = \left\{t \mid t \in \bigcup_{i=1}^{m} T'_{p_i} \text{ and } \sum_{i=1}^{m} 1_{T'_{p_i}}(t) > \frac{m}{2}\right\}
\end{equation}

where $1_{T'_{p_i}}(t)$ is an indicator function that equals 1 if tuple $t$ appears in $T'_{p_i}$ and 0 otherwise.

\subsection{Schema-Constrained Generation with Batch Inference}

The computational overhead of LLM-MvP is substantial, as the total number of prompts to be processed equals the number of test instances multiplied by the number of possible permutations. To efficiently handle this large volume of inference requests while ensuring valid output structures, LLM-MvP employs two key techniques: batched inference and schema-constrained decoding.

\subsubsection{Efficient Batched Inference}
To manage the substantial computational demands of processing all permuted prompts, we use vLLM (v0.20.0) \citep{kwon2023efficient} with prefix caching. Since ABSA operates on sentence-level inputs, unlike many NLP tasks that process longer documents, all test instances within a given permutation share an identical prompt prefix, comprising the task description, the output format specification, and the few-shot examples, and differ only in the input sentence and the trailing \texttt{Sentiment elements:} tokens. The prefix key-value cache is thus computed once per permutation and reused across all instances, avoiding redundant prefill computation. Since only the short instance-specific suffix has to be prefilled and the shared prefix blocks are stored only once in GPU memory, vLLM can additionally schedule many more requests concurrently within a single batch.

\subsubsection{Schema-Constrained Decoding}\label{sec:constrained-decoding}
LLM-MvP employs schema-constrained decoding, also referred to as guided decoding (GD), based on context-free grammars (CFGs), implemented via XGrammar \citep{dong2024xgrammar} in vLLM, to guarantee syntactically valid sentiment tuples. The valid phrase set $\mathcal{P}(x)$ is constructed by splitting the input sentence $x$ at whitespace, punctuation, hyphen, camel-case, and non-ASCII character boundaries, then enumerating all contiguous substring spans over the resulting tokens. Matching is case-sensitive. Since $\mathcal{P}(x)$ differs between inputs, a separate grammar is compiled for each input sentence and passed to vLLM as a per-request structured-output constraint, so that each request within a batch is decoded under its own grammar. The CFG enforces that (1) aspect and opinion terms are restricted to $\mathcal{P}(x)$, (2) polarities are drawn from $\{\textit{positive},\textit{negative},\textit{neutral}\}$, and (3) aspect categories belong to the dataset-specific set $\mathcal{C}$:

\begin{align*}
\text{OUTPUT} &\;\longrightarrow\; \texttt{[}\;\text{TUPLE}\;(\texttt{,}\;\text{TUPLE})^{*}\;\texttt{]}              \\[2pt]
\text{TUPLE}  &\;\longrightarrow\; \texttt{(}\;e_{\pi(1)},\,e_{\pi(2)},\,\ldots,\,e_{\pi(k)}\;\texttt{)}           \\[2pt]
e_\text{at}   &\;\longrightarrow\; p \in \mathcal{P}(x) \;\mid\; \texttt{NULL}                                      \\
e_\text{ot}   &\;\longrightarrow\; p \in \mathcal{P}(x)                                                             \\
e_\text{ac}   &\;\longrightarrow\; c \in \mathcal{C}                                                                 \\
e_\text{p}    &\;\longrightarrow\; \textit{positive} \;\mid\; \textit{negative} \;\mid\; \textit{neutral}
\end{align*}

\noindent where $\pi$ denotes the element ordering of the given permutation. Ensuring syntactically valid tuples eliminates the need for post-hoc validation and regeneration employed by \citet{hellwig2025still}, which necessitated additional LLM executions and thus increased inference time and energy consumption.
\section{Experiments}

\subsection{Tasks and Datasets}

We evaluated LLM-MvP on ASQP and TASD. We included the restaurant-domain corpora from SemEval-2015 (Rest15) \citep{pontiki2015semeval} and SemEval-2016 (Rest16) \citep{pontiki2016semeval}. The originally released versions of these datasets do not provide annotations for opinion terms, which are required for ASQP. Hence, for the ASQP task, we employed the extended labels of \citet{zhang2021aspect}, which include opinion terms.

We also included three additional datasets, following the evaluation setup of \citet{hellwig2025still}: FlightABSA \citep{hellwig2025still}, a corpus of airline reviews sourced from TripAdvisor; OATS Coursera \citep{chebolu2024oats}, a dataset derived from e-learning courses on Coursera; and OATS Hotels \citep{chebolu2024oats}, a collection of hotel reviews from TripAdvisor (hereafter Coursera and Hotels). This extended set of five English-language datasets allows us to assess the generalizability of LLM-MvP across a broader range of domains. Appendix~\ref{appendix:datasets} summarizes the number of training and test instances across all datasets, while Appendix~\ref{sec:distributions} provides a detailed distributional analysis of aspect categories and sentiment polarities.

\subsection{Implementation Details}

For our experiments, we employed Google's open-source model Gemma 4\footnote{\url{https://ai.google.dev/gemma/docs/core/model_card_4}} with 31 billion parameters, 4-bit quantized, which requires approximately 17.4\,GB of VRAM for inference\footnote{\url{https://ai.google.dev/gemma/docs/core}}, making it compatible with consumer-grade GPUs. Generation stopped once the model produced the sequence \texttt{")]"}, which ends the output label. We set the temperature to~0 and the maximum context size to 16,384 tokens, which accommodates all shot settings.

We evaluated LLM-MvP across the few-shot settings commonly employed in prior work \citep{hellwig2025still, ventirozos-etal-2025-aspect, bai2024compound}: 0-shot, 10-shot, 50-shot, and 100-shot. To account for the choice of few-shot examples, each configuration was repeated five times with independently sampled sets of few-shot examples, and all reported scores are the mean over these five runs. We further adopted a cumulative sampling strategy in which examples from lower-shot settings were retained in higher-shot configurations. For instance, the 10 examples used in the 10-shot condition were preserved in the 50-shot condition, with 40 additional examples appended, isolating the effect of demonstration count from that of example selection.

Following \citet{gou2023mvp}, we applied majority voting over the $m$ lowest-entropy outputs from the set of all possible permutations $P$. We fixed $m=5$ as a single default ensemble size for majority voting across both tasks, matching the five outputs aggregated by the SC baseline of \citet{hellwig2025still}. All experimental conditions of our study, including energy efficiency measurements, were executed on an NVIDIA RTX PRO 6000 Blackwell Generation GPU with 96\,GB VRAM.

\subsection{Evaluation Metrics}

Following standard practice in ABSA research, micro-averaged F1 served as the primary evaluation metric \citep{zhang2022survey}, counting a tuple as correct only if all sentiment elements exactly matched the gold annotation: $F_1 = 2\,\mathrm{TP} / (2\,\mathrm{TP} + \mathrm{FP} + \mathrm{FN})$, where $\mathrm{TP}$ is the number of predicted tuples that match a gold tuple, $\mathrm{FP}$ the number of predicted tuples without a match, and $\mathrm{FN}$ the number of gold tuples that were not predicted, each counted over the entire test set. Precision, recall, and macro-averaged F1 scores are reported in Appendix~\ref{sec:task-performance} for all conditions. All predictions are publicly available in our GitHub repository, allowing the calculation of further metrics.

\subsection{Baseline Approaches}

\begin{table*}[t]
\centering
\small
\setlength{\tabcolsep}{1.0pt}
\renewcommand{\arraystretch}{0.8} 
\resizebox{2.0\columnwidth}{!}{%

\begin{tabular}{@{}ll @{\hskip 6pt} rrrr @{\hskip 3pt} r @{\hskip 3pt} rrr @{\hspace{10pt}} rrrr @{\hskip 3pt} r @{\hskip 3pt} rrr@{}}
\toprule
\multirow{3}{*}{\textbf{Dataset}} & \multirow{3}{*}{\textbf{Approach}} & \multicolumn{8}{c}{\textbf{TASD} ($F_1$)} & \multicolumn{8}{c}{\textbf{ASQP} ($F_1$)} \\
\cmidrule(lr){3-10}\cmidrule(lr){11-18}
& & \multicolumn{4}{c}{\textit{\# Shots}} & \multirow{2}{*}{\textit{Avg.}} & \multicolumn{3}{c}{\textit{FT Baselines}} & \multicolumn{4}{c}{\textit{\# Shots}} & \multirow{2}{*}{\textit{Avg.}} & \multicolumn{3}{c}{\textit{FT Baselines}} \\
\cmidrule(lr){3-6}\cmidrule(lr){8-10}\cmidrule(lr){11-14}\cmidrule(lr){16-18}
& & \textbf{0} & \textbf{10} & \textbf{50} & \textbf{100} & & Para.$^{\star}$ & MvP$^{\star}$ & LLM$^{\star}$ & \textbf{0} & \textbf{10} & \textbf{50} & \textbf{100} & & Para.$^{\star}$ & MvP$^{\star}$ & LLM$^{\star}$ \\
\midrule
\multirow{4}{*}{\textsc{Rest15}} & Single-order & 40.48 & 58.82 & 63.05 & 65.27 & 56.91 & \multicolumn{1}{c}{\multirow{4}{*}{63.06}} & \multicolumn{1}{c}{\multirow{4}{*}{64.53}} & \multicolumn{1}{c}{\multirow{4}{*}{\underline{\textbf{72.41}}}} & 25.59 & 40.43 & 47.99 & 50.70 & 41.18 & \multicolumn{1}{c}{\multirow{4}{*}{46.93}} & \multicolumn{1}{c}{\multirow{4}{*}{51.04}} & \multicolumn{1}{c}{\multirow{4}{*}{\underline{\textbf{56.78}}}} \\
& Reasoning & \textbf{48.18} & 50.55 & 55.06 & 57.08 & 52.72 & & & & \textbf{33.33} & 36.39 & 37.51 & 39.95 & 36.79 & & & \\
& Self-consistency & 42.14 & 59.53 & 63.42 & 66.14 & 57.81 & & & & 27.10 & 41.99 & 49.27 & 51.74 & 42.53 & & & \\
& LLM-MvP (ours) & 40.88 & \textbf{63.29} & \textbf{68.18} & \textbf{70.50} & \textbf{60.71} & & & & 32.88 & \textbf{45.56} & \textbf{50.99} & \textbf{53.40} & \textbf{45.71} & & & \\
\midrule
\multirow{4}{*}{\textsc{Rest16}} & Single-order & 50.10 & 64.30 & 69.30 & 70.74 & 63.61 & \multicolumn{1}{c}{\multirow{4}{*}{71.97}} & \multicolumn{1}{c}{\multirow{4}{*}{72.76}} & \multicolumn{1}{c}{\multirow{4}{*}{\underline{\textbf{79.78}}}} & 30.13 & 48.98 & 54.70 & 56.36 & 47.54 & \multicolumn{1}{c}{\multirow{4}{*}{57.93}} & \multicolumn{1}{c}{\multirow{4}{*}{60.39}} & \multicolumn{1}{c}{\multirow{4}{*}{\underline{\textbf{66.49}}}} \\
& Reasoning & \textbf{54.59} & 55.97 & 59.80 & 60.00 & 57.59 & & & & 38.27 & 42.05 & 44.67 & 46.62 & 42.90 & & & \\
& Self-consistency & 50.98 & 64.68 & 69.81 & 71.89 & 64.34 & & & & 34.52 & 50.52 & 56.38 & 58.31 & 49.93 & & & \\
& LLM-MvP (ours) & 51.03 & \textbf{67.79} & \textbf{73.19} & \textbf{74.95} & \textbf{66.74} & & & & \textbf{39.15} & \textbf{52.46} & \textbf{58.08} & \textbf{60.61} & \textbf{52.57} & & & \\
\midrule
\multirow{4}{*}{\textsc{FlightABSA}} & Single-order & 37.93 & 59.27 & 66.38 & 67.59 & 57.79 & \multicolumn{1}{c}{\multirow{4}{*}{69.74}} & \multicolumn{1}{c}{\multirow{4}{*}{68.67}} & \multicolumn{1}{c}{\multirow{4}{*}{\underline{\textbf{73.36}}}} & 24.32 & 47.29 & 53.33 & 54.90 & 44.96 & \multicolumn{1}{c}{\multirow{4}{*}{57.76}} & \multicolumn{1}{c}{\multirow{4}{*}{57.90}} & \multicolumn{1}{c}{\multirow{4}{*}{\underline{\textbf{64.49}}}} \\
& Reasoning & \textbf{54.34} & 58.62 & 61.43 & 63.09 & 59.37 & & & & \textbf{38.43} & 42.62 & 46.88 & 45.28 & 43.30 & & & \\
& Self-consistency & 40.96 & 59.07 & 67.43 & 68.17 & 58.91 & & & & 28.62 & 48.75 & 55.12 & 56.24 & 47.18 & & & \\
& LLM-MvP (ours) & 39.11 & \textbf{62.22} & \textbf{68.36} & \textbf{70.11} & \textbf{59.95} & & & & 34.06 & \textbf{50.47} & \textbf{55.80} & \textbf{57.37} & \textbf{49.42} & & & \\
\midrule
\multirow{4}{*}{\textsc{Coursera}} & Single-order & 21.13 & 38.52 & 44.06 & 46.09 & 37.45 & \multicolumn{1}{c}{\multirow{4}{*}{51.86}} & \multicolumn{1}{c}{\multirow{4}{*}{50.97}} & \multicolumn{1}{c}{\multirow{4}{*}{\underline{\textbf{52.94}}}} & 6.26 & 22.01 & 27.49 & 28.14 & 20.97 & \multicolumn{1}{c}{\multirow{4}{*}{32.34}} & \multicolumn{1}{c}{\multirow{4}{*}{32.50}} & \multicolumn{1}{c}{\multirow{4}{*}{\underline{\textbf{33.66}}}} \\
& Reasoning & \textbf{26.26} & 33.50 & 36.92 & 39.96 & 34.16 & & & & \textbf{13.57} & 19.71 & 21.97 & 23.97 & 19.81 & & & \\
& Self-consistency & 22.36 & 39.43 & 44.42 & 47.20 & 38.35 & & & & 7.47 & \textbf{24.28} & \textbf{29.50} & \textbf{30.70} & \textbf{22.99} & & & \\
& LLM-MvP (ours) & 21.81 & \textbf{41.01} & \textbf{47.22} & \textbf{49.12} & \textbf{39.79} & & & & 9.46 & 22.93 & 28.75 & 30.13 & 22.82 & & & \\
\midrule
\multirow{4}{*}{\textsc{Hotels}} & Single-order & 33.93 & 61.09 & 64.44 & 66.06 & 56.38 & \multicolumn{1}{c}{\multirow{4}{*}{67.70}} & \multicolumn{1}{c}{\multirow{4}{*}{69.37}} & \multicolumn{1}{c}{\multirow{4}{*}{\underline{\textbf{71.47}}}} & 18.91 & 44.27 & 47.74 & 49.66 & 40.15 & \multicolumn{1}{c}{\multirow{4}{*}{53.87}} & \multicolumn{1}{c}{\multirow{4}{*}{55.03}} & \multicolumn{1}{c}{\multirow{4}{*}{\underline{\textbf{56.54}}}} \\
& Reasoning & \textbf{44.09} & 53.88 & 57.26 & 56.75 & 53.00 & & & & \textbf{32.61} & 39.91 & 42.30 & 42.26 & 39.27 & & & \\
& Self-consistency & 36.94 & 62.20 & 65.14 & 66.72 & 57.75 & & & & 21.21 & 47.23 & 49.45 & 50.95 & 42.21 & & & \\
& LLM-MvP (ours) & 37.07 & \textbf{64.33} & \textbf{67.42} & \textbf{68.38} & \textbf{59.30} & & & & 27.40 & \textbf{47.90} & \textbf{52.78} & \textbf{54.16} & \textbf{45.56} & & & \\
\bottomrule
\end{tabular}

}
\caption{\textbf{Main performance comparison ($F_1$ score) across Target Aspect Sentiment Detection (TASD) and Aspect Sentiment Quad Prediction (ASQP).} \textbf{Bold} values indicate the best performance per dataset and shot configuration within each task. \underline{Underlined} scores denote the overall best performance for a dataset within its respective task.}
\label{tab:results}
\end{table*}

We evaluated LLM-MvP against two categories of prior SOTA methods. The first category comprises fine-tuned approaches, namely \textbf{Paraphrase}~\citep{zhang2021aspect}, \textbf{MvP}~\citep{gou2023mvp}, and a \textbf{LoRA-adapted Gemma 4 (31B)}, which are outlined in detail in Appendix~\ref{appendix:fine-tuning}. The second category encompasses prompting-based approaches, which are detailed below. First, \textbf{single-order prompting} (Figure~\ref{fig:main-approach}) executes the LLM once with a fixed permutation of sentiment elements, as practiced by \citet{hellwig2025still}, \citet{bai2024compound}, \citet{gou2023mvp}, and \citet{zhang2024sentiment}. Second, \textbf{Self-Consistency (SC) prompting}, as proposed by \citet{hellwig2025still}, executes the same prompt five times with temperature 0.8 to obtain diverse outputs while keeping the element ordering fixed. Third, a \textbf{Reasoning LLM} performs chain-of-thought reasoning before generating the prediction. Since reasoning outputs comprise thousands of tokens rather than a short list of tuples, running our full experimental grid with a dense Gemma 4-based reasoning model was computationally infeasible. We therefore employed OpenAI's open-source mixture-of-experts model gpt-oss with 120 billion parameters, of which only 5.1 billion are active per token \citep{agarwal2025gpt}, as a substantially faster alternative. We configured it with a ``medium'' reasoning budget to cap reasoning length. We also considered other open-source reasoning models (e.g., GLM-4.7-Flash\footnote{\hficon\url{https://hf.co/zai-org/GLM-4.7-Flash}}), but they frequently produced reasoning chains exceeding our 16,384-token context window in initial tests.

For all prompt-based approaches, we employed guided decoding to enforce valid output formatting and vLLM for efficient inference. Unless stated otherwise (i.e., temperature 0.8 for SC and gpt-oss for the reasoning baseline), all approaches used Gemma 4 (31B) with temperature set to 0. This standardization allowed for a controlled comparison focused on the prompting strategies themselves. In total, the experiments reported in this paper comprised \val{prompts.total} prompt executions. %

\section{Results}

\subsection{Task Performance}

We present performance scores for all datasets and prompting techniques in Table~\ref{tab:results} for TASD and ASQP (significance testing details in Appendix~\ref{sec:significance}). No statistically significant differences in task performance were observed between methods.

\paragraph{LLM-MvP and reasoning models showed strong zero-shot performance.}
Differences are most pronounced under zero-shot conditions. For instance, on the ASQP task with Rest16, LLM-MvP achieves an F1 score of \val{f1.rest16.asqp.mvp.0}, compared to \val{f1.rest16.asqp.so.0} for single-order prompting. While LLM-MvP achieves higher scores than other prompting approaches across nearly all configurations, reasoning-based prompting excels on many task-dataset combinations in the zero-shot setting. For example, on Hotels (TASD), reasoning achieves \val{f1.hotels.tasd.reasoning.0} compared to \val{f1.hotels.tasd.mvp.0} for LLM-MvP.

\paragraph{LLM-MvP approached or exceeded SOTA fine-tuned performance.}
Examining few-shot performance revealed a consistent upward trend in F1 scores with increasing numbers of demonstrations, with only rare exceptions. In the few-shot regime, LLM-MvP achieved the highest performance across most configurations, while reasoning-based models yielded the lowest scores. Performance gains are particularly pronounced for ASQP. For instance, on Hotels (ASQP) in the 100-shot setting, LLM-MvP achieves an F1 score of \val{f1.hotels.asqp.mvp.100} compared to \val{f1.hotels.asqp.so.100} for single-order prompting.

Using only 100 annotated examples, LLM-MvP surpassed both fine-tuned T5-based models (Paraphrase and MvP) on Rest15, Rest16, and FlightABSA (TASD), and surpassed Paraphrase but not MvP on Hotels; only on Coursera and FlightABSA (ASQP) did it remain below both, as illustrated by the line plots in Appendix~\ref{sec:fs-performance}. A gap of roughly \val{gap.ft.min.int}--\val{gap.ft.max.int} percentage points remained relative to a LoRA-adapted Gemma 4 fine-tuned on about a thousand labeled examples, narrowest for TASD on Rest15 (\val{gap.ft.min} points) and widest for ASQP on FlightABSA (\val{gap.ft.max} points).

Precision, recall, and macro-averaged F1 (see Appendix~\ref{sec:task-performance}) give a more nuanced picture. In the zero-shot setting, the reasoning LLM achieved the highest scores on most datasets. In the few-shot settings, the F1 gains of LLM-MvP on ASQP stem primarily from higher recall: self-consistency achieved higher precision than LLM-MvP in most few-shot ASQP configurations, whereas LLM-MvP achieved the highest recall in nearly all few-shot settings. The exceptions are FlightABSA (TASD), where the reasoning LLM yielded higher recall at all shot counts, and isolated cases on Coursera (ASQP). For macro-averaged F1, LLM-MvP performed best in most few-shot settings, with exceptions only on Coursera.

\paragraph{New SOTA performance scores for prompting-based ABSA.} Prior work, including \citet{gou2023mvp}, \citet{bai2024compound}, \citet{li2025aligningblack}, \citet{hellwig2025still}, and \citet{ventirozos-etal-2025-aspect}, considered zero-shot, 10-shot, or 100-shot learning paradigms. Across all settings, LLM-MvP outperformed all previously reported scores, as did the reasoning LLM in the zero-shot setting (see Table~\ref{tab:past-paper-performance}).

\begin{table}[t]
\centering
\setlength{\tabcolsep}{6pt}
\renewcommand{\arraystretch}{1.0}
\newcommand{\srcrow}[1]{\quad \scriptsize{#1}}
\resizebox{1.0\columnwidth}{!}{%
\begin{tabular}{
    @{}l
    S[table-format=2.2]
    S[table-format=2.2]
    S[table-format=2.2]
    S[table-format=2.2]@{}
}
\toprule
& \multicolumn{2}{c}{\textbf{Rest15}} & \multicolumn{2}{c}{\textbf{Rest16}} \\
\cmidrule(lr){2-3} \cmidrule(lr){4-5}
\textbf{Approach} & \textbf{TASD} & \textbf{ASQP} & \textbf{TASD} & \textbf{ASQP} \\
\midrule
\rowcolor{gray!10} \multicolumn{5}{l}{\textit{Zero-shot}} \\
LLM-MvP (Ours) & 40.88 & 32.88 & 51.03 & \textbf{39.15} \\
gpt-oss-120B (CoT) & \textbf{48.18} & \textbf{33.33} & \textbf{54.59} & 38.27 \\
ChatABSA \citep{bai2024compound} & 39.21 & 27.11 & 41.28 & 30.42 \\
gpt-3.5-turbo \citep{zhang2024sentiment} & \multicolumn{1}{c}{-} & 10.46 & \multicolumn{1}{c}{-} & 14.02 \\
gpt4o-mini \citep{li2025aligningblack} & \multicolumn{1}{c}{-} & 25.24 & \multicolumn{1}{c}{-} & 34.31 \\
gpt4o-mini (CoT) \citep{li2025aligningblack} & \multicolumn{1}{c}{-} & 21.55 & \multicolumn{1}{c}{-} & 26.73 \\
Gemma-3-27B (SC) \citep{hellwig2025still} & 30.36 & 24.73 & 45.51 & 28.96 \\
\midrule
\rowcolor{gray!10} \multicolumn{5}{l}{\textit{10-shot}} \\
LLM-MvP (Ours) & \textbf{63.29} & \textbf{45.56} & \textbf{67.79} & \textbf{52.46} \\
gpt-oss-120B (CoT) & 50.55 & 36.39 & 55.97 & 42.05 \\
ChatABSA \citep{bai2024compound} & 45.93 & 32.14 & 47.00 & 33.26 \\
gpt-3.5-turbo \citep{zhang2024sentiment} & \multicolumn{1}{c}{-} & 30.92 & \multicolumn{1}{c}{-} & 40.15 \\
Gemma-3-27B (SC) \citep{hellwig2025still} & 54.47 & 39.95 & 66.75 & 46.23 \\
\midrule
\rowcolor{gray!10} \multicolumn{5}{l}{\textit{100-shot}} \\
LLM-MvP (Ours) & \textbf{70.50} & \textbf{53.40} & \textbf{74.95} & \textbf{60.61} \\
gpt-oss-120B (CoT) & 57.08 & 39.95 & 60.00 & 46.62 \\
Gemini 2.0 Flash \citep{ventirozos-etal-2025-aspect} & \multicolumn{1}{c}{-} & 50.70 & \multicolumn{1}{c}{-} & 54.40 \\
\bottomrule
\end{tabular}
}
\caption{\textbf{Performance of LLM-MvP compared to prior benchmarks reported on TASD and ASQP.}}
\label{tab:past-paper-performance}
\end{table}

\paragraph{Ablation: Component analysis.}
The number of views $m=5$ used throughout this paper was fixed a priori rather than tuned on the test sets. Varying $m$ (Appendix~\ref{appendix:m-variation}) shows that $m=\val{mvar.tasd.best.m}$ is the best choice for TASD (\val{mvar.tasd.best.avg} average $F_1$), while for ASQP the best value, $m=\val{mvar.asqp.best.m}$, raises the average $F_1$ only from \val{mvar.asqp.m5.avg} to \val{mvar.asqp.best.avg}. Comparing LLM-MvP with single-order prompting for each of the 6 (TASD) and 24 (ASQP) element orders individually (Appendix~\ref{appendix:permutations}), LLM-MvP outperformed the default order and the average over all orders in every setting, and even surpassed the best order chosen in hindsight in \val{perm.tasd.beats.oracle} (TASD) and \val{perm.asqp.beats.oracle} (ASQP) of 20 settings. We evaluated the individual contributions of LLM-MvP's components in Table~\ref{tab:mvp-variants} (results for all datasets can be found in Appendix~\ref{appendix:mvp-variants-all}). Removing guided decoding (\mbox{LLM-MvP w/o GD}) led to a performance decline in most settings, particularly on the more complex ASQP task in the zero-shot setting (e.g., \val{f1.rest16.asqp.nogd.0} vs.\ \val{f1.rest16.asqp.mvp.0} on Rest16). A notable exception is Coursera (ASQP), where \mbox{LLM-MvP w/o GD} outperformed the full approach in all few-shot settings. Prefix caching was not ablated, as it affects only inference time and energy consumption, not predictive performance.

Furthermore, we investigated whether the computational cost of multi-view prompting was necessary for every test instance. We hypothesized that ABSA could be performed for less complex examples via single-order inference, thereby reducing resource consumption. To test this, the cost-efficient variant \mbox{LLM-MvP\textsubscript{eff}} restricted the computation of all permutations to the 25\% of test examples with the lowest mean token-level confidence in the single-order prediction. \mbox{LLM-MvP\textsubscript{eff}} improved over single-order prompting in all but one setting (FlightABSA, TASD, 0-shot), suggesting that token-level confidence is a useful proxy for instance difficulty. Nevertheless, the full \mbox{LLM-MvP} framework remained superior in nearly all settings, with the exceptions of Rest15 (TASD, 0-shot) and the aforementioned Coursera (ASQP) results.

\begin{table}[t]
\centering
\setlength{\tabcolsep}{3pt}
\resizebox{1.0\columnwidth}{!}{%
\begin{tabular}{
    @{}l
    l
    l
    S[table-format=2.2]
    S[table-format=2.2]
    S[table-format=2.2]
    S[table-format=2.2]@{}
}
\toprule
\multirow{2}{*}{\textbf{Task}} & \multirow{2}{*}{\textbf{Dataset}} & \multirow{2}{*}{\textbf{Approach}} & \multicolumn{4}{c}{\textbf{\# Shots} ($F_1$)} \\
\cmidrule(lr){4-7}
& & & \textbf{0} & \textbf{10} & \textbf{50} & \textbf{100} \\
\midrule
\multirow{8}{*}{TASD} & \multirow{4}{*}{\textsc{Rest15}} 
& Single-order & 40.48 & 58.82 & 63.05 & 65.27 \\
& & LLM-MvP\textsubscript{eff} & \textbf{41.03} & 60.90 & 65.38 & 67.37 \\
& & LLM-MvP (w/o GD) & 41.02 & 62.26 & 66.14 & 67.97 \\
& & LLM-MvP & 40.88 & \textbf{63.29} & \textbf{68.18} & \textbf{70.50} \\
\cmidrule(lr){2-7}
& \multirow{4}{*}{\textsc{Rest16}} 
& Single-order & 50.10 & 64.30 & 69.30 & 70.74 \\
& & LLM-MvP\textsubscript{eff} & 50.52 & 65.46 & 71.45 & 72.96 \\
& & LLM-MvP (w/o GD) & 50.30 & 67.19 & 71.35 & 73.70 \\
& & LLM-MvP & \textbf{51.03} & \textbf{67.79} & \textbf{73.19} & \textbf{74.95} \\
\midrule
\multirow{8}{*}{ASQP} & \multirow{4}{*}{\textsc{Rest15}} 
& Single-order & 25.59 & 40.43 & 47.99 & 50.70 \\
& & LLM-MvP\textsubscript{eff} & 27.51 & 42.08 & 49.19 & 51.99 \\
& & LLM-MvP (w/o GD) & 20.33 & 43.00 & 49.91 & 52.65 \\
& & LLM-MvP & \textbf{32.88} & \textbf{45.56} & \textbf{50.99} & \textbf{53.40} \\
\cmidrule(lr){2-7}
& \multirow{4}{*}{\textsc{Rest16}} 
& Single-order & 30.13 & 48.98 & 54.70 & 56.36 \\
& & LLM-MvP\textsubscript{eff} & 32.64 & 49.87 & 56.46 & 58.16 \\
& & LLM-MvP (w/o GD) & 29.91 & 50.85 & 57.04 & 59.82 \\
& & LLM-MvP & \textbf{39.15} & \textbf{52.46} & \textbf{58.08} & \textbf{60.61} \\
\bottomrule
\end{tabular}

}
\caption{\textbf{Ablation study: comparison of the LLM-MvP framework against its efficient variant (LLM-MvP\textsubscript{eff}), a version without guided decoding (GD), and the single-order baseline.} LLM-MvP\textsubscript{eff} reduces inference cost by requesting all permutations only for the 25\% of test instances with the lowest mean token-level confidence in the single-order prediction.}
\label{tab:mvp-variants}
\end{table}

\paragraph{Ablation: LLM selection.}
To assess the generalizability of LLM-MvP across model architectures, we further evaluated it on models developed by different research groups across distinct geographic regions, including the French Mistral Small 3.2 (24B)\footnote{\hficon\url{https://hf.co/mistralai/Mistral-Small-3.2-24B-Instruct-2506}} and the Chinese Qwen 3 (30B)\footnote{\hficon\url{https://hf.co/Qwen/Qwen3-30B-A3B-Instruct-2507}}. The results (see Appendix~\ref{appendix:llm-comparison}) showed that LLM-MvP consistently improved over single-order prompting across all evaluated models with only a few exceptions. For both tasks, Gemma 4 achieved the strongest performance in nearly all few-shot settings (the exception being FlightABSA (ASQP) at 50 and 100 shots, where Mistral performed best), while in the zero-shot setting, LLM-MvP with Mistral outperformed Gemma in all configurations except Rest15 (TASD), and Qwen did so in half of the configurations. Finally, we also evaluated the mixture-of-experts variant Gemma 4 (26B MoE)\footnote{\hficon\url{https://hf.co/google/gemma-4-26B-A4B}}; with LLM-MvP, it consistently underperformed the dense Gemma~4 (31B) across all tasks, datasets, and shot configurations.

\begin{table}[t]
\centering
\setlength{\tabcolsep}{3pt}
\resizebox{1.0\columnwidth}{!}{%
\begin{tabular}{@{}llrrrr@{}}
\toprule
& & \multicolumn{4}{c}{\textbf{\# Shots}} \\
\cmidrule(lr){3-6}
\textbf{Task} & \textbf{Approach} & \textbf{0} & \textbf{10} & \textbf{50} & \textbf{100} \\
\midrule
\multirow{7}{*}{\textsc{TASD}}
& Single-order          & \textbf{7.77} & \textbf{7.21} & \textbf{9.44} & \textbf{10.26} \\
& Single-order (nc)     & 41.49 & 72.99 & 226.65 & 417.06 \\
\cmidrule(lr){2-6}
& Reasoning             & 10.25 & 13.29 & 16.68 & 19.17 \\
\cmidrule(lr){2-6}
& Self-consistency      & 26.91 & 23.94 & 27.04 & 28.96 \\
& Self-consistency (nc) & 206.57 & 366.24 & 1144.76 & 2107.03 \\
\cmidrule(lr){2-6}
& LLM-MvP               & 49.65 & 43.06 & 54.00 & 64.41 \\
& LLM (FT)\rlap{$^\dagger$}              & \multicolumn{4}{c}{12.61} \\
\midrule
\multirow{7}{*}{\textsc{ASQP}}
& Single-order          & \textbf{9.28} & \textbf{10.62} & \textbf{12.33} & \textbf{14.09} \\
& Single-order (nc)     & 50.27 & 88.15 & 270.39 & 516.45 \\
\cmidrule(lr){2-6}
& Reasoning             & 14.21 & 16.40 & 20.87 & 25.29 \\
\cmidrule(lr){2-6}
& Self-consistency      & 34.85 & 33.41 & 37.92 & 40.32 \\
& Self-consistency (nc) & 248.99 & 440.01 & 1363.75 & 2618.89 \\
\cmidrule(lr){2-6}
& LLM-MvP               & 219.23 & 264.86 & 307.62 & 352.60 \\
& LLM (FT)\rlap{$^\dagger$}              & \multicolumn{4}{c}{15.92} \\
\bottomrule
\multicolumn{6}{l}{\footnotesize $^\dagger$Fine-tuned on full training set; shown as a shot-independent reference.}\\
\end{tabular}
}
\caption{\textbf{Energy consumption in milliwatt-hours (mWh) across prompting methods and few-shot configurations per predicted test instance.} vLLM's prefix caching reduces energy consumption when compared to executing methods without caching (nc). \textbf{Bold} values indicate the lowest energy consumption.}
\label{tab:energy-consumption}
\end{table}

\paragraph{Scaling efficiency through prefix caching.}
As shown in Table~\ref{tab:energy-consumption}, prefix caching substantially reduced energy consumption compared to the independent execution (no caching) of the baseline approaches. Although LLM-MvP processes multiple views, it consumed less energy than non-cached self-consistency (SC-nc) in all settings. For TASD (six views), it also consumed less energy than non-cached single-order prompting (SO-nc) from 10 shots onwards; for instance, in the 50-shot TASD setting, LLM-MvP is significantly more efficient than both non-cached variants ($p < .001$, see Appendix~\ref{sec:significance}). For ASQP (24 views), LLM-MvP fell below SO-nc only at 100 shots, so the overhead of multi-view prediction is offset by prefix caching primarily at higher shot counts.

Prefix caching also improved scaling with the number of shots: while energy consumption of non-cached methods grew roughly linearly with the number of shots due to the increasing prompt length (e.g., SO-nc on ASQP: \val{energy.asqp.so-nc.0.int} to \val{energy.asqp.so-nc.100.int}\,mWh from 0 to 100 shots), the increase for cached methods was far smaller (e.g., LLM-MvP on ASQP: \val{energy.asqp.mvp.0.int} to \val{energy.asqp.mvp.100.int}\,mWh). This efficiency gain is not exclusive to LLM-MvP: all cached strategies benefit from the shared prefix computation, as also reflected in the inference durations (Table~\ref{tab:time-statistics}).

\begin{table}[t]
\centering
\small
\setlength{\tabcolsep}{3pt}
\resizebox{1.0\columnwidth}{!}{%
\begin{tabular}{@{}lcccccccc@{}}
\toprule
& \multicolumn{8}{c}{\textbf{\# Shots}} \\
\cmidrule(lr){2-9}
& \multicolumn{4}{c}{\textbf{\textsc{TASD}}} & \multicolumn{4}{c}{\textbf{\textsc{ASQP}}} \\
\cmidrule(lr){2-5} \cmidrule(lr){6-9}
\textbf{Approach} & \textbf{0} & \textbf{10} & \textbf{50} & \textbf{100} & \textbf{0} & \textbf{10} & \textbf{50} & \textbf{100} \\
\midrule
Single-order & \textbf{0.047} & \textbf{0.043} & \textbf{0.057} & \textbf{0.062} & \textbf{0.056} & \textbf{0.064} & \textbf{0.074} & \textbf{0.085} \\
Single-order (nc) & 0.249 & 0.438 & 1.360 & 2.503 & 0.302 & 0.529 & 1.623 & 3.099 \\
\midrule
Reasoning & 0.062 & 0.080 & 0.103 & 0.141 & 0.086 & 0.099 & 0.134 & 0.202 \\
\midrule
Self-consistency & 0.162 & 0.144 & 0.163 & 0.174 & 0.209 & 0.201 & 0.228 & 0.243 \\
Self-consistency (nc) & 1.240 & 2.198 & 6.869 & 12.644 & 1.495 & 2.641 & 8.183 & 15.715 \\
\midrule
LLM-MvP & 0.299 & 0.259 & 0.325 & 0.388 & 1.319 & 1.593 & 1.850 & 2.121 \\
LLM (FT) & \multicolumn{4}{c}{0.083} & \multicolumn{4}{c}{0.103} \\
\bottomrule
\end{tabular}

}
\caption{\textbf{Inference duration in seconds per predicted test instance.} \textbf{Bold} values indicate the fastest method (excluding the shot-independent LLM (FT) reference).}
\label{tab:time-statistics}
\end{table} %

\section{Discussion}

This work introduced LLM-based Multi-view Prompting (LLM-MvP), a resource-efficient approach to ABSA that narrows the gap between few-shot prompting and models fine-tuned on the full training set, using at most 100 demonstrations. While \citet{gou2023mvp} established the multi-view principle in a fine-tuning regime demanding about a thousand annotated examples, LLM-MvP adapts this principle to the prompting setting: it uses token-level entropy as a supervision-free signal for view selection and combines schema-constrained decoding with prefix-cached batch inference. The gap to LoRA-adapted Gemma 4 trained on the full training set remains narrow on most datasets, amounting to roughly \val{gap.ft.min.int}--\val{gap.ft.max.int} percentage points, despite the fine-tuned model having access to roughly an order of magnitude more labeled data than the 100-shot setting. Beyond performance, prefix-cached batch inference keeps the cost of processing up to 24 permutations per instance manageable: LLM-MvP is more energy-efficient than non-cached self-consistency in all settings and than non-cached single-order prompting at higher shot counts, with inference times below \val{time.tasd.mvp.max.ceil}\,s per example for TASD and roughly \val{time.asqp.mvp.min}--\val{time.asqp.mvp.max}\,s for ASQP. A fine-tuned model remains cheaper at inference time (e.g., \val{energy.tasd.ft} vs.\ \val{energy.tasd.mvp.min}--\val{energy.tasd.mvp.max}\,mWh per instance for TASD).

Our performance analysis yields several observations that extend prior work. Consistent with findings by \citet{hellwig2025still}, \citet{vsmid2024llama}, and \citet{gou2023mvp}, increasing the number of few-shot examples improves performance. The largest gain occurs from 0 to 10 shots (e.g., \val{f1.rest15.tasd.mvp.0} to \val{f1.rest15.tasd.mvp.10} for LLM-MvP on Rest15, TASD), while 10 to 100 shots still add up to about \val{gain.mvp.10to100.max.int} points (e.g., \val{f1.rest15.tasd.mvp.10} to \val{f1.rest15.tasd.mvp.100} on Rest15, TASD; \val{f1.rest16.asqp.mvp.10} to \val{f1.rest16.asqp.mvp.100} on Rest16, ASQP). LLM-MvP outperforms single-order and self-consistency prompting across most shot settings and tasks, with occasional exceptions. While reasoning models occasionally outperform LLM-MvP by several percentage points in zero-shot settings, they substantially underperform other prompting strategies in few-shot configurations. %

Employing all permutations for majority voting did not yield the best results: performance peaked at an intermediate $m$ and then declined, suggesting that low-confidence permutations introduce noise that outweighs the benefit of additional views. In the fine-tuning regime, \citet{jun-lee-2025-dynamic} sidestep this by learning to predict the most informative views from labeled data. Without such supervision, LLM-MvP\textsubscript{eff} offers a practical alternative, indicating that token-level confidence is useful both for view selection and for deciding whether multi-view inference is warranted at all.

Future work could adapt the technical principles underlying LLM-MvP for LLM fine-tuning, combining multi-view inference with parameter-efficient adaptation, or extend the approach to ABSA tasks with fewer sentiment elements (e.g., end-to-end ABSA: aspect term + polarity) or to structurally similar tasks such as Named Entity Recognition (NER) \citep{yadav-bethard-2018-survey, JEHANGIR2023100017} and relation extraction \citep{bassignana-plank-2022-mean, DETROJA2023200244}.

\section{Limitations}

While our evaluation spans five diverse datasets and demonstrates consistent performance gains across tasks, shot counts, and model selections, we do not claim statistical significance for every individual sub-setting. Given the ongoing \textit{replication crisis} in NLP and the documented risk of overestimating performance margins under many simultaneous tests \citep{10.1145/3360311}, we report significance conservatively.

Beyond statistical considerations, several practical limitations apply. First, we cannot exclude that training examples from the evaluated datasets may have been present in the LLMs' pre-training corpora, except for FlightABSA, whose labels are available only upon request. Second, the computational efficiency (but not predictive performance) of LLM-MvP depends on available GPU memory, since larger KV caches translate directly into faster inference, though smaller GPU configurations still benefit from caching to a lesser extent. Third, LLM-MvP requires open-weight models for logit access and guided decoding, limiting its direct applicability to closed-source proprietary systems. That said, since our open-source implementation builds on the open-source framework vLLM, supported by major cloud providers including Google Cloud Vertex AI, LLM-MvP remains readily deployable in production environments. Fourth, the reasoning baseline uses gpt-oss-120B rather than Gemma 4 (31B), since chain-of-thought generation with the dense Gemma model across our full experimental grid was computationally infeasible. Differences between the reasoning baseline and the other prompting strategies therefore reflect both the prompting strategy and the underlying model.
\section{Ethical Considerations}

While LLM-MvP reduces the need for extensive human annotation and fine-tuning compute compared to fine-tuned approaches, repeated inference on LLMs incurs non-negligible energy consumption and environmental impact. The use of AI assistants in this work is disclosed in the Acknowledgments.

\section*{Acknowledgments}

We thank the anonymous reviewers for their comments and suggestions. We also thank our colleagues at the Media Informatics Group and the Information Science Group at the University of Regensburg for reading and commenting on earlier drafts of this paper.

The proposed method and the experimental design originate from the authors. AI assistants were not used for brainstorming or idea generation. We used Claude Sonnet 4.5\footnote{\url{https://www.anthropic.com/claude/sonnet}} as an AI assistant for paraphrasing and polishing text written by the authors, and for refining and debugging the experiment and evaluation code. Figures~\ref{fig:main-approach} and~\ref{fig:prompt} were designed by the authors in PowerPoint and visually refined with Claude into vector graphics. The authors reviewed all AI-assisted content and take full responsibility for all claims, results, and scientific contributions.

\bibliography{bibliography}

\begin{thebibliography}{39}
\providecommand{\natexlab}[1]{#1}

\bibitem[{Bai et~al.(2024)Bai, Han, Zhao, Gao, Zhang, Wang, and
  Hu}]{bai2024compound}
Yinhao Bai, Zhixin Han, Yuhua Zhao, Hang Gao, Zhuowei Zhang, Xunzhi Wang, and
  Mengting Hu. 2024.
\newblock \href {https://doi.org/10.18653/v1/2024.findings-emnlp.460} {Is
  compound aspect-based sentiment analysis addressed by {LLM}s?}
\newblock In \emph{Findings of the Association for Computational Linguistics:
  EMNLP 2024}, pages 7836--7861, Miami, Florida, USA. Association for
  Computational Linguistics.

\bibitem[{Bassignana and Plank(2022)}]{bassignana-plank-2022-mean}
Elisa Bassignana and Barbara Plank. 2022.
\newblock \href {https://doi.org/10.18653/v1/2022.acl-srw.7} {What do you mean
  by relation extraction? a survey on datasets and study on scientific relation
  classification}.
\newblock In \emph{Proceedings of the 60th Annual Meeting of the Association
  for Computational Linguistics: Student Research Workshop}, pages 67--83,
  Dublin, Ireland. Association for Computational Linguistics.

\bibitem[{Chebolu et~al.(2024)Chebolu, Dernoncourt, Lipka, and
  Solorio}]{chebolu2024oats}
Siva Uday~Sampreeth Chebolu, Franck Dernoncourt, Nedim Lipka, and Thamar
  Solorio. 2024.
\newblock \href {https://aclanthology.org/2024.lrec-main.1080/} {{OATS}: A
  challenge dataset for opinion aspect target sentiment joint detection for
  aspect-based sentiment analysis}.
\newblock In \emph{Proceedings of the 2024 Joint International Conference on
  Computational Linguistics, Language Resources and Evaluation (LREC-COLING
  2024)}, pages 12336--12347, Torino, Italia. ELRA and ICCL.

\bibitem[{Cockburn et~al.(2020)Cockburn, Dragicevic, Besan\c{c}on, and
  Gutwin}]{10.1145/3360311}
Andy Cockburn, Pierre Dragicevic, Lonni Besan\c{c}on, and Carl Gutwin. 2020.
\newblock \href {https://doi.org/10.1145/3360311} {Threats of a replication
  crisis in empirical computer science}.
\newblock \emph{Commun. ACM}, 63(8):70–79.

\bibitem[{Daniel~Han and team(2023)}]{unsloth}
Michael~Han Daniel~Han and Unsloth team. 2023.
\newblock \href {http://github.com/unslothai/unsloth} {Unsloth}.

\bibitem[{Dao and Gu(2024)}]{mamba2}
Tri Dao and Albert Gu. 2024.
\newblock Transformers are {SSM}s: Generalized models and efficient algorithms
  through structured state space duality.
\newblock In \emph{International Conference on Machine Learning (ICML)}.

\bibitem[{Detroja et~al.(2023)Detroja, Bhensdadia, and
  Bhatt}]{DETROJA2023200244}
Kartik Detroja, C.K. Bhensdadia, and Brijesh~S. Bhatt. 2023.
\newblock \href {https://doi.org/10.1016/j.iswa.2023.200244} {A survey on
  relation extraction}.
\newblock \emph{Intelligent Systems with Applications}, 19:200244.

\bibitem[{Dong and Wei(2025)}]{DONG2025125933}
Hao Dong and Wei Wei. 2025.
\newblock \href {https://doi.org/10.1016/j.eswa.2024.125933} {Pgso:
  Prompt-based generative sequence optimization network for aspect-based
  sentiment analysis}.
\newblock \emph{Expert Systems with Applications}, 265:125933.

\bibitem[{Dong et~al.(2025)Dong, Ruan, Cai, Xu, Zhao, Lai, and
  Chen}]{dong2024xgrammar}
Yixin Dong, Charlie~F. Ruan, Yaxing Cai, Ziyi Xu, Yilong Zhao, Ruihang Lai, and
  Tianqi Chen. 2025.
\newblock \href
  {https://proceedings.mlsys.org/paper_files/paper/2025/file/5c20ca4b0b20b0bd2f1d839dc605e70f-Paper-Conference.pdf}
  {Xgrammar: Flexible and efficient structured generation engine for large
  language models}.
\newblock In \emph{Proceedings of Machine Learning and Systems}, volume~7.
  MLSys.

\bibitem[{Fehle et~al.(2026)Fehle, Kruschwitz, Hellwig, and
  Wolff}]{fehle2026leveraging}
Jakob Fehle, Udo Kruschwitz, Nils~Constantin Hellwig, and Christian Wolff.
  2026.
\newblock \href {https://doi.org/10.1016/j.knosys.2026.115277} {Leveraging
  fine-tuning of large language models for aspect-based sentiment analysis in
  resource-scarce environments}.
\newblock \emph{Knowledge-Based Systems}, 336:115277.

\bibitem[{Fisher(1992)}]{Fisher1992}
R.~A. Fisher. 1992.
\newblock \href {https://doi.org/10.1007/978-1-4612-4380-9_6}
  {\emph{Statistical Methods for Research Workers}}, pages 66--70.
\newblock Springer New York, New York, NY.

\bibitem[{{Ghiasvand Mohammadkhani} et~al.(2025){Ghiasvand Mohammadkhani},
  Ranjbar, and Momtazi}]{GHIASVANDMOHAMMADKHANI2025107847}
Mohammad {Ghiasvand Mohammadkhani}, Niloofar Ranjbar, and Saeedeh Momtazi.
  2025.
\newblock \href {https://doi.org/10.1016/j.neunet.2025.107847} {E2tp: Element
  to tuple prompting improves aspect sentiment tuple prediction}.
\newblock \emph{Neural Networks}, 191:107847.

\bibitem[{Gou et~al.(2023)Gou, Guo, and Yang}]{gou2023mvp}
Zhibin Gou, Qingyan Guo, and Yujiu Yang. 2023.
\newblock \href {https://doi.org/10.18653/v1/2023.acl-long.240} {{M}v{P}:
  Multi-view prompting improves aspect sentiment tuple prediction}.
\newblock In \emph{Proceedings of the 61st Annual Meeting of the Association
  for Computational Linguistics (Volume 1: Long Papers)}, pages 4380--4397,
  Toronto, Canada. Association for Computational Linguistics.

\bibitem[{Hasan et~al.(2024)Hasan, Das, Anjum, Alam, Anjum, Sarker, and
  Noori}]{hasan2024zero}
Md.~Arid Hasan, Shudipta Das, Afiyat Anjum, Firoj Alam, Anika Anjum, Avijit
  Sarker, and Sheak Rashed~Haider Noori. 2024.
\newblock \href {https://aclanthology.org/2024.lrec-main.1549/} {Zero- and
  few-shot prompting with {LLM}s: A comparative study with fine-tuned models
  for {B}angla sentiment analysis}.
\newblock In \emph{Proceedings of the 2024 Joint International Conference on
  Computational Linguistics, Language Resources and Evaluation (LREC-COLING
  2024)}, pages 17808--17818, Torino, Italia. ELRA and ICCL.

\bibitem[{Hellwig et~al.(2025)Hellwig, Fehle, Kruschwitz, and
  Wolff}]{hellwig2025still}
Nils~Constantin Hellwig, Jakob Fehle, Udo Kruschwitz, and Christian Wolff.
  2025.
\newblock \href {https://doi.org/10.18653/v1/2025.xllm-1.15} {Do we still need
  human annotators? prompting large language models for aspect sentiment quad
  prediction}.
\newblock In \emph{Proceedings of the 1st Joint Workshop on Large Language
  Models and Structure Modeling (XLLM 2025)}, pages 153--172, Vienna, Austria.
  Association for Computational Linguistics.

\bibitem[{Holm(1979)}]{1979holmbonferroni}
Sture Holm. 1979.
\newblock \href {http://www.jstor.org/stable/4615733} {A simple sequentially
  rejective multiple test procedure}.
\newblock \emph{Scandinavian Journal of Statistics}, 6(2):65--70.

\bibitem[{Hu et~al.(2022{\natexlab{a}})Hu, Shen, Wallis, Allen-Zhu, Li, Wang,
  Wang, and Chen}]{hu2022lora}
Edward~J Hu, Yelong Shen, Phillip Wallis, Zeyuan Allen-Zhu, Yuanzhi Li, Shean
  Wang, Lu~Wang, and Weizhu Chen. 2022{\natexlab{a}}.
\newblock \href {https://openreview.net/forum?id=nZeVKeeFYf9} {Lo{RA}: Low-rank
  adaptation of large language models}.
\newblock In \emph{International Conference on Learning Representations}.

\bibitem[{Hu et~al.(2022{\natexlab{b}})Hu, Wu, Gao, Bai, and
  Zhao}]{hu-etal-2022-improving-aspect}
Mengting Hu, Yike Wu, Hang Gao, Yinhao Bai, and Shiwan Zhao.
  2022{\natexlab{b}}.
\newblock \href {https://doi.org/10.18653/v1/2022.emnlp-main.538} {Improving
  aspect sentiment quad prediction via template-order data augmentation}.
\newblock In \emph{Proceedings of the 2022 Conference on Empirical Methods in
  Natural Language Processing}, pages 7889--7900, Abu Dhabi, United Arab
  Emirates. Association for Computational Linguistics.

\bibitem[{Jehangir et~al.(2023)Jehangir, Radhakrishnan, and
  Agarwal}]{JEHANGIR2023100017}
Basra Jehangir, Saravanan Radhakrishnan, and Rahul Agarwal. 2023.
\newblock \href {https://doi.org/10.1016/j.nlp.2023.100017} {A survey on named
  entity recognition — datasets, tools, and methodologies}.
\newblock \emph{Natural Language Processing Journal}, 3:100017.

\bibitem[{Jun and Lee(2025)}]{jun-lee-2025-dynamic}
Yonghyun Jun and Hwanhee Lee. 2025.
\newblock \href {https://doi.org/10.18653/v1/2025.acl-short.48} {Dynamic order
  template prediction for generative aspect-based sentiment analysis}.
\newblock In \emph{Proceedings of the 63rd Annual Meeting of the Association
  for Computational Linguistics (Volume 2: Short Papers)}, pages 614--626,
  Vienna, Austria. Association for Computational Linguistics.

\bibitem[{Kwon et~al.(2023)Kwon, Li, Zhuang, Sheng, Zheng, Yu, Gonzalez, Zhang,
  and Stoica}]{kwon2023efficient}
Woosuk Kwon, Zhuohan Li, Siyuan Zhuang, Ying Sheng, Lianmin Zheng, Cody~Hao Yu,
  Joseph Gonzalez, Hao Zhang, and Ion Stoica. 2023.
\newblock \href {https://doi.org/10.1145/3600006.3613165} {Efficient memory
  management for large language model serving with pagedattention}.
\newblock In \emph{Proceedings of the 29th Symposium on Operating Systems
  Principles}, SOSP '23, page 611–626, New York, NY, USA. Association for
  Computing Machinery.

\bibitem[{Lepagnol et~al.(2024)Lepagnol, Gerald, Ghannay, Servan, and
  Rosset}]{lepagnol2024small}
Pierre Lepagnol, Thomas Gerald, Sahar Ghannay, Christophe Servan, and Sophie
  Rosset. 2024.
\newblock \href {https://aclanthology.org/2024.lrec-main.1299/} {Small language
  models are good too: An empirical study of zero-shot classification}.
\newblock In \emph{Proceedings of the 2024 Joint International Conference on
  Computational Linguistics, Language Resources and Evaluation (LREC-COLING
  2024)}, pages 14923--14936, Torino, Italia. ELRA and ICCL.

\bibitem[{Li et~al.(2025)Li, Zhang, Wang, and Li}]{li2025aligningblack}
Shichen Li, Jiawei Zhang, Zhongqing Wang, and Peifeng Li. 2025.
\newblock \href {https://doi.org/10.18653/v1/2025.findings-emnlp.53} {Aligning
  black-box {LLM}s for aspect sentiment quad prediction}.
\newblock In \emph{Findings of the Association for Computational Linguistics:
  EMNLP 2025}, pages 1012--1025, Suzhou, China. Association for Computational
  Linguistics.

\bibitem[{M{\"u}nker et~al.(2025)M{\"u}nker, Kugler, and
  Rettinger}]{munker2025zero}
Simon M{\"u}nker, Kai Kugler, and Achim Rettinger. 2025.
\newblock \href {https://doi.org/10.18653/v1/2025.acl-srw.4} {Zero-shot
  prompt-based classification: topic labeling in times of foundation models in
  {G}erman tweets}.
\newblock In \emph{Proceedings of the 63rd Annual Meeting of the Association
  for Computational Linguistics (Volume 4: Student Research Workshop)}, pages
  53--63, Vienna, Austria. Association for Computational Linguistics.

\bibitem[{Ne{\v{s}}i{\'c} et~al.(2024)Ne{\v{s}}i{\'c}, Petalinkar,
  {\v{S}}kori{\'c}, Stankovi{\'c}, and Rujevi{\'c}}]{nevsic2024advancing}
Milica~Ikoni{\'c} Ne{\v{s}}i{\'c}, Sa{\v{s}}a Petalinkar, Mihailo
  {\v{S}}kori{\'c}, Ranka Stankovi{\'c}, and Biljana Rujevi{\'c}. 2024.
\newblock \href {https://aclanthology.org/2024.clib-1.5/} {Advancing sentiment
  analysis in {S}erbian literature: A zero and few{--}shot learning approach
  using the mistral model}.
\newblock In \emph{Proceedings of the Sixth International Conference on
  Computational Linguistics in Bulgaria (CLIB 2024)}, pages 58--70, Sofia,
  Bulgaria. Department of Computational Linguistics, Institute for Bulgarian
  Language, Bulgarian Academy of Sciences.

\bibitem[{OpenAI et~al.(2025)OpenAI, Agarwal, Ahmad, Ai, Altman, Applebaum,
  Arbus, Arora, Bai, Baker, Bao, Barak, Bennett, Bertao, Brett, Brevdo,
  Brockman, Bubeck, Chang, Chen, Chen, Cheung, Clark, Cook, Dukhan, Dvorak,
  Fives, Fomenko, Garipov, Georgiev, Glaese, Gogineni, Goucher, Gross, Guzman,
  Hallman, Hehir, Heidecke, Helyar, Hu, Huet, Huh, Jain, Johnson, Koch, Kofman,
  Kundel, Kwon, Kyrylov, Le, Leclerc, Lennon, Lessans, Lezcano-Casado, Li, Li,
  Lin, Liss, Lily, Liu, Liu, Lu, Lu, Martinovic, McCallum, McGrath, McKinney,
  McLaughlin, Mei, Mostovoy, Mu, Myles, Neitz, Nichol, Pachocki, Paino, Palmie,
  Pantuliano, Parascandolo, Park, Pathak, Paz, Peran, Pimenov, Pokrass, Proehl,
  Qiu, Raila, Raso, Ren, Richardson, Robinson, Rotsted, Salman, Sanjeev,
  Schwarzer, Sculley, Sikchi, Simon, Singhal, Song, Stuckey, Sun, Tillet,
  Toizer, Tsimpourlas, Vyas, Wallace, Wang, Wang, Watkins, Weil, Wendling,
  Whinnery, Whitney, Wong, Yang, Yang, Yasunaga, Ying, Zaremba, Zhan, Zhang,
  Zhang, Zhang, and Zhao}]{agarwal2025gpt}
OpenAI, Sandhini Agarwal, Lama Ahmad, Jason Ai, Sam Altman, Andy Applebaum,
  Edwin Arbus, Rahul~K. Arora, Yu~Bai, Bowen Baker, Haiming Bao, Boaz Barak,
  Ally Bennett, Tyler Bertao, Nivedita Brett, Eugene Brevdo, Greg Brockman,
  Sebastien Bubeck, Che Chang, and 107 others. 2025.
\newblock \href {https://arxiv.org/abs/2508.10925} {gpt-oss-120b \& gpt-oss-20b
  model card}.
\newblock \emph{Preprint}, arXiv:2508.10925.

\bibitem[{Pontiki et~al.(2016)Pontiki, Galanis, Papageorgiou, Androutsopoulos,
  Manandhar, AL-Smadi, Al-Ayyoub, Zhao, Qin, De~Clercq, Hoste, Apidianaki,
  Tannier, Loukachevitch, Kotelnikov, Bel, Jim{\'e}nez-Zafra, and
  Eryi{\u{g}}it}]{pontiki2016semeval}
Maria Pontiki, Dimitris Galanis, Haris Papageorgiou, Ion Androutsopoulos,
  Suresh Manandhar, Mohammad AL-Smadi, Mahmoud Al-Ayyoub, Yanyan Zhao, Bing
  Qin, Orph{\'e}e De~Clercq, V{\'e}ronique Hoste, Marianna Apidianaki, Xavier
  Tannier, Natalia Loukachevitch, Evgeniy Kotelnikov, Nuria Bel,
  Salud~Mar{\'i}a Jim{\'e}nez-Zafra, and G{\"u}l{\c{s}}en Eryi{\u{g}}it. 2016.
\newblock \href {https://doi.org/10.18653/v1/S16-1002} {{S}em{E}val-2016 task
  5: Aspect based sentiment analysis}.
\newblock In \emph{Proceedings of the 10th International Workshop on Semantic
  Evaluation ({S}em{E}val-2016)}, pages 19--30, San Diego, California.
  Association for Computational Linguistics.

\bibitem[{Pontiki et~al.(2015)Pontiki, Galanis, Papageorgiou, Manandhar, and
  Androutsopoulos}]{pontiki2015semeval}
Maria Pontiki, Dimitris Galanis, Haris Papageorgiou, Suresh Manandhar, and Ion
  Androutsopoulos. 2015.
\newblock \href {https://doi.org/10.18653/v1/S15-2082} {{S}em{E}val-2015 task
  12: Aspect based sentiment analysis}.
\newblock In \emph{Proceedings of the 9th International Workshop on Semantic
  Evaluation ({S}em{E}val 2015)}, pages 486--495, Denver, Colorado. Association
  for Computational Linguistics.

\bibitem[{Raffel et~al.(2020)Raffel, Shazeer, Roberts, Lee, Narang, Matena,
  Zhou, Li, and Liu}]{raffel2020exploring}
Colin Raffel, Noam Shazeer, Adam Roberts, Katherine Lee, Sharan Narang, Michael
  Matena, Yanqi Zhou, Wei Li, and Peter~J. Liu. 2020.
\newblock \href {https://dl.acm.org/doi/abs/10.5555/3455716.3455856} {Exploring
  the limits of transfer learning with a unified text-to-text transformer}.
\newblock \emph{J. Mach. Learn. Res.}, 21(1).

\bibitem[{{\v{S}}m{\'i}d et~al.(2024){\v{S}}m{\'i}d, Priban, and
  Kral}]{vsmid2024llama}
Jakub {\v{S}}m{\'i}d, Pavel Priban, and Pavel Kral. 2024.
\newblock \href {https://doi.org/10.18653/v1/2024.wassa-1.6} {{LL}a{MA}-based
  models for aspect-based sentiment analysis}.
\newblock In \emph{Proceedings of the 14th Workshop on Computational Approaches
  to Subjectivity, Sentiment, {\&} Social Media Analysis}, pages 63--70,
  Bangkok, Thailand. Association for Computational Linguistics.

\bibitem[{Su et~al.(2025)Su, Zhang, Wang, Wu, and Sha}]{su-etal-2025-unified}
Guixin Su, Yongcheng Zhang, Tongguan Wang, Mingmin Wu, and Ying Sha. 2025.
\newblock \href {https://aclanthology.org/2025.coling-main.269/} {Unified grid
  tagging scheme for aspect sentiment quad prediction}.
\newblock In \emph{Proceedings of the 31st International Conference on
  Computational Linguistics}, pages 3997--4010, Abu Dhabi, UAE. Association for
  Computational Linguistics.

\bibitem[{Ventirozos et~al.(2025)Ventirozos, Appleby, and
  Shardlow}]{ventirozos-etal-2025-aspect}
Filippos Ventirozos, Peter Appleby, and Matthew Shardlow. 2025.
\newblock \href {https://aclanthology.org/2025.ranlp-1.152/}
  {Aspect{--}sentiment quad prediction with distilled large language models}.
\newblock In \emph{Proceedings of the 15th International Conference on Recent
  Advances in Natural Language Processing - Natural Language Processing in the
  Generative AI Era}, pages 1309--1319, Varna, Bulgaria. INCOMA Ltd., Shoumen,
  Bulgaria.

\bibitem[{Wei et~al.(2022)Wei, Wang, Schuurmans, Bosma, Ichter, Xia, Chi, Le,
  and Zhou}]{10.5555/3600270.3602070}
Jason Wei, Xuezhi Wang, Dale Schuurmans, Maarten Bosma, Brian Ichter, Fei Xia,
  Ed~H. Chi, Quoc~V. Le, and Denny Zhou. 2022.
\newblock \href {https://dl.acm.org/doi/10.5555/3600270.3602070}
  {Chain-of-thought prompting elicits reasoning in large language models}.
\newblock In \emph{Proceedings of the 36th International Conference on Neural
  Information Processing Systems}, NIPS '22, Red Hook, NY, USA. Curran
  Associates Inc.

\bibitem[{Yadav and Bethard(2018)}]{yadav-bethard-2018-survey}
Vikas Yadav and Steven Bethard. 2018.
\newblock \href {https://aclanthology.org/C18-1182/} {A survey on recent
  advances in named entity recognition from deep learning models}.
\newblock In \emph{Proceedings of the 27th International Conference on
  Computational Linguistics}, pages 2145--2158, Santa Fe, New Mexico, USA.
  Association for Computational Linguistics.

\bibitem[{Yang et~al.(2025)Yang, Kautz, and Hatamizadeh}]{yang2025gated}
Songlin Yang, Jan Kautz, and Ali Hatamizadeh. 2025.
\newblock \href {https://openreview.net/forum?id=r8H7xhYPwz} {Gated delta
  networks: Improving mamba2 with delta rule}.
\newblock In \emph{The Thirteenth International Conference on Learning
  Representations}.

\bibitem[{Zhang et~al.(2021{\natexlab{a}})Zhang, Deng, Li, Yuan, Bing, and
  Lam}]{zhang2021aspect}
Wenxuan Zhang, Yang Deng, Xin Li, Yifei Yuan, Lidong Bing, and Wai Lam.
  2021{\natexlab{a}}.
\newblock \href {https://doi.org/10.18653/v1/2021.emnlp-main.726} {Aspect
  sentiment quad prediction as paraphrase generation}.
\newblock In \emph{Proceedings of the 2021 Conference on Empirical Methods in
  Natural Language Processing}, pages 9209--9219, Online and Punta Cana,
  Dominican Republic. Association for Computational Linguistics.

\bibitem[{Zhang et~al.(2024)Zhang, Deng, Liu, Pan, and
  Bing}]{zhang2024sentiment}
Wenxuan Zhang, Yue Deng, Bing Liu, Sinno Pan, and Lidong Bing. 2024.
\newblock \href {https://doi.org/10.18653/v1/2024.findings-naacl.246}
  {Sentiment analysis in the era of large language models: A reality check}.
\newblock In \emph{Findings of the Association for Computational Linguistics:
  NAACL 2024}, pages 3881--3906, Mexico City, Mexico. Association for
  Computational Linguistics.

\bibitem[{Zhang et~al.(2021{\natexlab{b}})Zhang, Li, Deng, Bing, and
  Lam}]{zhang-etal-2021-towards-generative}
Wenxuan Zhang, Xin Li, Yang Deng, Lidong Bing, and Wai Lam. 2021{\natexlab{b}}.
\newblock \href {https://doi.org/10.18653/v1/2021.acl-short.64} {Towards
  generative aspect-based sentiment analysis}.
\newblock In \emph{Proceedings of the 59th Annual Meeting of the Association
  for Computational Linguistics and the 11th International Joint Conference on
  Natural Language Processing (Volume 2: Short Papers)}, pages 504--510,
  Online. Association for Computational Linguistics.

\bibitem[{Zhang et~al.(2023)Zhang, Li, Deng, Bing, and Lam}]{zhang2022survey}
Wenxuan Zhang, Xin Li, Yang Deng, Lidong Bing, and Wai Lam. 2023.
\newblock \href {https://doi.org/10.1109/TKDE.2022.3230975} {{ A Survey on
  Aspect-Based Sentiment Analysis: Tasks, Methods, and Challenges }}.
\newblock \emph{IEEE Transactions on Knowledge \& Data Engineering},
  35(11):11019--11038.

\end{thebibliography}

\appendix

\renewcommand{\dbltopfraction}{0.9}
\renewcommand{\textfraction}{0.07}
\renewcommand{\dblfloatpagefraction}{0.7}

\makeatletter
\setlength{\@fptop}{0pt}
\setlength{\@fpsep}{\floatsep}
\setlength{\@fpbot}{0pt plus 1fil}
\setlength{\@dblfptop}{0pt}
\setlength{\@dblfpsep}{\dblfloatsep}
\setlength{\@dblfpbot}{0pt plus 1fil}
\makeatother

\section{Task Examples}\label{appendix:tuples-examples}

Table~\ref{tab:task-examples} shows annotated example sentences with their TASD triplets and ASQP quadruplets.

\begin{table*}[t]
\centering
\small
\resizebox{\textwidth}{!}{%
\begin{tabular}{@{}lcccc|lcccc@{}}
\toprule
\multicolumn{5}{c|}{\textit{``The wine list is excellent, but the service was slow.''}} &
\multicolumn{5}{c}{\textit{``Beautiful experience.''}} \\
\midrule
\textbf{} & \textbf{\textit{a}} & \textbf{\textit{c}} & \textbf{\textit{o}} & \textbf{\textit{p}} &
\textbf{} & \textbf{\textit{a}} & \textbf{\textit{c}} & \textbf{\textit{o}} & \textbf{\textit{p}} \\
\midrule
\multirow{2}{*}{\textbf{TASD}}
  & \textit{wine list} & drinks\#style\_options& --               & \colorbox{green!20}{positive}
  & \multirow{2}{*}{\textbf{TASD}}
  & \textit{NULL}      & restaurant\#general & --           & \colorbox{green!20}{positive} \\
  & \textit{service}   & service\#general & --               & \colorbox{red!20}{negative}
  &                    &                  &                  &                   & \\
\midrule
\multirow{2}{*}{\textbf{\textsc{ASQP}}}
  & \textit{wine list} & drinks\#style\_options& \textit{excellent} & \colorbox{green!20}{positive}
  & \multirow{2}{*}{\textbf{\textsc{ASQP}}}
  & \textit{NULL}      & restaurant\#general & \textit{Beautiful} & \colorbox{green!20}{positive} \\
  & \textit{service}   & service\#general & \textit{slow}      & \colorbox{red!20}{negative}
  &                    &                  &                    &                   & \\
\bottomrule
\end{tabular}
}
\caption{\textbf{Examples of TASD and ASQP extraction tasks.} Both tasks extract structured sentiment information as tuples (\textit{a}: aspect term, \textit{c}: aspect category, \textit{o}: opinion term, \textit{p}: polarity). NULL indicates implicit aspects where no specific aspect target (aspect term) is mentioned in the text.}
\label{tab:task-examples}

\vspace{0.8em}
{%
\scriptsize
\setlength{\tabcolsep}{3pt}
\renewcommand{\arraystretch}{0.8} %
\resizebox{\textwidth}{!}{%
\begin{tabular}{@{}l cccc cccc cccc cccc@{}}
\toprule
\multirow{3}{*}{\textbf{Dataset}} & \multicolumn{8}{c}{\textbf{TASD}} & \multicolumn{8}{c}{\textbf{ASQP}} \\
\cmidrule(lr){2-9} \cmidrule(lr){10-17}
& \multicolumn{4}{c}{Train} & \multicolumn{4}{c}{Test} & \multicolumn{4}{c}{Train} & \multicolumn{4}{c}{Test} \\
\cmidrule(lr){2-5} \cmidrule(lr){6-9} \cmidrule(lr){10-13} \cmidrule(lr){14-17}
& \# Sents & \# Tuples & Uniq. Cat. & Avg Len & \# Sents & \# Tuples & Uniq. Cat. & Avg Len & \# Sents & \# Tuples & Uniq. Cat. & Avg Len & \# Sents & \# Tuples & Uniq. Cat. & Avg Len \\
\midrule
\textsc{Rest15}     & 1,120 & 1,654 & 13 & 12.3 & 582 & 845 & 12 & 13.5 & 834 & 1,354 & 13 & 13.8 & 537 & 795 & 12 & 15.2 \\
\textsc{Rest16}     & 1,708 & 2,507 & 12 & 12.7 & 587 & 859 & 12 & 12.7 & 1,264 & 1,989 & 12 & 14.5 & 544 & 799 & 12 & 14.5 \\
\textsc{FlightABSA} & 1,351 & 1,900 & 13 & 13.8 & 387 & 529 & 13 & 14.0 & 1,351 & 2,116 & 13 & 13.8 & 387 & 590 & 13 & 14.0 \\
\textsc{Coursera}   & 1,400 & 1,711 & 26 & 15.8 & 400 & 488 & 24 & 15.5 & 1,400 & 1,802 & 26 & 15.8 & 400 & 502 & 24 & 15.5 \\
\textsc{Hotels}     & 1,400 & 2,285 & 33 & 16.1 & 400 & 629 & 30 & 16.1 & 1,400 & 2,677 & 33 & 16.1 & 400 & 721 & 30 & 16.1 \\
\bottomrule
\end{tabular}

}
\caption{\textbf{Detailed dataset statistics.} Comparison of dataset sizes, tuple counts, unique category diversity, and average number of tokens per sentence for the TASD and ASQP tasks.}
\label{tab:detailed_stats}
}
\end{table*}

\section{Datasets}
\subsection{Overview}\label{appendix:datasets}

Table~\ref{tab:detailed_stats} reports the number of sentences and tuples, the number of unique aspect categories, and the average sentence length of the train and test splits of each dataset.

\subsection{Datasets: Aspect Category and Sentiment Distributions}\label{sec:distributions}

\begin{figure*}[p]
    \centering
    \includegraphics[width=\textwidth]{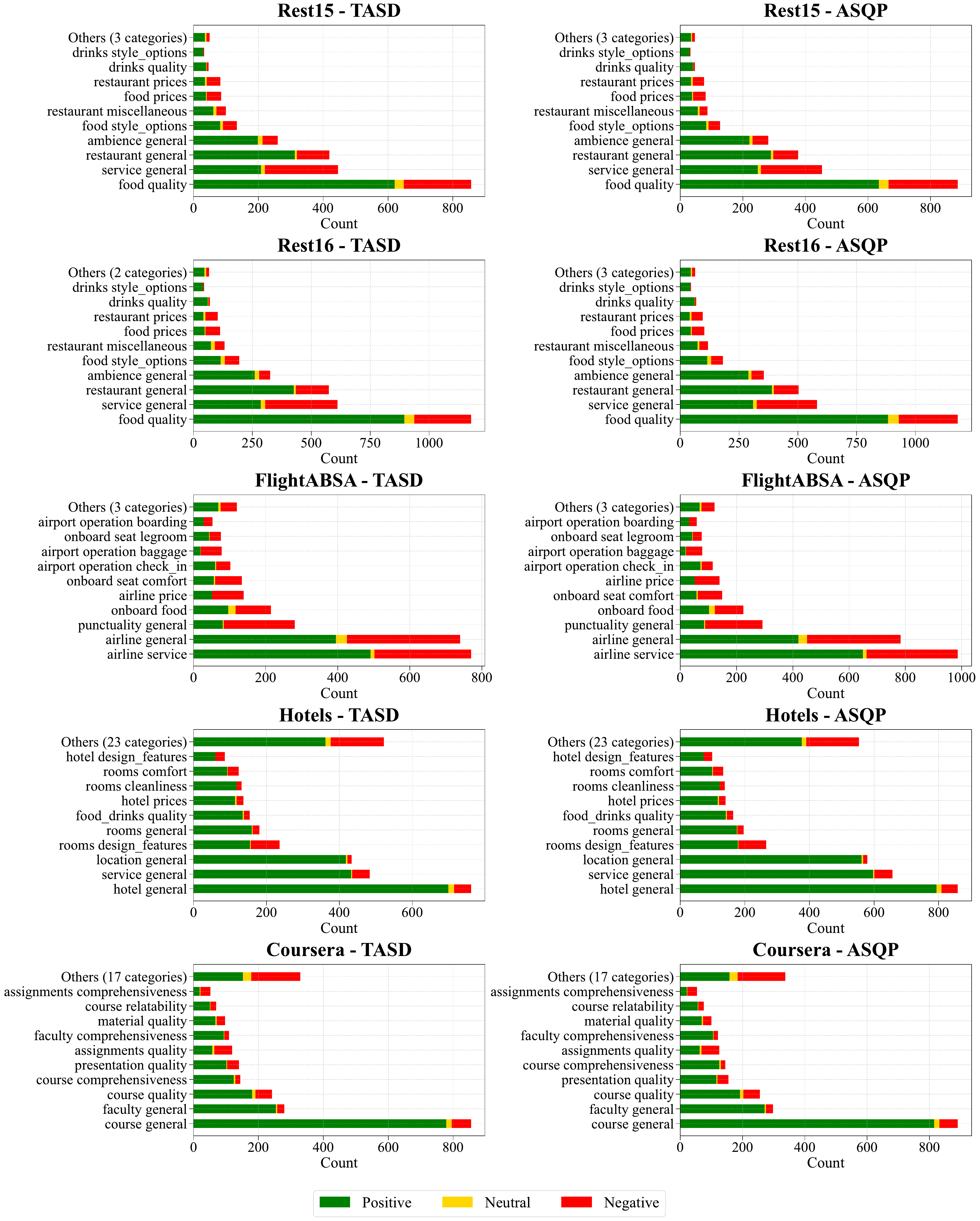}
    \caption{\textbf{Distribution of aspect categories and sentiments across datasets (train, test, and validation) for TASD and ASQP tasks.} Each subplot shows the top 10 aspect categories (sorted by total frequency), with stacked bars representing positive (green), neutral (yellow), and negative (red) sentiments. The `Others' category aggregates the remaining aspects. Results are aggregated across five datasets: Rest15 \citep{zhang2021aspect, pontiki2015semeval}, Rest16 \citep{zhang2021aspect, pontiki2016semeval}, FlightABSA \citep{hellwig2025still}, Coursera \citep{chebolu2024oats}, and Hotels \citep{chebolu2024oats}. The distributions of polarities and aspect categories are imbalanced and differ across datasets.}
    \label{fig:ac_pol_dist}
\end{figure*}

Figure~\ref{fig:ac_pol_dist} shows the distribution of the ten most frequent aspect categories and their sentiment polarities for each dataset and task.

\section{Fine-tuned Approaches}\label{appendix:fine-tuning}

First, we considered Multi-view Prompting (\textbf{MvP}) \citep{gou2023mvp}, the best-performing fine-tuned approach based on the T5-base encoder-decoder model. Furthermore, we evaluated \textbf{Paraphrase}, which, unlike MvP, considers only a single permutation of sentiment elements for each example. Paraphrase represents tuples as natural language text. For instance, the tuple (a: ``wine list'', c: ``drinks style'', p: ``positive'', o: ``excellent'') is transformed into \textit{drinks style is great because wine list is excellent}, where polarities are mapped to adjectives (``positive'' $\rightarrow$ ``great'', ``negative'' $\rightarrow$ ``bad'', ``neutral'' $\rightarrow$ ``okay''), and multiple tuples are separated by \texttt{[SSEP]} tokens. We took performance scores for MvP and Paraphrase on the Rest15 and Rest16 datasets from their respective original papers \citep{gou2023mvp,zhang2021aspect}.\footnote{Neither \citet{gou2023mvp} nor \citet{zhang2021aspect} reported macro-averaged F1 scores.} Scores for the remaining three datasets are adopted from \citet{hellwig2025still}, who employed the same hyperparameters as reported by \citet{gou2023mvp} and \citet{zhang2021aspect}: a batch size of 16, a learning rate of 3e-4 (1e-4 for MvP), and 20 training epochs.

Finally, as a comparative baseline, we \textbf{fine-tuned Gemma 4 (31B)}\footnote{\hficon\url{https://hf.co/unsloth/gemma-4-31B-it-unsloth-bnb-4bit}} using the unsloth framework \citep{unsloth}. Following \citet{vsmid2024llama}, we employed a learning rate of 2e-4 and trained for 5 epochs using Low-Rank Adaptation (LoRA) \citep{hu2022lora}. We applied LoRA adapters with rank $r=64$ and scaling factor $\alpha=16$ to both attention and MLP modules, while keeping all other layers frozen. The LoRA dropout rate was set to 0. The maximum sequence length was limited to 512 tokens. For inference, we used vLLM (v0.20.0) with the trained LoRA adapters, prefix caching, and temperature set to 0, allowing a fair comparison with LLM-MvP. Table~\ref{tab:asqp_tasd_results_related} shows that our fine-tuned Gemma 4 (31B) outperforms all existing baselines and sets new SOTA scores on all four benchmarks.

\begin{table}[t]
\centering
\small
\resizebox{\columnwidth}{!}{%
\begin{tabular}{l cccc}
\toprule
\multirow{2}{*}{\textbf{Approach Name}} & \multicolumn{2}{c}{\textbf{ASQP}} & \multicolumn{2}{c}{\textbf{TASD}} \\
\cmidrule(lr){2-3} \cmidrule(lr){4-5}
& \textbf{Rest15} & \textbf{Rest16} & \textbf{Rest15} & \textbf{Rest16} \\
\midrule
GAS & 45.98 & 56.04 & 60.63 & 68.3 \\
Paraphrase & 46.93 & 57.93 & 63.06 & 71.97 \\
MvP & 51.04 & 60.39 & 64.53 & 72.76 \\
UGTS & 52.59 & 65.10 & -- & -- \\
E2TP & 51.94 & 62.57 & 65.80 & 73.17 \\
PGSO & -- & -- & 65.40 & 72.74 \\
\midrule
Orca 2 (13B) FT & 52.29 & 60.82 & 70.49 & 78.82 \\
Orca 2 (7B) FT & 51.50 & 58.63 & 69.74 & 76.10 \\
\midrule
LLM-MvP (100-shot) (Ours) & \val{f1.rest15.asqp.mvp.100} & \val{f1.rest16.asqp.mvp.100} & \val{f1.rest15.tasd.mvp.100} & \val{f1.rest16.tasd.mvp.100} \\
\textbf{Gemma 4 (31B) FT} (Ours) & \textbf{\val{f1.rest15.asqp.ft-llm}} & \textbf{\val{f1.rest16.asqp.ft-llm}} & \textbf{\val{f1.rest15.tasd.ft-llm}} & \textbf{\val{f1.rest16.tasd.ft-llm}} \\
\bottomrule
\end{tabular}
}
\caption{\textbf{Performance comparison ($F_1$ score) of fine-tuned approaches across the ASQP and TASD tasks on the Rest15 and Rest16 datasets.} \textbf{Bold} values indicate the best performance per task and dataset. Scores of prior approaches are taken from the original publications: GAS \citep{zhang-etal-2021-towards-generative}, Paraphrase \citep{zhang2021aspect}, MvP \citep{gou2023mvp}, UGTS \citep{su-etal-2025-unified}, E2TP \citep{GHIASVANDMOHAMMADKHANI2025107847}, PGSO \citep{DONG2025125933}, and Orca 2 \citep{vsmid2024llama}. Missing values indicate results not reported in these publications.}
\label{tab:asqp_tasd_results_related}
\end{table}

\section{Significance Tests}\label{sec:significance}

\subsection{Task Performance}

To assess statistical significance, we conducted pairwise comparisons between all seven methods across different shot settings for both TASD and ASQP tasks. These seven methods are single-order prompting, reasoning-based prompting, self-consistency prompting, and LLM-MvP, as well as the fine-tuned approaches Paraphrase, MvP, and the LoRA-adapted LLM. Since each method was evaluated on five datasets, five independent observations per method were available. This comparison was conducted for four shot configurations (0, 10, 50, and 100 shots) to cover a broad set of low-resource scenarios.

For each of these four configurations within a given subtask (TASD or ASQP), we first performed an omnibus test, either ANOVA \citep{Fisher1992} when all groups met normality assumptions (Shapiro--Wilk test, $\alpha = 0.05$) or Kruskal--Wallis otherwise, to determine whether significant differences existed among the seven methods. If the omnibus test indicated significance, we proceeded with pairwise comparisons using either Welch's $t$-test (for normally distributed data) or the Mann--Whitney U test (otherwise). To control for multiple comparisons, we applied the Bonferroni--Holm correction \citep{1979holmbonferroni} separately for each task, adjusting for \val{sig.n.comparisons} pairwise comparisons (\val{sig.n.pairs} method pairs $\times$ 4 shot configurations) per task.

\subsection{Energy Consumption}

For the energy efficiency analysis, we conducted statistical comparisons between the methods across both TASD and ASQP tasks. Each method was evaluated using five datasets, providing five independent observations per configuration. The evaluation included seven approaches: four batch-processed prompting approaches (single-order, reasoning, self-consistency, and LLM-MvP), single-order and self-consistency prompting without prefix caching, and the fine-tuned LLM. We analyzed four shot configurations (0, 10, 50, and 100 shots). 

The statistical testing followed the same protocol as for task performance. For each configuration, an omnibus test (ANOVA or Kruskal--Wallis) was performed, followed by post-hoc pairwise comparisons where significant results were found. We applied the Bonferroni--Holm correction separately for each task, adjusting for \val{sig.n.comparisons} pairwise comparisons (\val{sig.n.pairs} method pairs $\times$ 4 shot configurations) per task. 

As shown in Table~\ref{tab:significance_energy}, we observed no statistically significant difference in energy consumption between single-order prompting (SO) and the fine-tuned LLM baseline (FT) across any shot configuration (e.g., $p=\val{p.energy.tasd.100.so-ft}$ at 100 shots). While reasoning-based prompting also showed no significant difference from FT in lower-shot settings, it consumed significantly more energy than FT for TASD at 100 shots ($p=\val{p.energy.tasd.100.reasoning-ft}$).

With prefix caching, LLM-MvP benefits most: for TASD, despite its multi-view nature, it is significantly more energy-efficient than non-cached self-consistency (SC-nc) across all shot settings and than non-cached single-order prompting (SO-nc) from 10 shots onwards ($p < .001$ at 50 and 100 shots); at 0 shots, the difference from SO-nc is not significant ($p=\val{p.energy.tasd.0.sonc-mvp}$). In contrast, for the ASQP task, significant efficiency differences were primarily confined to the 0-shot setting and largely vanished as the number of shots increased ($p=\val{p.energy.asqp.ge10.mode}$ for most pairs at 10+ shots).

\begin{table*}[p]
\centering
\small
\resizebox{\textwidth}{!}{%
\begin{tabular}{@{}l|ccccccc|ccccccc@{}}
\toprule
& \multicolumn{7}{c|}{\textbf{\textsc{TASD} --- 0 shots}} & \multicolumn{7}{c}{\textbf{\textsc{ASQP} --- 0 shots}} \\
& \multicolumn{7}{c|}{ANOVA: $p < .001$} & \multicolumn{7}{c}{ANOVA: $p < .001$} \\
\midrule
& \textbf{SO} & \textbf{\scalebox{0.8}{SO (nc)}} & \textbf{R} & \textbf{SC} & \textbf{\scalebox{0.8}{SC (nc)}} & \textbf{\scalebox{0.85}{LLM-MvP}} & \textbf{FT} & \textbf{SO} & \textbf{\scalebox{0.8}{SO (nc)}} & \textbf{R} & \textbf{SC} & \textbf{\scalebox{0.8}{SC (nc)}} & \textbf{\scalebox{0.85}{LLM-MvP}} & \textbf{FT} \\
\midrule
\textbf{Single-order}          & - & 0.018* & 0.178 & 0.011* & 0.012* & 0.022* & 0.081 & - & 0.007* & 0.042* & <.001* & 0.005* & 0.007* & 0.524 \\
\textbf{Single-order (nc)}     & - & - & 0.022* & 0.102 & 0.015* & 0.445 & 0.022* & - & - & 0.008* & 0.188 & 0.007* & 0.009* & 0.002* \\
\textbf{Reasoning}             & - & - & - & 0.014* & 0.012* & 0.023* & 0.426 & - & - & - & <.001* & 0.005* & 0.007* & 1.000 \\
\textbf{Self-consistency}      & - & - & - & - & 0.014* & 0.079 & 0.015* & - & - & - & - & 0.007* & 0.009* & 0.002* \\
\textbf{Self-consistency (nc)} & - & - & - & - & - & 0.015* & 0.012* & - & - & - & - & - & 0.787 & 0.005* \\
\textbf{LLM-MvP}              & - & - & - & - & - & - & 0.023* & - & - & - & - & - & - & 0.007* \\
\textbf{LLM (FT)}             & - & - & - & - & - & - & - & - & - & - & - & - & - & - \\
\bottomrule
\end{tabular}
}

\vspace{0.3cm}
\resizebox{\textwidth}{!}{%
\begin{tabular}{@{}l|ccccccc|ccccccc@{}}
\toprule
& \multicolumn{7}{c|}{\textbf{\textsc{TASD} --- 10 shots}} & \multicolumn{7}{c}{\textbf{\textsc{ASQP} --- 10 shots}} \\
& \multicolumn{7}{c|}{ANOVA: $p < .001$} & \multicolumn{7}{c}{Kruskal-Wallis: $p < .001$} \\
\midrule
& \textbf{SO} & \textbf{\scalebox{0.8}{SO (nc)}} & \textbf{R} & \textbf{SC} & \textbf{\scalebox{0.8}{SC (nc)}} & \textbf{\scalebox{0.85}{LLM-MvP}} & \textbf{FT} & \textbf{SO} & \textbf{\scalebox{0.8}{SO (nc)}} & \textbf{R} & \textbf{SC} & \textbf{\scalebox{0.8}{SC (nc)}} & \textbf{\scalebox{0.85}{LLM-MvP}} & \textbf{FT} \\
\midrule
\textbf{Single-order}          & - & <.001* & 0.011* & <.001* & <.001* & 0.022* & 0.063 & - & 0.524 & 0.524 & 0.524 & 0.524 & 0.524 & 0.524 \\
\textbf{Single-order (nc)}     & - & - & <.001* & <.001* & <.001* & 0.016* & <.001* & - & - & 0.524 & 0.524 & 0.524 & 0.524 & 0.524 \\
\textbf{Reasoning}             & - & - & - & 0.004* & <.001* & 0.028* & 0.616 & - & - & - & 0.524 & 0.524 & 0.524 & 1.000 \\
\textbf{Self-consistency}      & - & - & - & - & <.001* & 0.081 & 0.004* & - & - & - & - & 0.524 & 0.524 & 0.524 \\
\textbf{Self-consistency (nc)} & - & - & - & - & - & <.001* & <.001* & - & - & - & - & - & 0.524 & 0.524 \\
\textbf{LLM-MvP}              & - & - & - & - & - & - & 0.023* & - & - & - & - & - & - & 0.524 \\
\textbf{LLM (FT)}             & - & - & - & - & - & - & - & - & - & - & - & - & - & - \\
\bottomrule
\end{tabular}
}

\vspace{0.3cm}
\resizebox{\textwidth}{!}{%
\begin{tabular}{@{}l|ccccccc|ccccccc@{}}
\toprule
& \multicolumn{7}{c|}{\textbf{\textsc{TASD} --- 50 shots}} & \multicolumn{7}{c}{\textbf{\textsc{ASQP} --- 50 shots}} \\
& \multicolumn{7}{c|}{ANOVA: $p < .001$} & \multicolumn{7}{c}{Kruskal-Wallis: $p < .001$} \\
\midrule
& \textbf{SO} & \textbf{\scalebox{0.8}{SO (nc)}} & \textbf{R} & \textbf{SC} & \textbf{\scalebox{0.8}{SC (nc)}} & \textbf{\scalebox{0.85}{LLM-MvP}} & \textbf{FT} & \textbf{SO} & \textbf{\scalebox{0.8}{SO (nc)}} & \textbf{R} & \textbf{SC} & \textbf{\scalebox{0.8}{SC (nc)}} & \textbf{\scalebox{0.85}{LLM-MvP}} & \textbf{FT} \\
\midrule
\textbf{Single-order}          & - & <.001* & 0.022* & <.001* & <.001* & 0.022* & 0.359 & - & 0.524 & 0.524 & 0.524 & 0.524 & 0.524 & 0.571 \\
\textbf{Single-order (nc)}     & - & - & <.001* & <.001* & <.001* & <.001* & <.001* & - & - & 0.524 & 0.524 & 0.524 & 0.889 & 0.524 \\
\textbf{Reasoning}             & - & - & - & 0.004* & <.001* & 0.029* & 0.158 & - & - & - & 0.524 & 0.524 & 0.524 & 0.524 \\
\textbf{Self-consistency}      & - & - & - & - & <.001* & 0.079 & <.001* & - & - & - & - & 0.524 & 0.524 & 0.524 \\
\textbf{Self-consistency (nc)} & - & - & - & - & - & <.001* & <.001* & - & - & - & - & - & 0.524 & 0.524 \\
\textbf{LLM-MvP}              & - & - & - & - & - & - & 0.023* & - & - & - & - & - & - & 0.524 \\
\textbf{LLM (FT)}             & - & - & - & - & - & - & - & - & - & - & - & - & - & - \\
\bottomrule
\end{tabular}
}

\vspace{0.5cm}
\resizebox{\textwidth}{!}{%
\begin{tabular}{@{}l|ccccccc|ccccccc@{}}
\toprule
& \multicolumn{7}{c|}{\textbf{\textsc{TASD} --- 100 shots}} & \multicolumn{7}{c}{\textbf{\textsc{ASQP} --- 100 shots}} \\
& \multicolumn{7}{c|}{ANOVA: $p < .001$} & \multicolumn{7}{c}{Kruskal-Wallis: $p < .001$} \\
\midrule
& \textbf{SO} & \textbf{\scalebox{0.8}{SO (nc)}} & \textbf{R} & \textbf{SC} & \textbf{\scalebox{0.8}{SC (nc)}} & \textbf{\scalebox{0.85}{LLM-MvP}} & \textbf{FT} & \textbf{SO} & \textbf{\scalebox{0.8}{SO (nc)}} & \textbf{R} & \textbf{SC} & \textbf{\scalebox{0.8}{SC (nc)}} & \textbf{\scalebox{0.85}{LLM-MvP}} & \textbf{FT} \\
\midrule
\textbf{Single-order}          & - & <.001* & 0.011* & <.001* & <.001* & 0.008* & 0.426 & - & 0.524 & 0.524 & 0.524 & 0.524 & 0.524 & 1.000 \\
\textbf{Single-order (nc)}     & - & - & <.001* & <.001* & <.001* & <.001* & <.001* & - & - & 0.524 & 0.524 & 0.524 & 0.524 & 0.524 \\
\textbf{Reasoning}             & - & - & - & 0.005* & <.001* & 0.011* & 0.036* & - & - & - & 0.524 & 0.524 & 0.524 & 0.524 \\
\textbf{Self-consistency}      & - & - & - & - & <.001* & 0.022* & <.001* & - & - & - & - & 0.524 & 0.524 & 0.524 \\
\textbf{Self-consistency (nc)} & - & - & - & - & - & <.001* & <.001* & - & - & - & - & - & 0.524 & 0.524 \\
\textbf{LLM-MvP}              & - & - & - & - & - & - & 0.006* & - & - & - & - & - & - & 0.524 \\
\textbf{LLM (FT)}             & - & - & - & - & - & - & - & - & - & - & - & - & - & - \\
\bottomrule
\end{tabular}
}

\caption{\textbf{Pairwise statistical significance testing results with Bonferroni--Holm correction for energy efficiency (milliwatt-hours, mWh) comparisons.} Corrected $p$-values are shown for each method pair. Values marked with * indicate statistical significance at $\alpha = 0.05$. The omnibus test (Kruskal--Wallis or ANOVA) $p$-value is displayed in each subtitle. Diagonal entries represent self-comparisons and are marked with hyphens.}
\label{tab:significance_energy}
\end{table*}

\section{Few-Shot Performance Visualization: LLM-based Prompting Methods vs. Fine-Tuning Baselines}\label{sec:fs-performance}

\begin{figure*}[p]
    \centering
    \includegraphics[width=0.95\textwidth]{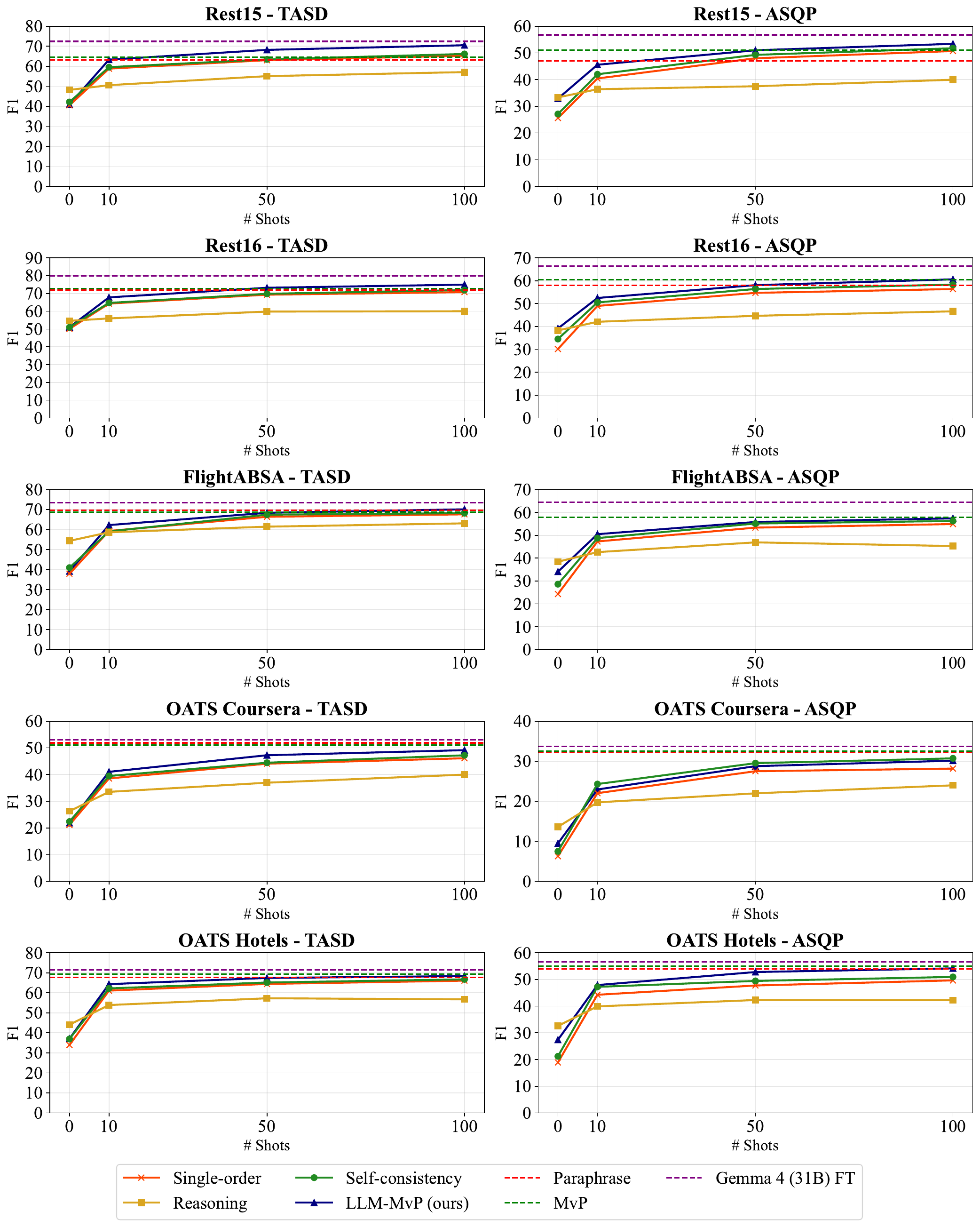}
    \caption{\textbf{LLM-MvP performance comparison against baselines.} The plots show F1 scores for Target Aspect Sentiment Detection (TASD) and Aspect Sentiment Quad Prediction (ASQP) tasks across varying numbers of in-context examples (0, 10, 50, and 100 shots). LLM-MvP (shown in navy blue with triangles) outperformed the other prompting strategies (Self-consistency, Reasoning, Single-order) in most few-shot configurations and approached or exceeded the T5-based fine-tuning baselines at higher shot counts. Horizontal dashed lines represent three fine-tuning baselines: Paraphrase \citep{zhang2021aspect}, MvP \citep{gou2023mvp}, and Gemma 4 (31B) LoRA fine-tuning.}
    \label{fig:fs-performance}
\end{figure*}

Figure~\ref{fig:fs-performance} plots the $F_1$ scores of all prompting methods against the number of few-shot examples, with the fine-tuned baselines shown as horizontal lines.

\section{Task Performance}\label{sec:task-performance}

Tables~\ref{tab:results-tasd-precision}, \ref{tab:results-tasd-recall}, and~\ref{tab:results-tasd-f1-macro} report precision, recall, and macro-averaged $F_1$ for all methods, datasets, and shot settings.

\begin{table*}[p]
\centering
\small
\setlength{\tabcolsep}{1.0pt}
\renewcommand{\arraystretch}{0.8} 
\resizebox{\textwidth}{!}{%
\begin{tabular}{@{}ll @{\hskip 6pt} rrrr @{\hskip 3pt} r @{\hskip 3pt} rrr @{\hspace{10pt}} rrrr @{\hskip 3pt} r @{\hskip 3pt} rrr@{}}
\toprule
\multirow{3}{*}{\textbf{Dataset}} & \multirow{3}{*}{\textbf{Approach}} & \multicolumn{8}{c}{\textbf{TASD} (Precision)} & \multicolumn{8}{c}{\textbf{ASQP} (Precision)} \\
\cmidrule(lr){3-10}\cmidrule(lr){11-18}
& & \multicolumn{4}{c}{\textit{\# Shots}} & \multirow{2}{*}{\textit{Avg.}} & \multicolumn{3}{c}{\textit{FT Baselines}} & \multicolumn{4}{c}{\textit{\# Shots}} & \multirow{2}{*}{\textit{Avg.}} & \multicolumn{3}{c}{\textit{FT Baselines}} \\
\cmidrule(lr){3-6}\cmidrule(lr){8-10}\cmidrule(lr){11-14}\cmidrule(lr){16-18}
& & \textbf{0} & \textbf{10} & \textbf{50} & \textbf{100} & & Para.$^{\star}$ & MvP$^{\star}$ & LLM$^{\star}$ & \textbf{0} & \textbf{10} & \textbf{50} & \textbf{100} & & Para.$^{\star}$ & MvP$^{\star}$ & LLM$^{\star}$ \\
\midrule
\multirow{4}{*}{\textsc{Rest15}} & Single-order & 39.90 & 60.95 & 65.53 & 67.58 & 58.49 & \multicolumn{1}{c}{\multirow{4}{*}{-}} & \multicolumn{1}{c}{\multirow{4}{*}{-}} & \multicolumn{1}{c}{\multirow{4}{*}{\underline{74.11}}} & 23.53 & 39.42 & 47.54 & 50.42 & 40.23 & \multicolumn{1}{c}{\multirow{4}{*}{46.16}} & \multicolumn{1}{c}{\multirow{4}{*}{-}} & \multicolumn{1}{c}{\multirow{4}{*}{\underline{\textbf{56.78}}}} \\
& Reasoning & \textbf{46.23} & 47.68 & 52.61 & 54.43 & 50.24 & & & & 31.39 & 33.84 & 35.53 & 38.03 & 34.70 & & & \\
& Self-consistency & 44.95 & 63.76 & 67.60 & 70.19 & 61.63 & & & & 29.42 & 44.04 & \textbf{51.88} & \textbf{54.50} & 44.96 & & & \\
& LLM-MvP (ours) & 42.89 & \textbf{66.81} & \textbf{72.17} & \textbf{74.09} & \textbf{63.99} & & & & \textbf{31.81} & \textbf{44.82} & 51.45 & 53.94 & \textbf{45.50} & & & \\
\midrule
\multirow{4}{*}{\textsc{Rest16}} & Single-order & 49.16 & 66.59 & 70.78 & 71.96 & 64.62 & \multicolumn{1}{c}{\multirow{4}{*}{-}} & \multicolumn{1}{c}{\multirow{4}{*}{-}} & \multicolumn{1}{c}{\multirow{4}{*}{\underline{81.00}}} & 27.34 & 46.69 & 53.69 & 55.22 & 45.74 & \multicolumn{1}{c}{\multirow{4}{*}{56.63}} & \multicolumn{1}{c}{\multirow{4}{*}{-}} & \multicolumn{1}{c}{\multirow{4}{*}{\underline{\textbf{64.88}}}} \\
& Reasoning & 53.53 & 53.19 & 57.39 & 57.66 & 55.44 & & & & 36.48 & 39.50 & 42.08 & 44.18 & 40.56 & & & \\
& Self-consistency & \textbf{53.56} & 68.71 & 72.78 & 74.30 & 67.34 & & & & 36.49 & \textbf{51.65} & \textbf{58.45} & \textbf{60.16} & \textbf{51.69} & & & \\
& LLM-MvP (ours) & 52.14 & \textbf{71.02} & \textbf{75.60} & \textbf{76.92} & \textbf{68.92} & & & & \textbf{37.27} & 51.26 & 57.65 & 59.79 & 51.49 & & & \\
\midrule
\multirow{4}{*}{\textsc{FlightABSA}} & Single-order & 32.98 & 59.90 & 67.31 & 68.21 & 57.10 & \multicolumn{1}{c}{\multirow{4}{*}{70.22}} & \multicolumn{1}{c}{\multirow{4}{*}{67.84}} & \multicolumn{1}{c}{\multirow{4}{*}{\underline{\textbf{72.32}}}} & 22.59 & 45.91 & 52.50 & 53.73 & 43.68 & \multicolumn{1}{c}{\multirow{4}{*}{57.37}} & \multicolumn{1}{c}{\multirow{4}{*}{56.09}} & \multicolumn{1}{c}{\multirow{4}{*}{\underline{\textbf{64.02}}}} \\
& Reasoning & \textbf{50.29} & 53.80 & 55.70 & 57.45 & 54.31 & & & & \textbf{36.45} & 39.51 & 43.73 & 42.27 & 40.49 & & & \\
& Self-consistency & 39.78 & 61.54 & 70.16 & 70.36 & 60.46 & & & & 32.36 & \textbf{50.73} & \textbf{57.44} & \textbf{58.06} & \textbf{49.65} & & & \\
& LLM-MvP (ours) & 36.83 & \textbf{64.13} & \textbf{70.80} & \textbf{71.97} & \textbf{60.93} & & & & 33.10 & 49.85 & 55.35 & 56.75 & 48.76 & & & \\
\midrule
\multirow{4}{*}{\textsc{Coursera}} & Single-order & 18.35 & 40.19 & 46.15 & 47.31 & 38.00 & \multicolumn{1}{c}{\multirow{4}{*}{52.73}} & \multicolumn{1}{c}{\multirow{4}{*}{51.42}} & \multicolumn{1}{c}{\multirow{4}{*}{\underline{\textbf{53.35}}}} & 5.33 & 21.58 & 27.04 & 27.19 & 20.28 & \multicolumn{1}{c}{\multirow{4}{*}{32.06}} & \multicolumn{1}{c}{\multirow{4}{*}{32.04}} & \multicolumn{1}{c}{\multirow{4}{*}{\textbf{33.12}}} \\
& Reasoning & \textbf{24.23} & 31.28 & 34.38 & 37.72 & 31.90 & & & & \textbf{12.60} & 18.46 & 20.69 & 23.14 & 18.72 & & & \\
& Self-consistency & 23.84 & 44.37 & 49.10 & 51.49 & 42.20 & & & & 8.39 & \textbf{28.40} & \textbf{34.30} & \underline{\textbf{34.71}} & \textbf{26.45} & & & \\
& LLM-MvP (ours) & 21.99 & \textbf{44.49} & \textbf{50.97} & \textbf{52.33} & \textbf{42.44} & & & & 8.01 & 23.69 & 30.05 & 31.05 & 23.20 & & & \\
\midrule
\multirow{4}{*}{\textsc{Hotels}} & Single-order & 31.09 & 62.98 & 66.57 & 68.34 & 57.25 & \multicolumn{1}{c}{\multirow{4}{*}{68.41}} & \multicolumn{1}{c}{\multirow{4}{*}{69.58}} & \multicolumn{1}{c}{\multirow{4}{*}{\textbf{72.11}}} & 17.32 & 42.85 & 46.03 & 48.51 & 38.68 & \multicolumn{1}{c}{\multirow{4}{*}{52.61}} & \multicolumn{1}{c}{\multirow{4}{*}{54.38}} & \multicolumn{1}{c}{\multirow{4}{*}{\underline{\textbf{55.91}}}} \\
& Reasoning & \textbf{41.09} & 50.09 & 53.30 & 53.10 & 49.40 & & & & \textbf{32.50} & 39.29 & 42.36 & 42.72 & 39.22 & & & \\
& Self-consistency & 38.50 & 66.79 & 68.97 & 70.75 & 61.25 & & & & 24.90 & \textbf{49.92} & 51.20 & 52.99 & 44.75 & & & \\
& LLM-MvP (ours) & 36.54 & \textbf{68.72} & \textbf{71.12} & \underline{\textbf{72.46}} & \textbf{62.21} & & & & 27.56 & 48.45 & \textbf{52.73} & \textbf{54.70} & \textbf{45.86} & & & \\
\bottomrule
\end{tabular}

}
\caption{\textbf{Precision across TASD and ASQP.} Layout as in Table~\ref{tab:results}.}
\label{tab:results-tasd-precision}

\vspace{1.5em}
\resizebox{\textwidth}{!}{%
\begin{tabular}{@{}ll @{\hskip 6pt} rrrr @{\hskip 3pt} r @{\hskip 3pt} rrr @{\hspace{10pt}} rrrr @{\hskip 3pt} r @{\hskip 3pt} rrr@{}}
\toprule
\multirow{3}{*}{\textbf{Dataset}} & \multirow{3}{*}{\textbf{Approach}} & \multicolumn{8}{c}{\textbf{TASD} (Recall)} & \multicolumn{8}{c}{\textbf{ASQP} (Recall)} \\
\cmidrule(lr){3-10}\cmidrule(lr){11-18}
& & \multicolumn{4}{c}{\textit{\# Shots}} & \multirow{2}{*}{\textit{Avg.}} & \multicolumn{3}{c}{\textit{FT Baselines}} & \multicolumn{4}{c}{\textit{\# Shots}} & \multirow{2}{*}{\textit{Avg.}} & \multicolumn{3}{c}{\textit{FT Baselines}} \\
\cmidrule(lr){3-6}\cmidrule(lr){8-10}\cmidrule(lr){11-14}\cmidrule(lr){16-18}
& & \textbf{0} & \textbf{10} & \textbf{50} & \textbf{100} & & Para.$^{\star}$ & MvP$^{\star}$ & LLM$^{\star}$ & \textbf{0} & \textbf{10} & \textbf{50} & \textbf{100} & & Para.$^{\star}$ & MvP$^{\star}$ & LLM$^{\star}$ \\
\midrule
\multirow{4}{*}{\textsc{Rest15}} & Single-order & 41.09 & 56.83 & 60.76 & 63.12 & 55.45 & \multicolumn{1}{c}{\multirow{4}{*}{-}} & \multicolumn{1}{c}{\multirow{4}{*}{-}} & \multicolumn{1}{c}{\multirow{4}{*}{\underline{70.79}}} & 28.05 & 41.53 & 48.45 & 50.99 & 42.26 & \multicolumn{1}{c}{\multirow{4}{*}{47.72}} & \multicolumn{1}{c}{\multirow{4}{*}{-}} & \multicolumn{1}{c}{\multirow{4}{*}{\underline{\textbf{56.83}}}} \\
& Reasoning & \textbf{50.30} & 53.82 & 57.78 & 60.02 & 55.48 & & & & \textbf{35.52} & 39.37 & 39.75 & 42.09 & 39.18 & & & \\
& Self-consistency & 39.67 & 55.83 & 59.74 & 62.53 & 54.44 & & & & 25.13 & 40.18 & 46.92 & 49.26 & 40.37 & & & \\
& LLM-MvP (ours) & 39.05 & \textbf{60.14} & \textbf{64.62} & \textbf{67.24} & \textbf{57.76} & & & & 34.04 & \textbf{46.39} & \textbf{50.54} & \textbf{52.88} & \textbf{45.96} & & & \\
\midrule
\multirow{4}{*}{\textsc{Rest16}} & Single-order & 51.08 & 62.19 & 67.89 & 69.57 & 62.68 & \multicolumn{1}{c}{\multirow{4}{*}{-}} & \multicolumn{1}{c}{\multirow{4}{*}{-}} & \multicolumn{1}{c}{\multirow{4}{*}{\underline{78.60}}} & 33.57 & 51.54 & 55.77 & 57.55 & 49.61 & \multicolumn{1}{c}{\multirow{4}{*}{59.30}} & \multicolumn{1}{c}{\multirow{4}{*}{-}} & \multicolumn{1}{c}{\multirow{4}{*}{\underline{\textbf{68.19}}}} \\
& Reasoning & \textbf{55.69} & 59.09 & 62.44 & 62.54 & 59.94 & & & & 40.25 & 44.98 & 47.61 & 49.36 & 45.55 & & & \\
& Self-consistency & 48.64 & 61.12 & 67.08 & 69.64 & 61.62 & & & & 32.77 & 49.46 & 54.47 & 56.60 & 48.32 & & & \\
& LLM-MvP (ours) & 49.97 & \textbf{64.84} & \textbf{70.94} & \textbf{73.08} & \textbf{64.71} & & & & \textbf{41.23} & \textbf{53.72} & \textbf{58.52} & \textbf{61.45} & \textbf{53.73} & & & \\
\midrule
\multirow{4}{*}{\textsc{FlightABSA}} & Single-order & 44.61 & 58.71 & 65.48 & 66.99 & 58.95 & \multicolumn{1}{c}{\multirow{4}{*}{69.26}} & \multicolumn{1}{c}{\multirow{4}{*}{69.53}} & \multicolumn{1}{c}{\multirow{4}{*}{\underline{\textbf{74.44}}}} & 26.34 & 48.88 & 54.20 & 56.14 & 46.39 & \multicolumn{1}{c}{\multirow{4}{*}{58.17}} & \multicolumn{1}{c}{\multirow{4}{*}{59.83}} & \multicolumn{1}{c}{\multirow{4}{*}{\underline{\textbf{64.98}}}} \\
& Reasoning & \textbf{59.09} & \textbf{64.46} & \textbf{68.51} & \textbf{69.98} & \textbf{65.51} & & & & \textbf{40.64} & 46.31 & 50.54 & 48.81 & 46.58 & & & \\
& Self-consistency & 42.23 & 56.86 & 64.91 & 66.12 & 57.53 & & & & 25.66 & 47.02 & 52.98 & 54.54 & 45.05 & & & \\
& LLM-MvP (ours) & 41.70 & 60.49 & 66.09 & 68.36 & 59.16 & & & & 35.08 & \textbf{51.25} & \textbf{56.27} & \textbf{58.03} & \textbf{50.16} & & & \\
\midrule
\multirow{4}{*}{\textsc{Coursera}} & Single-order & 24.92 & 37.01 & 42.17 & 44.96 & 37.26 & \multicolumn{1}{c}{\multirow{4}{*}{51.02}} & \multicolumn{1}{c}{\multirow{4}{*}{50.53}} & \multicolumn{1}{c}{\multirow{4}{*}{\underline{\textbf{52.54}}}} & 7.57 & \textbf{22.47} & \textbf{27.97} & 29.20 & 21.80 & \multicolumn{1}{c}{\multirow{4}{*}{32.63}} & \multicolumn{1}{c}{\multirow{4}{*}{32.97}} & \multicolumn{1}{c}{\multirow{4}{*}{\underline{\textbf{34.22}}}} \\
& Reasoning & \textbf{28.69} & 36.07 & 39.88 & 42.50 & 36.78 & & & & \textbf{14.70} & 21.16 & 23.43 & 24.86 & 21.04 & & & \\
& Self-consistency & 21.07 & 35.49 & 40.57 & 43.61 & 35.18 & & & & 6.73 & 21.24 & 25.90 & 27.57 & 20.36 & & & \\
& LLM-MvP (ours) & 21.64 & \textbf{38.07} & \textbf{44.02} & \textbf{46.31} & \textbf{37.51} & & & & 11.55 & 22.23 & 27.57 & \textbf{29.28} & \textbf{22.66} & & & \\
\midrule
\multirow{4}{*}{\textsc{Hotels}} & Single-order & 37.33 & 59.40 & 62.45 & 63.94 & 55.78 & \multicolumn{1}{c}{\multirow{4}{*}{67.01}} & \multicolumn{1}{c}{\multirow{4}{*}{69.16}} & \multicolumn{1}{c}{\multirow{4}{*}{\underline{\textbf{70.84}}}} & 20.83 & 45.99 & 49.63 & 50.90 & 41.84 & \multicolumn{1}{c}{\multirow{4}{*}{55.19}} & \multicolumn{1}{c}{\multirow{4}{*}{55.69}} & \multicolumn{1}{c}{\multirow{4}{*}{\underline{\textbf{57.20}}}} \\
& Reasoning & \textbf{47.57} & 58.31 & 61.88 & 60.95 & \textbf{57.18} & & & & \textbf{32.73} & 40.58 & 42.25 & 41.83 & 39.35 & & & \\
& Self-consistency & 35.52 & 58.28 & 61.75 & 63.15 & 54.67 & & & & 18.47 & 44.99 & 47.85 & 49.10 & 40.10 & & & \\
& LLM-MvP (ours) & 37.62 & \textbf{60.51} & \textbf{64.10} & \textbf{64.74} & 56.74 & & & & 27.24 & \textbf{47.57} & \textbf{52.87} & \textbf{53.65} & \textbf{45.33} & & & \\
\bottomrule
\end{tabular}

}
\caption{\textbf{Recall across TASD and ASQP.} Layout as in Table~\ref{tab:results}.}
\label{tab:results-tasd-recall}
\end{table*}

\begin{table*}[t]
\centering
\small
\setlength{\tabcolsep}{1.0pt}
\renewcommand{\arraystretch}{0.8} 
\resizebox{\textwidth}{!}{%
\begin{tabular}{@{}ll @{\hskip 6pt} rrrr @{\hskip 3pt} r @{\hskip 3pt} rrr @{\hspace{10pt}} rrrr @{\hskip 3pt} r @{\hskip 3pt} rrr@{}}
\toprule
\multirow{3}{*}{\textbf{Dataset}} & \multirow{3}{*}{\textbf{Approach}} & \multicolumn{8}{c}{\textbf{TASD} ($F_1$ Macro)} & \multicolumn{8}{c}{\textbf{ASQP} ($F_1$ Macro)} \\
\cmidrule(lr){3-10}\cmidrule(lr){11-18}
& & \multicolumn{4}{c}{\textit{\# Shots}} & \multirow{2}{*}{\textit{Avg.}} & \multicolumn{3}{c}{\textit{FT Baselines}} & \multicolumn{4}{c}{\textit{\# Shots}} & \multirow{2}{*}{\textit{Avg.}} & \multicolumn{3}{c}{\textit{FT Baselines}} \\
\cmidrule(lr){3-6}\cmidrule(lr){8-10}\cmidrule(lr){11-14}\cmidrule(lr){16-18}
& & \textbf{0} & \textbf{10} & \textbf{50} & \textbf{100} & & Para.$^{\star}$ & MvP$^{\star}$ & LLM$^{\star}$ & \textbf{0} & \textbf{10} & \textbf{50} & \textbf{100} & & Para.$^{\star}$ & MvP$^{\star}$ & LLM$^{\star}$ \\
\midrule
\multirow{4}{*}{\textsc{Rest15}} & Single-order & 26.02 & 38.50 & 47.15 & 48.76 & 40.11 & \multicolumn{1}{c}{\multirow{4}{*}{-}} & \multicolumn{1}{c}{\multirow{4}{*}{-}} & \multicolumn{1}{c}{\multirow{4}{*}{\underline{60.26}}} & 20.25 & 29.85 & 37.69 & 38.71 & 31.63 & \multicolumn{1}{c}{\multirow{4}{*}{-}} & \multicolumn{1}{c}{\multirow{4}{*}{-}} & \multicolumn{1}{c}{\multirow{4}{*}{\underline{45.18}}} \\
& Reasoning & \textbf{36.54} & 36.86 & 39.76 & 41.61 & 38.69 & & & & \textbf{26.46} & 28.39 & 28.55 & 32.15 & 28.88 & & & \\
& Self-consistency & 27.31 & 38.47 & 47.00 & 49.28 & 40.52 & & & & 19.71 & 31.48 & 37.23 & 38.51 & 31.73 & & & \\
& LLM-MvP (ours) & 27.04 & \textbf{42.23} & \textbf{53.35} & \textbf{54.83} & \textbf{44.36} & & & & 21.82 & \textbf{33.37} & \textbf{39.84} & \textbf{41.78} & \textbf{34.20} & & & \\
\midrule
\multirow{4}{*}{\textsc{Rest16}} & Single-order & 34.59 & 50.93 & 57.25 & 58.78 & 50.39 & \multicolumn{1}{c}{\multirow{4}{*}{-}} & \multicolumn{1}{c}{\multirow{4}{*}{-}} & \multicolumn{1}{c}{\multirow{4}{*}{\underline{72.40}}} & 19.88 & 37.87 & 44.33 & 48.47 & 37.64 & \multicolumn{1}{c}{\multirow{4}{*}{-}} & \multicolumn{1}{c}{\multirow{4}{*}{-}} & \multicolumn{1}{c}{\multirow{4}{*}{\underline{58.15}}} \\
& Reasoning & \textbf{40.58} & 41.40 & 43.70 & 44.02 & 42.42 & & & & \textbf{31.86} & 31.38 & 33.71 & 36.08 & 33.26 & & & \\
& Self-consistency & 32.11 & 49.43 & 56.75 & 60.12 & 49.60 & & & & 22.57 & 39.03 & 45.34 & 49.43 & 39.10 & & & \\
& LLM-MvP (ours) & 29.00 & \textbf{51.56} & \textbf{59.98} & \textbf{61.15} & \textbf{50.42} & & & & 26.82 & \textbf{39.31} & \textbf{48.20} & \textbf{50.91} & \textbf{41.31} & & & \\
\midrule
\multirow{4}{*}{\textsc{FlightABSA}} & Single-order & 34.37 & 55.33 & 59.75 & 59.77 & 52.30 & \multicolumn{1}{c}{\multirow{4}{*}{60.97}} & \multicolumn{1}{c}{\multirow{4}{*}{63.65}} & \multicolumn{1}{c}{\multirow{4}{*}{\underline{\textbf{65.51}}}} & 23.15 & 44.65 & 49.98 & 50.97 & 42.19 & \multicolumn{1}{c}{\multirow{4}{*}{51.79}} & \multicolumn{1}{c}{\multirow{4}{*}{52.07}} & \multicolumn{1}{c}{\multirow{4}{*}{\underline{\textbf{54.39}}}} \\
& Reasoning & \textbf{49.42} & 53.10 & 54.15 & 55.62 & 53.07 & & & & \textbf{38.93} & 40.16 & 41.66 & 41.58 & 40.58 & & & \\
& Self-consistency & 38.95 & 55.22 & 58.92 & 59.74 & 53.21 & & & & 27.92 & 45.89 & 52.01 & 53.30 & 44.78 & & & \\
& LLM-MvP (ours) & 35.97 & \textbf{58.00} & \textbf{61.98} & \textbf{62.62} & \textbf{54.64} & & & & 33.55 & \textbf{47.27} & \textbf{53.10} & \textbf{54.28} & \textbf{47.05} & & & \\
\midrule
\multirow{4}{*}{\textsc{Coursera}} & Single-order & 11.51 & 17.16 & 20.60 & 23.25 & 18.13 & \multicolumn{1}{c}{\multirow{4}{*}{\underline{\textbf{45.86}}}} & \multicolumn{1}{c}{\multirow{4}{*}{44.18}} & \multicolumn{1}{c}{\multirow{4}{*}{35.28}} & 2.68 & 7.04 & 10.15 & 11.14 & 7.75 & \multicolumn{1}{c}{\multirow{4}{*}{\underline{\textbf{27.92}}}} & \multicolumn{1}{c}{\multirow{4}{*}{26.13}} & \multicolumn{1}{c}{\multirow{4}{*}{16.12}} \\
& Reasoning & \textbf{14.65} & \textbf{17.92} & \textbf{21.36} & 23.77 & \textbf{19.42} & & & & \textbf{5.97} & \textbf{7.41} & 8.49 & 8.94 & 7.70 & & & \\
& Self-consistency & 10.44 & 16.32 & 20.50 & 22.98 & 17.56 & & & & 2.66 & 7.36 & 10.25 & 11.24 & 7.88 & & & \\
& LLM-MvP (ours) & 12.29 & 17.14 & 20.72 & \textbf{25.32} & 18.86 & & & & 3.42 & 7.21 & \textbf{11.29} & \textbf{11.92} & \textbf{8.46} & & & \\
\midrule
\multirow{4}{*}{\textsc{Hotels}} & Single-order & 14.01 & 30.64 & 35.10 & 36.76 & 29.13 & \multicolumn{1}{c}{\multirow{4}{*}{45.89}} & \multicolumn{1}{c}{\multirow{4}{*}{\underline{\textbf{48.66}}}} & \multicolumn{1}{c}{\multirow{4}{*}{41.08}} & 11.58 & 25.43 & 28.45 & 28.83 & 23.57 & \multicolumn{1}{c}{\multirow{4}{*}{37.97}} & \multicolumn{1}{c}{\multirow{4}{*}{\underline{\textbf{39.46}}}} & \multicolumn{1}{c}{\multirow{4}{*}{32.11}} \\
& Reasoning & \textbf{23.13} & 28.09 & 31.86 & 32.07 & 28.79 & & & & \textbf{19.12} & 24.16 & 24.93 & 24.13 & 23.08 & & & \\
& Self-consistency & 15.12 & 30.89 & 34.66 & 37.55 & 29.55 & & & & 12.12 & 27.08 & 28.78 & 28.96 & 24.24 & & & \\
& LLM-MvP (ours) & 15.74 & \textbf{31.66} & \textbf{36.59} & \textbf{38.05} & \textbf{30.51} & & & & 16.34 & \textbf{28.17} & \textbf{30.74} & \textbf{32.13} & \textbf{26.84} & & & \\
\bottomrule
\end{tabular}

}
\caption{\textbf{Macro-averaged $F_1$ across TASD and ASQP.} Layout as in Table~\ref{tab:results}.}
\label{tab:results-tasd-f1-macro}
\end{table*}

\FloatBarrier

\section{Performance Comparison of Gemma, Mistral, and Qwen}\label{appendix:llm-comparison}

We did not evaluate the latest iterations of the Qwen series (versions 3.6 and 3.8)\footnote{\hficon\url{https://hf.co/collections/Qwen/qwen36}}. Unlike their transformer-based predecessors, these versions adopt a hybrid architecture that interleaves standard attention with Gated DeltaNet linear-attention layers \citep{yang2025gated}, a recurrent design building on Mamba2 \citep{mamba2}, for which prefix caching is not yet implemented in vLLM\footnote{\ghicon\url{https://github.com/vllm-project/vllm/issues/36493}}.

\begin{table*}[t]
\centering
\resizebox{0.9\textwidth}{!}{%
\begin{tabular}{@{}lll @{\hskip 6pt} rrrr @{\hskip 3pt} r @{\hspace{10pt}} rrrr @{\hskip 3pt} r @{}}
\toprule
\multirow{3}{*}{\textbf{Dataset}} & \multirow{3}{*}{\textbf{Model}} & \multirow{3}{*}{\textbf{Approach}} & \multicolumn{5}{c}{\textbf{TASD} ($F_1$)} & \multicolumn{5}{c}{\textbf{ASQP} ($F_1$)} \\
\cmidrule(lr){4-8}\cmidrule(lr){9-13}
& & & \multicolumn{4}{c}{\textit{\# Shots}} & \multirow{2}{*}{\textit{Avg.}} & \multicolumn{4}{c}{\textit{\# Shots}} & \multirow{3}{*}{\textit{Avg.}} \\
\cmidrule(lr){4-7}\cmidrule(lr){9-12}
& & & \textbf{0} & \textbf{10} & \textbf{50} & \textbf{100} & & \textbf{0} & \textbf{10} & \textbf{50} & \textbf{100} & \\
\midrule
\multirow{8}{*}{\textsc{Rest15}} 
& \multirow{2}{*}{Gemma 4 (31B)}           & Single   & 40.48 & 58.82 & 63.05 & 65.27 & 56.91 & 25.59 & 40.43 & 47.99 & 50.70 & 41.18 \\
&                                          & LLM-MvP  & \textbf{40.88} & \textbf{63.29} & \textbf{68.18} & \underline{\textbf{70.50}} & \textbf{60.71} & 32.88 & \textbf{45.56} & \textbf{50.99} & \underline{\textbf{53.40}} & \textbf{45.71} \\
\cmidrule(lr){2-8} \cmidrule(lr){9-13}
& \multirow{2}{*}{Gemma 4 (26B MoE)}       & Single   & 29.18 & 51.08 & 59.23 & 60.12 & 49.90 & 17.55 & 33.38 & 40.03 & 42.34 & 33.32 \\
&                                          & LLM-MvP  & 34.59 & 54.95 & 62.62 & 64.40 & 54.14 & 25.50 & 40.40 & 46.94 & 48.98 & 40.45 \\
\cmidrule(lr){2-8} \cmidrule(lr){9-13}
& \multirow{2}{*}{Mistral Small 3.2 (24B)} & Single   & 37.28 & 58.59 & 63.68 & 65.41 & 56.24 & 24.59 & 41.01 & 45.46 & 47.63 & 39.67 \\
&                                          & LLM-MvP  & 37.22 & 61.05 & 66.64 & 68.52 & 58.36 & \textbf{33.23} & 43.47 & 47.99 & 51.09 & 43.95 \\
\cmidrule(lr){2-8} \cmidrule(lr){9-13}
& \multirow{2}{*}{Qwen 3 (30B)}            & Single   & 26.15 & 44.32 & 47.95 & 48.08 & 41.62 & 14.97 & 26.26 & 28.67 & 29.64 & 24.89 \\
&                                          & LLM-MvP  & 31.41 & 50.83 & 58.80 & 60.45 & 50.37 & 28.19 & 36.66 & 38.93 & 40.63 & 36.10 \\
\midrule
\multirow{8}{*}{\textsc{Rest16}}
& \multirow{2}{*}{Gemma 4 (31B)}           & Single   & 50.10 & 64.30 & 69.30 & 70.74 & 63.61 & 30.13 & 48.98 & 54.70 & 56.36 & 47.54 \\
&                                          & LLM-MvP  & 51.03 & \textbf{67.79} & \textbf{73.19} & \underline{\textbf{74.95}} & \textbf{66.74} & 39.15 & \textbf{52.46} & \textbf{58.08} & \underline{\textbf{60.61}} & \textbf{52.57} \\
\cmidrule(lr){2-8} \cmidrule(lr){9-13}
& \multirow{2}{*}{Gemma 4 (26B MoE)}       & Single   & 44.72 & 57.38 & 63.25 & 65.12 & 57.62 & 26.74 & 40.62 & 47.17 & 50.38 & 41.23 \\
&                                          & LLM-MvP  & 48.48 & 58.68 & 65.54 & 68.26 & 60.24 & 30.54 & 48.05 & 52.20 & 53.92 & 46.18 \\
\cmidrule(lr){2-8} \cmidrule(lr){9-13}
& \multirow{2}{*}{Mistral Small 3.2 (24B)} & Single   & \textbf{54.58} & 63.99 & 69.59 & 71.40 & 64.89 & 29.67 & 49.84 & 53.59 & 55.61 & 47.18 \\
&                                          & LLM-MvP  & 54.42 & 65.31 & 71.44 & 73.05 & 66.05 & \textbf{39.97} & 51.65 & 56.49 & 57.63 & 51.43 \\
\cmidrule(lr){2-8} \cmidrule(lr){9-13}
& \multirow{2}{*}{Qwen 3 (30B)}            & Single   & 38.77 & 50.72 & 54.03 & 53.66 & 49.29 & 17.77 & 33.37 & 36.83 & 38.66 & 31.66 \\
&                                          & LLM-MvP  & 46.17 & 59.01 & 65.14 & 65.65 & 58.99 & 32.08 & 44.46 & 48.85 & 49.35 & 43.68 \\
\midrule
\multirow{8}{*}{\textsc{FlightABSA}}
& \multirow{2}{*}{Gemma 4 (31B)}           & Single   & 37.93 & 59.27 & 66.38 & 67.59 & 57.79 & 24.32 & 47.29 & 53.33 & 54.90 & 44.96 \\
&                                          & LLM-MvP  & 39.11 & \textbf{62.22} & \textbf{68.36} & \underline{\textbf{70.11}} & 59.95 & 34.06 & \textbf{50.47} & 55.80 & 57.37 & 49.42 \\
\cmidrule(lr){2-8} \cmidrule(lr){9-13}
& \multirow{2}{*}{Gemma 4 (26B MoE)}       & Single   & 37.53 & 53.26 & 62.31 & 63.35 & 54.11 & 25.77 & 41.55 & 48.54 & 49.50 & 41.34 \\
&                                          & LLM-MvP  & 34.87 & 55.15 & 65.14 & 67.37 & 55.63 & 18.29 & 48.27 & 54.53 & 56.20 & 44.32 \\
\cmidrule(lr){2-8} \cmidrule(lr){9-13}
& \multirow{2}{*}{Mistral Small 3.2 (24B)} & Single   & 52.10 & 60.78 & 65.91 & 68.54 & 61.83 & 27.05 & 47.20 & 52.99 & 55.12 & 45.59 \\
&                                          & LLM-MvP  & \textbf{53.70} & 61.43 & 67.60 & 69.33 & \textbf{63.01} & \textbf{41.55} & 50.33 & \textbf{56.46} & \underline{\textbf{58.13}} & \textbf{51.62} \\
\cmidrule(lr){2-8} \cmidrule(lr){9-13}
& \multirow{2}{*}{Qwen 3 (30B)}            & Single   & 40.17 & 50.84 & 54.21 & 54.68 & 49.97 & 23.90 & 34.19 & 39.72 & 39.79 & 34.40 \\
&                                          & LLM-MvP  & 44.23 & 57.76 & 62.80 & 64.93 & 57.43 & 35.97 & 44.39 & 47.48 & 50.04 & 44.47 \\
\midrule
\multirow{8}{*}{\textsc{Coursera}}
& \multirow{2}{*}{Gemma 4 (31B)}           & Single   & 21.13 & 38.52 & 44.06 & 46.09 & 37.45 & 6.26 & 22.01 & 27.49 & 28.14 & 20.97 \\
&                                          & LLM-MvP  & 21.81 & \textbf{41.01} & \textbf{47.22} & \underline{\textbf{49.12}} & \textbf{39.79} & 9.46 & \textbf{22.93} & \textbf{28.75} & \underline{\textbf{30.13}} & \textbf{22.82} \\
\cmidrule(lr){2-8} \cmidrule(lr){9-13}
& \multirow{2}{*}{Gemma 4 (26B MoE)}       & Single   & 15.03 & 32.20 & 41.33 & 42.74 & 32.83 & 8.31 & 16.58 & 23.40 & 25.38 & 18.42 \\
&                                          & LLM-MvP  & 18.13 & 34.97 & 44.05 & 44.73 & 35.47 & 6.52 & 18.77 & 26.51 & 28.26 & 20.01 \\
\cmidrule(lr){2-8} \cmidrule(lr){9-13}
& \multirow{2}{*}{Mistral Small 3.2 (24B)} & Single   & 21.63 & 33.65 & 45.98 & 46.20 & 36.87 & 9.02 & 19.06 & 25.65 & 27.50 & 20.31 \\
&                                          & LLM-MvP  & 24.25 & 36.26 & 46.92 & 48.38 & 38.95 & 9.61 & 21.03 & 27.31 & 28.82 & 21.69 \\
\cmidrule(lr){2-8} \cmidrule(lr){9-13}
& \multirow{2}{*}{Qwen 3 (30B)}            & Single   & 19.98 & 25.36 & 29.03 & 31.09 & 26.37 & 11.57 & 13.88 & 18.62 & 19.37 & 15.86 \\
&                                          & LLM-MvP  & \textbf{25.22} & 34.43 & 42.16 & 43.92 & 36.43 & \textbf{16.24} & 19.71 & 24.92 & 26.68 & 21.89 \\
\midrule
\multirow{8}{*}{\textsc{Hotels}}
& \multirow{2}{*}{Gemma 4 (31B)}           & Single   & 33.93 & 61.09 & 64.44 & 66.06 & 56.38 & 18.91 & 44.27 & 47.74 & 49.66 & 40.15 \\
&                                          & LLM-MvP  & 37.07 & \textbf{64.33} & \textbf{67.42} & \underline{\textbf{68.38}} & \textbf{59.30} & 27.40 & \textbf{47.90} & \textbf{52.78} & \underline{\textbf{54.16}} & \textbf{45.56} \\
\cmidrule(lr){2-8} \cmidrule(lr){9-13}
& \multirow{2}{*}{Gemma 4 (26B MoE)}       & Single   & 33.54 & 49.35 & 56.46 & 59.46 & 49.70 & 22.00 & 31.72 & 38.90 & 42.73 & 33.84 \\
&                                          & LLM-MvP  & 29.99 & 52.40 & 60.28 & 64.01 & 51.67 & 17.48 & 40.96 & 46.71 & 50.26 & 38.85 \\
\cmidrule(lr){2-8} \cmidrule(lr){9-13}
& \multirow{2}{*}{Mistral Small 3.2 (24B)} & Single   & 38.36 & 57.36 & 62.40 & 64.93 & 55.76 & 20.18 & 40.03 & 45.71 & 47.40 & 38.33 \\
&                                          & LLM-MvP  & \textbf{42.43} & 59.85 & 64.54 & 67.17 & 58.50 & 28.34 & 45.22 & 49.05 & 51.38 & 43.50 \\
\cmidrule(lr){2-8} \cmidrule(lr){9-13}
& \multirow{2}{*}{Qwen 3 (30B)}            & Single   & 30.26 & 41.84 & 44.07 & 44.64 & 40.20 & 21.99 & 28.62 & 30.50 & 31.27 & 28.09 \\
&                                          & LLM-MvP  & 36.10 & 56.78 & 58.58 & 61.86 & 53.33 & \textbf{32.54} & 42.36 & 45.51 & 48.17 & 42.14 \\
\bottomrule
\end{tabular}

}
\caption{\textbf{Evaluation of LLM-MvP on various LLMs.} The best score for a given combination of dataset, task, and shot count is highlighted in \textbf{bold}.}
\label{tab:llm_mvp_comparison}
\end{table*}

\section{Variation of the Number of Views}\label{appendix:m-variation}

Table~\ref{tab:m_variation_llm_mvp} presents the performance impact of varying the number of lowest-entropy views $m$ used for majority voting. For both TASD and ASQP, increasing the number of permutations $m$ generally improves performance, with modest gains for TASD (\val{mvar.tasd.m1.avg} to \val{mvar.tasd.best.avg} on average from $m=1$ to $m=\val{mvar.tasd.best.m}$) and larger gains for ASQP (\val{mvar.asqp.m1.avg} to \val{mvar.asqp.best.avg} from $m=1$ to $m=\val{mvar.asqp.best.m}$).

We focus on odd values of $m$, which avoid ties in the vote. While \citet{gou2023mvp} only explored a coarse selection of views ($m \in \{1, 3, 7, 15, 24\}$), we evaluated all odd values of $m$ as well as the full set of permutations, i.e., $m \in \{1, 3, 5, 6\}$ for TASD and $m \in \{1, 3, \ldots, 23, 24\}$ for ASQP. Our results for TASD show a peak at $m=\val{mvar.tasd.best.m}$. For the more complex ASQP task, performance keeps increasing up to $m=\val{mvar.asqp.best.m}$. This aligns with and slightly extends the findings of \citet{gou2023mvp}, who observed a peak at $m=15$ for ASQP. 

In both tasks, including all permutations ($m=6$ for TASD and $m=24$ for ASQP) does not yield the highest scores. This decline suggests that the lowest-confidence permutations add noise that outweighs the benefit of further views. Since performance changes only slightly between neighboring values of $m$, we recommend the best average values, $m=\val{mvar.tasd.best.m}$ for TASD and $m=\val{mvar.asqp.best.m}$ for ASQP, as defaults when no annotated data is available for selecting $m$.

\begin{table*}[t]
\centering
\resizebox{1.0\textwidth}{!}{%
\begin{tabular}{ll cccc ccccccccccccc}
\toprule
& & \multicolumn{4}{c}{\textbf{TASD}} & \multicolumn{13}{c}{\textbf{ASQP}} \\
\cmidrule(lr){3-6} \cmidrule(lr){7-19}
\textbf{Dataset} & \textbf{\# Shots} & \textbf{1} & \textbf{3} & \textbf{5} & \textbf{full (6)} & \textbf{1} & \textbf{3} & \textbf{5} & \textbf{7} & \textbf{9} & \textbf{11} & \textbf{13} & \textbf{15} & \textbf{17} & \textbf{19} & \textbf{21} & \textbf{23} & \textbf{full (24)} \\
\midrule
Rest15 & 0 & 38.72 & 39.07 & \textbf{40.88} & 40.54 & 30.71 & 32.24 & 32.88 & 33.47 & 33.00 & 33.49 & 33.61 & 34.09 & \textbf{34.28} & 33.83 & 33.15 & 33.29 & 33.33 \\
& 10 & 62.93 & 62.95 & \textbf{63.29} & 63.18 & 44.17 & 44.88 & 45.56 & 45.67 & 45.96 & 46.15 & \textbf{46.42} & 46.32 & 46.28 & 46.18 & 45.94 & 46.21 & 46.09 \\
& 50 & 67.72 & 68.09 & \textbf{68.18} & 67.75 & 49.92 & 50.51 & 50.99 & 51.02 & 51.07 & 51.20 & 51.91 & 51.91 & 52.07 & 51.87 & 51.95 & 52.21 & \textbf{52.24} \\
& 100 & 70.04 & 70.23 & \textbf{70.50} & 69.84 & 52.84 & 53.56 & 53.40 & 53.79 & 54.03 & 54.09 & 54.86 & 54.81 & 54.94 & 54.96 & \textbf{55.21} & 54.99 & 55.13 \\
\midrule
Rest16 & 0 & 50.12 & 51.56 & 51.03 & \textbf{51.68} & 35.95 & 38.13 & 39.15 & 39.34 & 39.69 & 39.95 & 40.39 & 40.96 & 41.15 & \textbf{41.24} & 40.57 & 40.28 & 40.24 \\
& 10 & 66.40 & 67.17 & \textbf{67.79} & 67.64 & 51.01 & 51.74 & 52.46 & 52.53 & 53.05 & 53.38 & 53.71 & 53.91 & 54.00 & 54.31 & 54.36 & \textbf{54.62} & 54.53 \\
& 50 & 72.13 & 72.97 & \textbf{73.19} & 72.70 & 57.10 & 58.46 & 58.08 & 58.45 & 58.47 & 59.03 & 59.60 & 59.88 & 59.65 & 59.85 & 59.99 & \textbf{60.27} & 60.08 \\
& 100 & 74.54 & 74.83 & \textbf{74.95} & 74.47 & 59.70 & 60.31 & 60.61 & 60.77 & 61.45 & 61.87 & 62.39 & 62.25 & 62.51 & 62.64 & 62.60 & \textbf{62.98} & 62.89 \\
\midrule
FlightABSA & 0 & 36.19 & 38.75 & \textbf{39.11} & 38.61 & 27.93 & 32.83 & \textbf{34.06} & 33.65 & 33.56 & 33.27 & 32.28 & 32.28 & 31.54 & 31.03 & 30.38 & 29.94 & 30.05 \\
& 10 & 62.24 & \textbf{62.44} & 62.22 & 62.35 & 49.07 & 49.96 & 50.47 & 50.88 & 50.93 & 51.17 & 51.33 & 51.44 & 51.46 & 51.46 & 51.37 & \textbf{51.75} & 51.50 \\
& 50 & 68.41 & \textbf{68.73} & 68.36 & 68.20 & 54.86 & 55.11 & 55.80 & 56.09 & 56.08 & 56.00 & 56.11 & 56.07 & 56.30 & 56.32 & 56.59 & \textbf{56.76} & 56.67 \\
& 100 & 69.92 & 70.10 & \textbf{70.11} & 69.64 & 56.31 & 57.07 & 57.37 & 57.65 & 57.93 & 58.38 & 58.41 & 58.39 & 58.45 & 58.47 & \textbf{58.63} & 58.19 & 57.97 \\
\midrule
OATS Coursera & 0 & 18.37 & 21.60 & 21.81 & \textbf{22.45} & 7.64 & 9.01 & 9.46 & 9.77 & 9.66 & 9.80 & 9.73 & 9.87 & 10.34 & \textbf{10.66} & 10.63 & 10.08 & 10.04 \\
& 10 & 40.36 & 40.33 & \textbf{41.01} & 40.71 & 21.69 & 22.58 & 22.93 & 23.51 & 23.73 & 24.05 & 24.28 & 24.66 & 24.87 & 25.39 & 25.47 & \textbf{25.76} & 25.76 \\
& 50 & 46.84 & 47.04 & 47.22 & \textbf{47.30} & 27.61 & 28.36 & 28.75 & 29.31 & 29.47 & 30.09 & 30.87 & 31.20 & \textbf{31.40} & 31.26 & 31.08 & 31.02 & 31.15 \\
& 100 & 48.53 & \textbf{49.37} & 49.12 & 48.94 & 29.02 & 29.97 & 30.13 & 30.52 & 30.98 & 31.27 & 31.57 & 31.86 & 32.09 & 32.10 & 32.25 & 32.48 & \textbf{32.79} \\
\midrule
OATS Hotels & 0 & 34.82 & 36.97 & \textbf{37.07} & 36.26 & 23.23 & 27.43 & 27.40 & 28.22 & \textbf{28.83} & 28.33 & 28.51 & 28.56 & 28.83 & 28.10 & 28.30 & 28.59 & 28.60 \\
& 10 & 63.29 & 64.20 & \textbf{64.33} & 63.70 & 46.31 & 47.57 & 47.90 & 48.11 & 48.55 & 48.65 & 48.63 & 48.77 & 48.80 & 48.82 & 48.77 & 48.97 & \textbf{49.03} \\
& 50 & 66.80 & 67.29 & \textbf{67.42} & 67.02 & 51.88 & 52.20 & 52.78 & 52.87 & 53.15 & 53.21 & 53.06 & \textbf{53.27} & 53.20 & 53.00 & 53.08 & 53.03 & 52.73 \\
& 100 & 68.14 & \textbf{68.51} & 68.38 & 68.33 & 53.78 & 54.04 & 54.16 & 54.65 & 54.71 & 54.45 & 54.78 & 54.83 & 55.11 & 55.07 & 54.99 & \textbf{55.11} & 55.00 \\
\midrule
\textbf{AVG} & & 56.32 & 57.11 & \textbf{57.30} & 57.07 & 41.54 & 42.80 & 43.22 & 43.51 & 43.72 & 43.89 & 44.12 & 44.27 & \textbf{44.36} & 44.33 & 44.27 & 44.33 & 44.29 \\
\bottomrule
\end{tabular}

}
\caption{\textbf{Variation of the number of views $m$ employed for the majority voting of LLM-MvP.} We compare the performance ($F_1$ score) for different values of $m$ in TASD and ASQP. The best value per row and task is highlighted in \textbf{bold}.}
\label{tab:m_variation_llm_mvp}

\vspace{1.5em}
{%
\setlength{\tabcolsep}{4pt}
\resizebox{1.0\textwidth}{!}{%
\begin{tabular}{@{}ll @{\hskip 8pt} *{5}{c} @{\hspace{12pt}} *{5}{c} @{}}
\toprule
\multirow{2}{*}{\textbf{\# Shots}} & \multirow{2}{*}{\textbf{Approach}} & \multicolumn{5}{c}{\textbf{TASD} ($F_1$)} & \multicolumn{5}{c}{\textbf{ASQP} ($F_1$)} \\
\cmidrule(lr){3-7} \cmidrule(lr){8-12}
& & \textbf{Rest15} & \textbf{Rest16} & \textbf{FlightABSA} & \textbf{Coursera} & \textbf{Hotels} & \textbf{Rest15} & \textbf{Rest16} & \textbf{FlightABSA} & \textbf{Coursera} & \textbf{Hotels} \\
\midrule
\multirow{4}{*}{0} 
& Single-order & 40.48 & 50.10 & 37.93 & 21.13 & 33.93 & 25.59 & 30.13 & 24.32 & 6.26 & 18.91 \\
& LLM-MvP\textsubscript{eff} & \textbf{41.03} & 50.52 & 37.62 & 21.64 & 34.81 & 27.51 & 32.64 & 25.62 & 7.17 & 19.66 \\
& LLM-MvP (w/o GD) & 41.02 & 50.30 & 37.07 & 20.22 & 36.52 & 20.33 & 29.91 & 27.26 & 5.25 & 18.11 \\
\cmidrule(lr){2-12}
& LLM-MvP & 40.88 & \textbf{51.03} & \textbf{39.11} & \textbf{21.81} & \textbf{37.07} & \textbf{32.88} & \textbf{39.15} & \textbf{34.06} & \textbf{9.46} & \textbf{27.40} \\
\midrule
\multirow{4}{*}{10} 
& Single-order & 58.82 & 64.30 & 59.27 & 38.52 & 61.09 & 40.43 & 48.98 & 47.29 & 22.01 & 44.27 \\
& LLM-MvP\textsubscript{eff} & 60.90 & 65.46 & 60.74 & 39.19 & 62.69 & 42.08 & 49.87 & 48.20 & 22.12 & 45.38 \\
& LLM-MvP (w/o GD) & 62.26 & 67.19 & 60.55 & 39.19 & 62.11 & 43.00 & 50.85 & 49.07 & \textbf{23.55} & 46.15 \\
\cmidrule(lr){2-12}
& LLM-MvP & \textbf{63.29} & \textbf{67.79} & \textbf{62.22} & \textbf{41.01} & \textbf{64.33} & \textbf{45.56} & \textbf{52.46} & \textbf{50.47} & 22.93 & \textbf{47.90} \\
\midrule
\multirow{4}{*}{50} 
& Single-order & 63.05 & 69.30 & 66.38 & 44.06 & 64.44 & 47.99 & 54.70 & 53.33 & 27.49 & 47.74 \\
& LLM-MvP\textsubscript{eff} & 65.38 & 71.45 & 67.15 & 44.91 & 65.49 & 49.19 & 56.46 & 53.91 & 27.50 & 48.27 \\
& LLM-MvP (w/o GD) & 66.14 & 71.35 & 67.66 & 47.17 & 67.07 & 49.91 & 57.04 & 54.64 & \textbf{30.12} & 50.53 \\
\cmidrule(lr){2-12}
& LLM-MvP & \textbf{68.18} & \textbf{73.19} & \textbf{68.36} & \textbf{47.22} & \textbf{67.42} & \textbf{50.99} & \textbf{58.08} & \textbf{55.80} & 28.75 & \textbf{52.78} \\
\midrule
\multirow{4}{*}{100} 
& Single-order & 65.27 & 70.74 & 67.59 & 46.09 & 66.06 & 50.70 & 56.36 & 54.90 & 28.14 & 49.66 \\
& LLM-MvP\textsubscript{eff} & 67.37 & 72.96 & 68.44 & 46.84 & 67.24 & 51.99 & 58.16 & 55.62 & 28.49 & 50.48 \\
& LLM-MvP (w/o GD) & 67.97 & 73.70 & 69.56 & 48.83 & 67.06 & 52.65 & 59.82 & 56.82 & \textbf{32.28} & 52.65 \\
\cmidrule(lr){2-12}
& LLM-MvP & \textbf{70.50} & \textbf{74.95} & \textbf{70.11} & \textbf{49.12} & \textbf{68.38} & \textbf{53.40} & \textbf{60.61} & \textbf{57.37} & 30.13 & \textbf{54.16} \\
\bottomrule
\end{tabular}

}
}
\caption{\textbf{Performance ($F_1$) of LLM-MvP variants on all five ABSA datasets.}}
\label{tab:mvp-variants-all}
\end{table*}

\section{Individual Element Orders vs.\ LLM-MvP}\label{appendix:permutations}

Table~\ref{tab:permutations_llm_mvp} compares LLM-MvP against single-order prompting with each individual element order. LLM-MvP outperformed both the default order and the average over all orders in all 20 settings of both tasks. Compared to the best order chosen in hindsight for each setting (oracle), LLM-MvP achieved higher scores in \val{perm.tasd.beats.oracle} (TASD) and \val{perm.asqp.beats.oracle} (ASQP) of 20 settings, and its average $F_1$ (\val{perm.tasd.mvp.avg} for TASD, \val{perm.asqp.mvp.avg} for ASQP) exceeded that of the oracle (\val{perm.tasd.oracle.avg} and \val{perm.asqp.oracle.avg}). 

\begin{table*}[t]
\centering
\setlength{\tabcolsep}{3pt}
\resizebox{\textwidth}{!}{%
\begin{tabular}{@{}l rrrr @{\hskip 8pt} rrrr @{\hskip 8pt} rrrr @{\hskip 8pt} rrrr @{\hskip 8pt} rrrr@{}}
\toprule
\multirow{2}{*}{\textbf{Element order}} & \multicolumn{4}{c}{\textbf{Rest15}} & \multicolumn{4}{c}{\textbf{Rest16}} & \multicolumn{4}{c}{\textbf{FlightABSA}} & \multicolumn{4}{c}{\textbf{Coursera}} & \multicolumn{4}{c}{\textbf{Hotels}} \\
\cmidrule(lr){2-5}\cmidrule(lr){6-9}\cmidrule(lr){10-13}\cmidrule(lr){14-17}\cmidrule(lr){18-21}
 & \textbf{0} & \textbf{10} & \textbf{50} & \textbf{100} & \textbf{0} & \textbf{10} & \textbf{50} & \textbf{100} & \textbf{0} & \textbf{10} & \textbf{50} & \textbf{100} & \textbf{0} & \textbf{10} & \textbf{50} & \textbf{100} & \textbf{0} & \textbf{10} & \textbf{50} & \textbf{100} \\
\midrule
\multicolumn{21}{l}{\textbf{TASD}} \\
\midrule
$(a, c, p)$$^{\dagger}$ & 40.48 & 58.82 & 63.05 & 65.27 & 50.10 & 64.30 & 69.30 & 70.74 & 37.93 & 59.27 & 66.38 & 67.59 & 21.13 & 38.52 & 44.06 & 46.09 & 33.93 & 61.09 & 64.44 & 66.06 \\
$(a, p, c)$ & 39.08 & 59.29 & 63.14 & 65.50 & 50.99 & 65.31 & 68.33 & 70.70 & \textbf{45.08} & 59.53 & 65.57 & 67.08 & \textbf{23.77} & 39.31 & 44.25 & 46.21 & \textbf{39.14} & 60.83 & 65.48 & 64.97 \\
$(c, a, p)$ & 35.56 & 63.51 & 68.09 & 70.56 & 46.57 & 64.74 & 71.50 & 74.05 & 32.14 & 60.14 & 66.62 & 68.90 & 18.95 & 39.57 & 47.20 & 48.37 & 33.02 & 59.79 & 65.40 & 66.10 \\
$(c, p, a)$ & 31.22 & 62.09 & 67.47 & 70.43 & 40.72 & 65.70 & 72.74 & 74.47 & 28.13 & 59.68 & 67.19 & 68.22 & 20.10 & 40.20 & 46.53 & 49.00 & 27.45 & 60.83 & 64.75 & 65.98 \\
$(p, a, c)$ & 39.46 & 61.65 & 65.22 & 66.74 & 47.52 & 65.90 & 70.19 & 72.00 & 29.80 & 58.18 & 65.82 & 68.06 & 19.63 & 37.64 & 45.66 & 46.80 & 32.75 & 60.39 & 65.14 & 65.95 \\
$(p, c, a)$ & 39.39 & \textbf{63.91} & \textbf{68.92} & \textbf{70.63} & 48.66 & 66.49 & \textbf{73.37} & 74.68 & 28.94 & 60.36 & 67.29 & 68.91 & 18.04 & \textbf{41.01} & 46.79 & 48.44 & 22.73 & 61.11 & 65.06 & 66.22 \\
\cmidrule(lr){1-21}
Avg.\ over orders & 37.53 & 61.54 & 65.98 & 68.19 & 47.43 & 65.41 & 70.90 & 72.77 & 33.67 & 59.53 & 66.48 & 68.13 & 20.27 & 39.37 & 45.75 & 47.48 & 31.50 & 60.67 & 65.05 & 65.88 \\
Best order (oracle) & 40.48 & 63.91 & 68.92 & 70.63 & 50.99 & 66.49 & 73.37 & 74.68 & 45.08 & 60.36 & 67.29 & 68.91 & 23.77 & 41.01 & 47.20 & 49.00 & 39.14 & 61.11 & 65.48 & 66.22 \\
LLM-MvP & \textbf{40.88} & 63.29 & 68.18 & 70.50 & \textbf{51.03} & \textbf{67.79} & 73.19 & \textbf{74.95} & 39.11 & \textbf{62.22} & \textbf{68.36} & \textbf{70.11} & 21.81 & \textbf{41.01} & \textbf{47.22} & \textbf{49.12} & 37.07 & \textbf{64.33} & \textbf{67.42} & \textbf{68.38} \\
\midrule
\multicolumn{21}{l}{\textbf{ASQP}} \\
\midrule
$(a, c, p, o)$$^{\dagger}$ & 25.59 & 40.43 & 47.99 & 50.70 & 30.13 & 48.98 & 54.70 & 56.36 & 24.32 & 47.29 & 53.33 & 54.90 & 6.26 & 22.01 & 27.49 & 28.14 & 18.91 & 44.27 & 47.74 & 49.66 \\
$(a, c, o, p)$ & 26.36 & 40.28 & 47.59 & 50.20 & 32.25 & 48.84 & 53.72 & 57.36 & 27.79 & 47.23 & 52.97 & 53.46 & 7.08 & 21.97 & 27.26 & 28.47 & 21.22 & 44.28 & 47.90 & 50.10 \\
$(a, p, c, o)$ & 26.81 & 41.10 & 47.53 & 50.03 & 33.63 & 49.40 & 53.87 & 55.92 & 32.57 & 46.78 & 53.74 & 54.36 & \textbf{11.19} & 22.50 & 27.54 & 29.13 & 22.87 & 44.73 & 47.89 & 49.55 \\
$(a, p, o, c)$ & 27.67 & 41.02 & 47.18 & 50.20 & 34.83 & 49.31 & 54.45 & 56.07 & \textbf{36.98} & 48.51 & 54.39 & 55.01 & 9.70 & 22.48 & 27.97 & 28.50 & 25.69 & 45.81 & 50.23 & 51.68 \\
$(a, o, c, p)$ & 25.29 & 40.74 & 47.31 & 49.75 & 30.46 & 47.77 & 53.59 & 55.75 & 26.06 & 46.40 & 53.70 & 55.63 & 7.25 & 22.00 & 28.28 & 29.30 & 20.16 & 44.65 & 49.19 & 50.51 \\
$(a, o, p, c)$ & 24.77 & 40.69 & 47.46 & 49.95 & 28.16 & 49.11 & 53.88 & 55.98 & 26.43 & 46.86 & 53.45 & 54.53 & 7.49 & 21.70 & 28.62 & 29.15 & 20.51 & 44.69 & 49.60 & 50.69 \\
$(c, a, p, o)$ & 25.70 & 42.76 & 50.37 & 52.48 & 30.82 & 50.84 & \textbf{58.35} & 59.73 & 29.29 & 47.49 & 53.26 & 54.95 & 6.02 & 22.22 & 28.35 & 28.80 & 20.98 & 42.46 & 47.04 & 49.12 \\
$(c, a, o, p)$ & 23.77 & 42.02 & 49.18 & 52.53 & 31.02 & 50.41 & 56.79 & 59.48 & 28.10 & 46.51 & 53.22 & 54.56 & 7.00 & 21.94 & 28.54 & 29.74 & 20.96 & 42.81 & 47.15 & 49.36 \\
$(c, p, a, o)$ & 28.34 & 43.11 & 49.78 & 52.14 & 33.12 & 51.47 & 57.72 & 59.07 & 29.02 & 46.95 & 52.58 & 54.71 & 11.14 & 22.41 & 28.01 & 29.36 & 25.53 & 43.82 & 46.32 & 48.53 \\
$(c, p, o, a)$ & 30.15 & 42.77 & 48.82 & 52.26 & 34.38 & 50.48 & 56.98 & 58.61 & 31.36 & 46.53 & 52.32 & 55.35 & 9.95 & 22.83 & 27.29 & 28.61 & 25.12 & 43.23 & 49.35 & 51.12 \\
$(c, o, a, p)$ & 28.84 & 42.42 & 48.49 & 51.36 & 34.59 & 49.18 & 55.96 & 57.57 & 29.55 & 46.31 & 52.12 & 54.52 & 9.19 & 22.06 & 26.82 & 28.85 & 24.41 & 42.12 & 48.67 & 51.17 \\
$(c, o, p, a)$ & 27.46 & 41.62 & 48.16 & 51.14 & 31.41 & 48.86 & 55.69 & 57.90 & 23.37 & 46.67 & 52.21 & 53.82 & 6.32 & 22.51 & 27.07 & 28.62 & 16.32 & 41.41 & 49.41 & 51.02 \\
$(p, a, c, o)$ & 20.20 & 42.62 & 49.55 & 51.60 & 24.08 & 50.31 & 57.65 & 59.70 & 17.53 & 48.73 & 53.87 & 56.04 & 6.27 & 21.14 & 28.29 & 29.82 & 14.39 & 45.11 & 47.83 & 49.89 \\
$(p, a, o, c)$ & 20.22 & 42.91 & 49.39 & 52.24 & 22.14 & 49.99 & 56.46 & 58.62 & 17.58 & 47.30 & 54.28 & 55.95 & 6.49 & 21.88 & \textbf{29.12} & 29.89 & 15.89 & 45.23 & 50.37 & 51.22 \\
$(p, c, a, o)$ & 29.45 & 44.25 & 49.87 & 53.08 & 33.29 & 50.35 & 57.84 & 59.81 & 30.05 & 47.54 & 54.02 & 55.97 & 10.43 & 21.90 & 28.76 & 29.54 & 21.25 & 42.71 & 46.39 & 49.20 \\
$(p, c, o, a)$ & 32.86 & 42.84 & 49.28 & 52.22 & 36.31 & 49.71 & 56.98 & 58.81 & 32.53 & 46.83 & 53.15 & 55.21 & 8.58 & 23.16 & 27.50 & 28.56 & 25.58 & 41.92 & 49.15 & 51.62 \\
$(p, o, a, c)$ & 32.84 & 42.61 & 49.95 & 52.12 & 34.57 & 49.88 & 56.08 & 57.85 & 30.65 & 48.05 & 53.68 & 55.03 & 9.61 & \textbf{23.64} & 28.60 & 29.01 & \textbf{29.51} & 42.70 & 49.39 & 51.41 \\
$(p, o, c, a)$ & 30.70 & 44.52 & 50.15 & 52.84 & 35.69 & 49.98 & 56.52 & 59.89 & 28.96 & 47.90 & 53.53 & 55.04 & 11.18 & 23.27 & 28.29 & 29.09 & 28.45 & 41.95 & 48.54 & 50.50 \\
$(o, a, c, p)$ & 28.62 & 41.85 & 46.64 & 49.01 & 35.68 & 48.30 & 52.95 & 55.54 & 20.92 & 47.75 & 52.70 & 54.77 & 7.66 & 21.90 & 26.26 & 26.71 & 24.37 & 42.90 & 49.03 & 50.38 \\
$(o, a, p, c)$ & 25.21 & 41.56 & 47.22 & 48.12 & 29.38 & 48.43 & 53.42 & 55.87 & 21.13 & 47.48 & 52.60 & 54.73 & 7.95 & 22.15 & 26.56 & 26.95 & 25.61 & 43.26 & 48.87 & 50.21 \\
$(o, c, a, p)$ & 26.10 & 42.75 & 47.85 & 48.91 & 33.41 & 47.81 & 54.82 & 56.02 & 21.49 & 46.79 & 52.53 & 53.77 & 8.94 & 20.89 & 26.54 & 26.78 & 24.49 & 40.65 & 47.58 & 49.54 \\
$(o, c, p, a)$ & 26.70 & 41.80 & 47.21 & 48.91 & 31.25 & 46.92 & 53.65 & 56.79 & 18.37 & 46.35 & 51.59 & 54.34 & 7.92 & 20.76 & 25.71 & 26.58 & 23.46 & 39.78 & 47.54 & 49.57 \\
$(o, p, a, c)$ & 24.33 & 39.48 & 45.94 & 48.24 & 28.47 & 47.58 & 52.62 & 55.23 & 18.38 & 46.83 & 52.77 & 54.40 & 7.45 & 21.60 & 25.98 & 26.44 & 21.64 & 41.92 & 48.27 & 49.28 \\
$(o, p, c, a)$ & 24.69 & 41.31 & 46.82 & 49.34 & 28.82 & 47.48 & 53.88 & 56.84 & 14.05 & 46.46 & 52.82 & 54.55 & 6.10 & 21.02 & 26.16 & 26.00 & 19.05 & 40.51 & 47.42 & 49.76 \\
\cmidrule(lr){1-21}
Avg.\ over orders & 26.78 & 41.98 & 48.32 & 50.81 & 31.58 & 49.23 & 55.36 & 57.53 & 25.69 & 47.15 & 53.12 & 54.82 & 8.22 & 22.08 & 27.54 & 28.42 & 22.35 & 43.04 & 48.37 & 50.21 \\
Best order (oracle) & 32.86 & 44.52 & 50.37 & 53.08 & 36.31 & 51.47 & 58.35 & 59.89 & 36.98 & 48.73 & 54.39 & 56.04 & 11.19 & 23.64 & 29.12 & 29.89 & 29.51 & 45.81 & 50.37 & 51.68 \\
LLM-MvP & \textbf{32.88} & \textbf{45.56} & \textbf{50.99} & \textbf{53.40} & \textbf{39.15} & \textbf{52.46} & 58.08 & \textbf{60.61} & 34.06 & \textbf{50.47} & \textbf{55.80} & \textbf{57.37} & 9.46 & 22.93 & 28.75 & \textbf{30.13} & 27.40 & \textbf{47.90} & \textbf{52.78} & \textbf{54.16} \\
\bottomrule
\end{tabular}

}
\caption{\textbf{Performance ($F_1$) of single-order prompting with each of the 6 (TASD) and 24 (ASQP) element orders compared to LLM-MvP (Gemma 4 (31B)).} Column headers 0, 10, 50, and 100 denote the number of few-shot examples. $^{\dagger}$ marks the default order used for Single-order in all other tables. \textit{Avg.\ over orders} averages over all element orders. \textit{Best order (oracle)} picks the best element order per column in hindsight. No individual order is consistently best, and LLM-MvP exceeds even the oracle in average $F_1$. The best score per column and task, excluding the oracle, is highlighted in \textbf{bold}.}
\label{tab:permutations_llm_mvp}
\end{table*}

\section{Component Analysis}\label{appendix:mvp-variants-all}

Table~\ref{tab:mvp-variants-all} extends the component analysis in Table~\ref{tab:mvp-variants} to all five datasets.

\end{document}